\documentclass[twocolumn]{svjour3}          
\smartqed  

\usepackage{graphicx}
\usepackage[T1]{fontenc} 
\usepackage[utf8]{inputenc}
\usepackage[colorlinks, citecolor = blue, urlcolor = magenta]{hyperref}

\usepackage{cuted}
\usepackage[switch]{lineno}
\usepackage{amsmath}
\usepackage{amssymb}
\usepackage{booktabs}
\usepackage{multirow}
\usepackage{algorithm}
\usepackage{algpseudocode}
\usepackage{subcaption}

\begin{document}

\title{Distilled Pooling Transformer Encoder for Efficient Realistic Image Dehazing}

\author{Le-Anh Tran \and Dong-Chul Park}



\institute{Le-Anh Tran \at
        leanhtran@mju.ac.kr \\
}

\date{Received: date / Accepted: date}

\maketitle

\vspace*{-5cm}

\begin{strip}
\begin{abstract}

This paper proposes a lightweight neural network designed for realistic image dehazing, utilizing a \textbf{D}istilled \textbf{P}ooling \textbf{T}ransformer \textbf{E}ncoder, named \textbf{DPTE-Net}. Recently, while vision transformers (ViTs) have achieved great success in various vision tasks, their self-attention (SA) module’s complexity scales quadratically with image resolution, hindering their applicability on resource-constrained devices. To overcome this, the proposed DPTE-Net substitutes traditional SA modules with efficient pooling mechanisms, significantly reducing computational demands while preserving ViTs’ learning capabilities. To further enhance semantic feature learning, a distillation-based training process is implemented which transfers rich knowledge from a larger teacher network to DPTE-Net. Additionally, DPTE-Net is trained within a generative adversarial network (GAN) framework, leveraging the strong generalization of GAN in image restoration, and employs a transmission-aware loss function to dynamically adapt to varying haze densities. Experimental results on various benchmark datasets have shown that the proposed DPTE-Net can achieve competitive dehazing performance when compared to state-of-the-art methods while maintaining low computational complexity, making it a promising solution for resource-limited applications. The code of this work is available at \href{https://github.com/tranleanh/dpte-net}{https://github.com/tranleanh/dpte-net}.

\end{abstract}
\end{strip}


\section{Introduction}
\label{sec:intro}

Image dehazing is an ill-posed and challenging computer vision problem that aims to enhance the visibility of images captured under inclement weather conditions caused by natural phenomena like haze and fog. Prior-based methods such as \cite{he2010single,zhu2015fast,berman2016non,tran2024single} achieved early success in the field that rely on domain-specific assumptions combined with the haze imaging model. However, those algorithms have limited effectiveness as they are built based on handcrafted priors which may be inapplicable to some challenging cases such as densely hazy scenes.

On the other hand, owing to the revolutionary improvement in performance achieved by convolutional neural networks (CNNs) over various vision tasks, studies on image dehazing have also gradually shifted from prior-based methods to deep learning-based schemes \cite{chen2019gated,liu2019griddehazenet,qin2020ffa,tran2022encoder}. 
Moreover, the breakthrough of an emerging neural network class, vision transformers (ViTs) \cite{dosovitskiy2020vit}, has opened up a new direction for neural architecture design. Several studies \cite{zamir2022restormer,guo2022image} have also adopted standard ViTs with self-attention (SA) to gain image dehazing performance. However, those performance improvements come at the cost of model complexity and inference latency. Even though SA is highly effective in capturing long-range pixel interactions, its complexity grows quadratically with the spatial resolution \cite{zamir2022restormer,pan2022edgevits}. Also, the SA module may not meet the requirements of general resource-constrained mobile devices that have not been integrated with latest deep learning operators. Accordingly, a fallback to CPU-reference implementation can result in unreasonably high latency \cite{chiang2020deploying}. Since haze removal usually serves as a pre-processing process for other high-level vision tasks such as classification or object detection, the trade-off between effectiveness and efficiency appears quite pronounced, whereas recently established ViT-based dehazing schemes are considered heavyweight. For instance, Dehamer \cite{guo2022image} has 132.5M parameters while Restormer \cite{zamir2022restormer} requires 564G multiply-accumulate operations (MACs) to process a 512$\times$512 input. To enhance efficiency, ViT models should be lightweight and employ non-complex architectures that involve straightforward operations.

On the other extreme, knowledge distillation (KD) \cite{hinton2015distilling,tran2024soft} has been known as an effectual machine learning technique to learn small and effective neural networks. Inspired by the success of KD in various vision tasks \cite{hinton2015distilling,romero2014fitnets}, several studies \cite{hong2020distilling,wu2020knowledge,suresh2022rich,tran2024lightweight} have also proposed to apply KD to image dehazing. However, most KD-based approaches only utilized the knowledge extracted from clean images (hard knowledge) and ignored exploiting the concise knowledge encoded from hazy images (soft knowledge). Contrarily, we claim that soft knowledge can be aptly exploited to train more effective student networks via a two-stage training procedure.

This study is primarily targeted toward two main issues: 1) how effective an SA-free ViT-based dehazing network can be; and 2) how to aptly exploit soft knowledge to train dehazing networks. For these objectives, we propose an efficient dehazing network with a \textbf{D}istilled \textbf{P}ooling \textbf{T}ransformer \textbf{E}ncoder, called \textbf{DPTE-Net}. The proposed DPTE-Net consists of multi-level SA-free ViT-like blocks in its encoder and is trained based on a generative adversarial network (GAN) framework combined with the KD training technique. In the proposed framework, knowledge for guiding the student network is exploited from a dehazing teacher network, simulating a same-task teacher-student scheme (soft knowledge distillation), an important aspect that has never been examined thoroughly in other KD-based dehazing studies. Additionally, our student network is much more compressed while those in other relevant existing methods \cite{hong2020distilling,wu2020knowledge} have a similar network complexity to the teacher models. 
Even though ViT models have been well-studied in recent years for various high-level vision tasks, there is currently a lack of investigation into an SA-free ViT-based architecture for image restoration. This is necessary as neural network approaches may result in variations in performance when applied to different types of tasks and data. In addition, we take advantage of the transmission feature estimated by prior-based methods and utilize it as a weight mask for a \textit{transmission-aware loss} in order to optimize the network based on different haze densities. Inspired by \textit{focal loss} \cite{lin2017focal} in object detection, transmission-aware loss down-weights and up-weights the loss assigned to sparse and dense haze regions, respectively. 
A typical comparison on a benchmark dataset, NH-HAZE \cite{ancuti2020nh}, is presented in Fig. \ref{fig01:chart}, which shows that DPTE-Net achieves competitive performance and appears to be more robust when compared to the other approaches. The robustness and effectiveness of the proposed network stem from the following key contributions:

\begin{figure}
  \centering
  \includegraphics[width=0.9\linewidth]{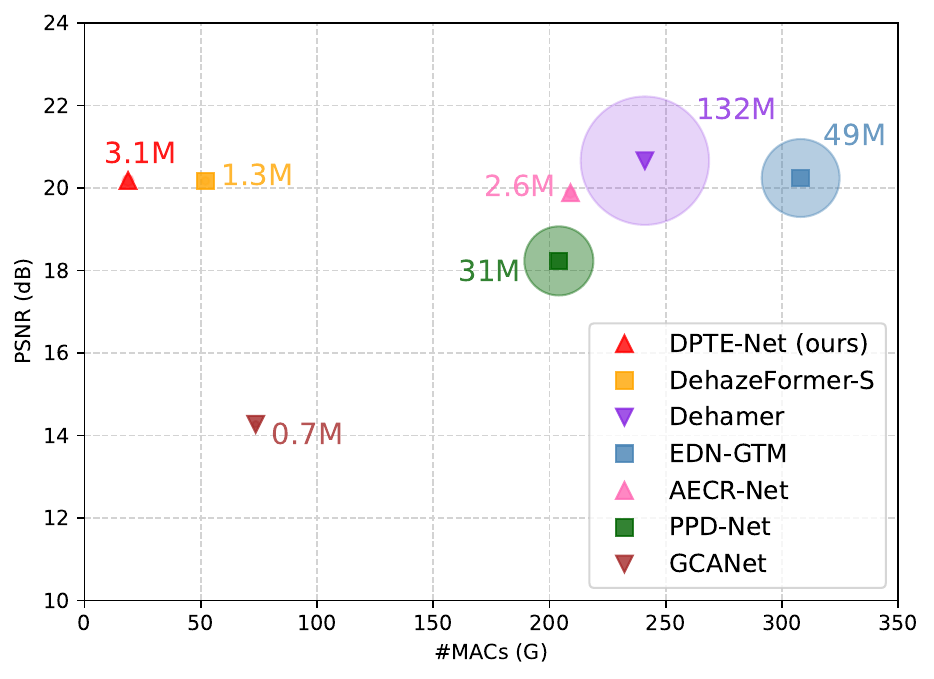}
  \caption{Comparison of various methods on the NH-HAZE dataset. The proposed DPTE-Net achieves competitive performance with low computational complexity. The circle radius denotes the model size.}
  \label{fig01:chart}
\end{figure}

\begin{itemize}

  \item The principal advancement is achieved by substituting the computationally expensive SA module in ViTs with a pooling mechanism. This approach reduces computational complexity while effectively capturing global contextual information, thereby facilitating efficient dehazing.
  
  \item The implementation of a two-stage knowledge distillation training process enables the student model to effectively replicate the semantic feature extraction capability of a larger teacher model, particularly in the encoding stage with enhanced pooling token mixing, thus improving the student model's generalizability.

  \item The utilization of the transmission-aware loss with a weight mask allows the network to adaptively adjust its learning process according to varying haze densities, promoting consistent performance across diverse environmental conditions.
  
  \item The proposed network has been trained and evaluated across multiple benchmark datasets, demonstrating a favorable trade-off between performance and computational complexity as compared to existing methods in the literature.
  
\end{itemize}

The rest of this paper is organized as follows. Section \ref{sec:relatedworks} briefly reviews related research works. In Section \ref{sec:proposedframework}, the proposed framework is elaborated. The experiments and analyses on various benchmark datasets are provided in Section \ref{sec:experiments}. Section \ref{sec:conclusions} concludes the paper.

\section{Related Works}
\label{sec:relatedworks}

\begin{figure*}[t]
    \centering
    \includegraphics[width=1.0\linewidth]{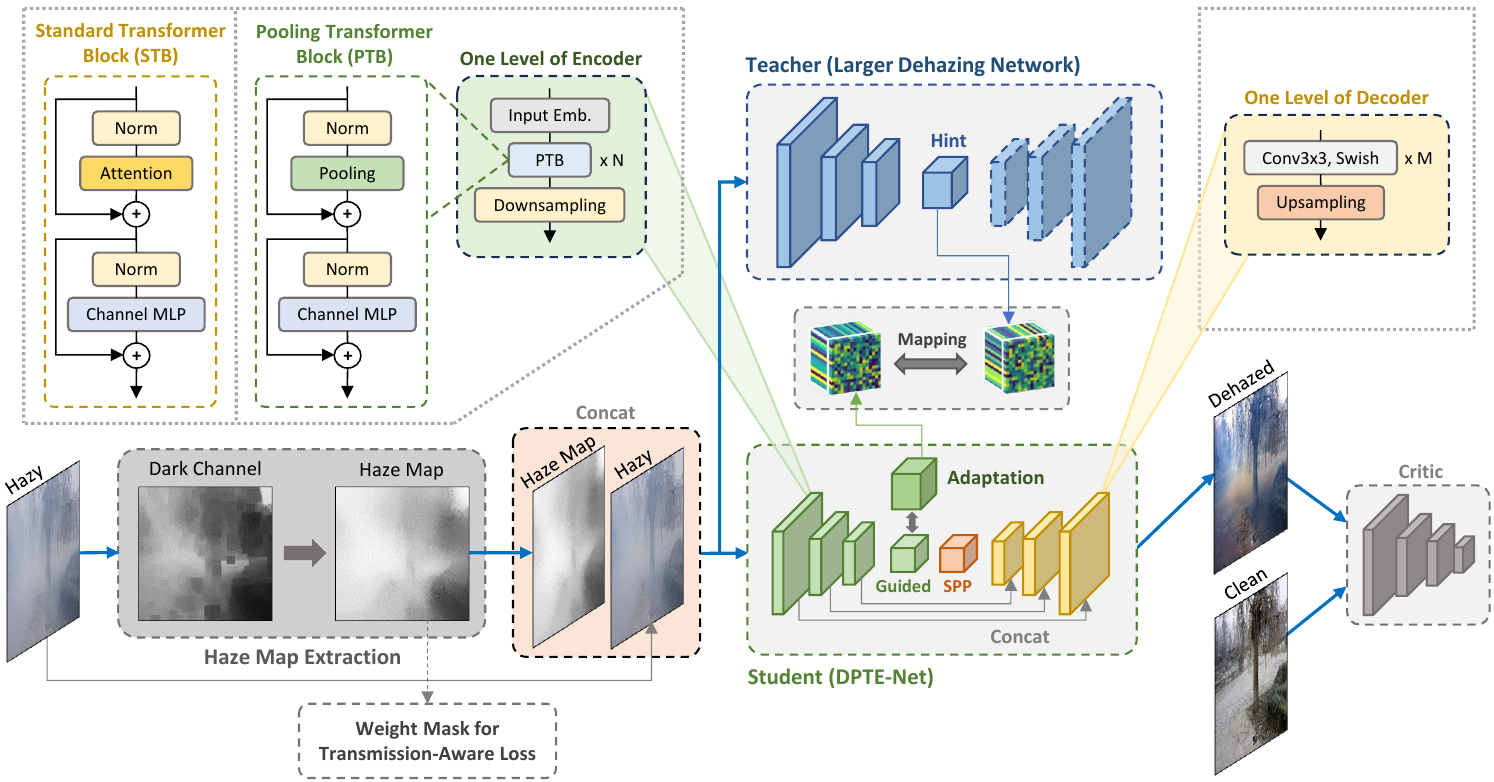}
    \caption{The proposed framework.}
    \label{fig:proposed-framework}
\end{figure*}

Traditional or prior-based approaches have been considered to achieve the initial success in the field, which leverage domain-specific knowledge or assumptions to guide the dehazing process \cite{he2010single,berman2016non,wang2017fast,galdran2018image,tran2024haze}. He \textit{et al.} \cite{he2010single} proposed the dark channel prior (DCP) algorithm which was developed based on a key observation that most local patches in haze-free outdoor images contain some pixels with very low intensities in at least one color channel. Wang \textit{et al.} \cite{wang2017fast} introduced a linear transformation method by assuming that a linear relationship exists in the minimum channel between the hazy image and the haze-free image. Galdran \cite{galdran2018image} presented a multi-scale Laplacian blending scheme to merge a set of multiply-exposed images and produce a haze-free image without relying on the haze formation model. These methods are computationally efficient and able to recover haze-free images with an adequate level of accuracy but struggle with complex conditions where their assumptions are inapplicable.

In recent years, learning-based methods with the great success of data-driven CNN architectures have gained significant progress due to their ability to learn complex patterns from large image databases. Early works on CNN-based approaches such as DehazeNet \cite{cai2016dehazenet}, MSCNN \cite{ren2016single}, and AOD-Net \cite{li2017aod} perform dehazing through the haze imaging model. However, these models are less effective in challenging and complex cases such as densely hazy scenes where the haze imaging model may be inapplicable. Accordingly, some follow-up approaches such as FFA-Net \cite{qin2020ffa} and AECR-Net \cite{wu2021contrastive} have tried to overcome that shortcoming by directly learning the hazy-to-clean mapping via end-to-end training. These types of methods further adopt attention mechanisms, feature enhancement modules, or contrastive learning for better dehazing performances. Besides, various multi-stage methods have been designed, including DRN \cite{li2022single}, UHD \cite{zheng2021ultra}, and PSD \cite{chen2021psd}. These sequential approaches break down complex dehazing tasks into manageable steps, where each step typically focuses on refining specific aspects of the image to progressively improve clarity and detail. In addition to that, several studies have also leveraged the generalization power of image-generation models (GAN-based architectures) to improve dehazing outcomes. Notable approaches following these strategies include EPDN \cite{qu2019enhanced}, Cycle-Dehaze \cite{engin2018cycle}, RefineDNet \cite{zhao2021refinednet}, and EDN-GTM \cite{tran2024encoder}. In recent years, with the growing popularity of ViTs which have shown significant performance gains in high-level vision tasks, several studies have also proposed ViT-based approaches to image restoration and obtained promising performances. For example, Dehamer \cite{guo2022image} adopts ViT with transmission-aware 3D position embedding; Restormer \cite{zamir2022restormer} captures long-range pixel interactions via multi-head self-attention (MHSA); DehazeFormer \cite{song2023vision} carries out various improvements on the Swin Transformer \cite{liu2021swin} baseline by modifying normalization layer, activation function, and spatial information aggregation; TransWeather \cite{valanarasu2022transweather} employs intra-patch transformer blocks in its encoder to enhance attention inside the patches to effectively address adverse weather removal problem. Even though the aforementioned methods have produced satisfactory results, they do not guarantee an optimal balance for the performance-complexity trade-off. For instance, EDN-GTM and Dehamer correspondingly have 49.3M and 132.5M parameters, while FFA-Net and Restormer consume 1,151G and 564G MACs to process a 512$\times$512 input, respectively, making them infeasible to be deployed to resource-limited platforms.

On the other extreme, KD aims to train a compact student network to learn the prediction behavior of a teacher network. Several KD-based dehazing methods such as KDDN \cite{hong2020distilling}, KTDN \cite{wu2020knowledge}, MC-FSRCNN \cite{suresh2022rich} have also been proposed in the literature. However, existing approaches only exploited the knowledge from the teacher networks trained on clean-to-clean translation tasks to optimize a student network trained for dehazing task. For example, the teacher networks in KDDN and KTDN are trained to perform a copying task while MC-FSRCNN transfers knowledge from a super-resolution teacher to a dehazing student. We argue that these training strategies may not always guarantee the suitableness of the transferred knowledge since the convergence feature spaces of the teacher and student networks that perform heterogeneous tasks may not be close in feature space. Hence, mapping the features of a dehazing student network with those of a teacher model that performs another task is not likely to guarantee the student network to achieve its utmost generalization capability.

\section{Proposed Framework}
\label{sec:proposedframework}

Motivated through observations on existing ViT-based and KD-based dehazing approaches which still show inherent shortcomings, in this study, a strong emphasis is placed on efficiency in an attempt to deliver a better effectiveness-complexity trade-off and promote efficient dehazing applications. First, we propose to design an SA-free ViT-based dehazing network. Specifically, the SA block is replaced with pooling operator, such replacement has been proven to yield competitive performance as compared to SA-based ViT models while reducing the computational complexity \cite{yu2022metaformer}. Second, we revisit the KD method for image dehazing and adopt a same-task teacher-student KD framework to provide the student network with more suitable knowledge and better convergence results.

\subsection{Architectures}
\label{subsec:distillededngtm}

\begin{algorithm}[t]
\caption{Implementation of the PTB module in a TensorFlow-like pseudocode.}\label{algo1}
\hspace*{0.0cm} \textbf{Input:} Feature map $X$ with shape $c \times h \times w$, MLP ratio $r$ \\
\hspace*{0.0cm} \textbf{Output:} Feature map $Z$ with shape $c \times h \times w$ \\
\hspace*{0.0cm} \textbf{PTB:}
\begin{algorithmic}[1]
\State \textcolor{magenta}{\textit{\# The 1st sub-block:}}
\State $X_n = \mathrm{BatchNormalization}()(X)$
\State $X_p = \mathrm{AveragePooling2D}(ksize=3\times3, stride=1)(X_n)$ 
\State $Y = \mathrm{Add}()([X,X_p])$
\State 
\State \textcolor{magenta}{\textit{\# The 2nd sub-block:}}
\State $Y_n = \mathrm{BatchNormalization}()(Y)$
\State $Y_1 = \mathrm{Conv2D}(filters=c \times r,ksize=1\times1, stride=1)(Y_n)$ 
\State $Y_1 = \mathrm{Swish}()(Y_1)$
\State $Y_2 = \mathrm{Conv2D}(filters=c,ksize=1\times1, stride=1)(Y_1)$ 
\State $Z = \mathrm{Add}()([Y,Y_2])$
\end{algorithmic}
\end{algorithm}

\subsubsection{Pooling Transformer Block (PTB)}

The ultimate goal of this study is to develop a ViT-like module to take advantage of the learning prowess of ViTs while constructing it as a non-complex structure. In a general ViT block, an input $I$ is first embedded as a sequence of features (or tokens) \cite{yu2022metaformer}:
\begin{equation}
    X = \mathrm{InputEmb}(I),
    \label{eq2}
\end{equation} 
where $X \in \mathbb{R}^{N \times C}$ denotes the token sequence with length $N$ and channel dimension $C$. The token sequence $X$ is then passed through two sub-blocks, in which the first block is expressed as:
\begin{equation}
    Y = X + \mathrm{TokenMixer}(\mathrm{Norm}(X)), \label{eq3}
\end{equation} 
while the second block is expressed as:
\begin{equation}
    Z = Y + \sigma(\mathrm{Norm}(Y)W_1)W_2, \label{eq4}
\end{equation} 
where $\mathrm{TokenMixer}(.)$ is the token mixing operation which propagates information among tokens, $\mathrm{Norm}(.)$ denotes the normalization layer, $\sigma(.)$ represents a non-linear activation, while $W_1$, $W_2$ are two layers of channel multilayer perceptron (MLP). Furthermore, considering empirical evidence demonstrating that the general structure of ViTs contributes mostly to the success of ViT-based models rather than the SA module \cite{yu2022metaformer}, we adopt the general structure of the ViT module and use average pooling for the $\mathrm{TokenMixer}(.)$. As a result, the first sub-block includes a normalization layer followed by a 2D average pooling layer. The pooling layer in this stage can be utilized to replace the attention mechanism in the standard ViT block for the token mixing operation since it can produce competitive performance with fewer MACs and parameters as well as keeping the model computational complexity at a low-level. On the other hand, the second sub-block is comprised of a normalization layer followed by a channel MLP to propagate the features among channels. The channel MLP can be constructed by adopting two point-wise convolution layers with an MLP expansion ratio $r$ and a non-linear activation between them. Skip connection is also applied to each sub-block. Specifically, we configure $\mathrm{InputEmb}$, $\mathrm{Norm}$, and $\sigma$ as $Conv3\times3$, batch normalization \cite{ioffe2015batch}, and Swich activation \cite{ramachandran2017searching}, respectively. A TensorFlow-like pseudocode of the PTB module is presented in Algorithm \ref{algo1}.

\subsubsection{Framework Architectures}

Motivated by various GAN-based frameworks that have offered significant advantages for image restoration in recent years \cite{qu2019enhanced,engin2018cycle,zhao2021refinednet,tran2024encoder}, the proposed network is trained based on a GAN structure, comprising a generative model and a critic model, with a 4-level U-Net \cite{ronneberger2015u,tran2019robust} used as the backbone, as illustrated in Fig. \ref{fig:proposed-framework}. The generative dehazing network (DPTE-Net) is derived by making various modifications on the backbone’s architecture, whereas the critic model is constructed as the backbone’s encoder.

As depicted in Fig. \ref{fig:proposed-framework}, the input hazy image is first passed through a DCP-based haze map extraction module to generate a transmission-aware mask (or haze map) via two steps: dark channel extraction and smoothing filtering. The haze map is then used for two purposes: 1) an additional channel of the input and 2) a weight mask for transmission-aware loss calculation (described in Section \ref{subsec:lossfunctions}). Note that we chose DCP to estimate transmission due to its robustness in computation \cite{guo2022image}, other methods can also be applicable. In the generative network’s design, each encoding level adopts one input embedding layer followed by multiple PTB modules and a max pooling layer for down-sampling, whereas the decoding part keeps the original structure as in the backbone but replaces the default ReLU function with the Swish function as the main activation. Inspired by \cite{tran2022anovel}, spatial pyramid pooling (SPP) is also integrated into the bottleneck to increase the receptive field and separate significant context features. The network is designed with small numbers of channels for the sake of complexity and memory. Specifically, the network has 5 stages (including the first/last stage and 4 down/up-sampling steps) with the shapes of $H \times W \times 16$, $\frac{H}{2} \times \frac{W}{2} \times 32$, $\frac{H}{4} \times \frac{W}{4} \times 64$, $\frac{H}{8} \times \frac{W}{8} \times 128$, and $\frac{H}{16} \times \frac{W}{16} \times 256$, where $H$ and $W$ denote height and width of input image. In fact, we have examined the effects of PTB and sub-pixel convolution \cite{shi2016real} for decoding, but these configurations do not provide good performances for our case (discussed in Section \ref{subsubsec:dptenet_decoder}). Therefore, we keep utilizing a simple form of decoder but with a replacement of activation. This choice, however, can yield a satisfactory performance while maintaining the low complexity of the network.

\begin{figure}
  \centering
  \includegraphics[width=0.95\linewidth]{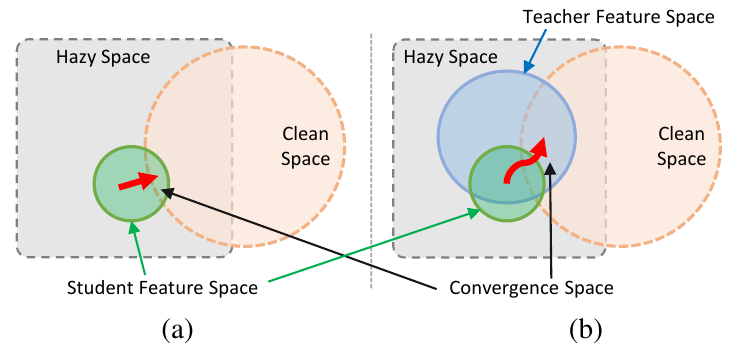}
  \caption{Convergence process of a dehazing network (a) with/(b) without the teacher network's guidance.}
  \label{fig03:featurespaces}
\end{figure}

On the other hand, a critic model is used to evaluate the DPTE-Net’s output during training. Unlike a standard GAN discriminator, which uses binary classification to determine if images are real or fake, the critic, inspired by WGAN \cite{arjovsky2017wasserstein}, assesses image quality by measuring the Wasserstein distance between generated and real image distributions. The network is trained by optimizing the Wasserstein-based loss function. Using a critic in GAN frameworks is generally more effective for image restoration, as it provides stable feedback, resulting in higher-quality images with finer details.

\begin{figure}
  \centering
  \includegraphics[width=0.95\linewidth]{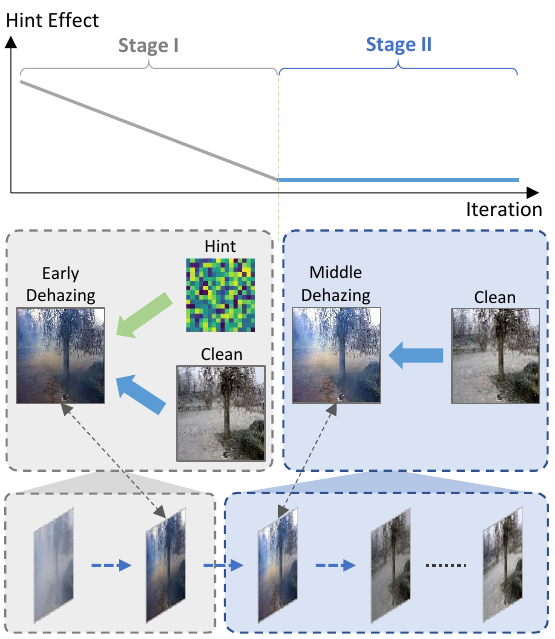}
  \caption{Two-stage training process.}
  \label{fig03as:featurespaces}
\end{figure}

\subsubsection{Knowledge Transfer}

Distinct from \cite{hong2020distilling,wu2020knowledge,suresh2022rich} which only investigated hard knowledge, we aptly exploit soft knowledge to train the student network. We hypothesize the convergence process of a dehazing network with/without the teacher network's guidance in Fig. \ref{fig03:featurespaces}. Generally, the weight exploration capability of a small network tends to be limited by its size and a larger network usually has a larger feature search space compared to a smaller network \cite{brutzkus2019larger}. When a teacher network is trained on the same task as the student, the convergence space of the teacher can be closer to that of the student, and thus the teacher can deliver a bridge to guide the student to a better position in the feature convergence space \cite{romero2014fitnets}, as illustrated in Fig. \ref{fig03:featurespaces}b. To this end, we first train a teacher dehazing network and utilize its bottleneck as \textit{guidance} (or \textit{hint}) since the bottleneck is known as the part that contains the richest semantic features in encoder-decoder-like networks. The EDN-GTM model \cite{tran2024encoder} is chosen as the teacher network as it has a general encoder-decoder structure and its performance has been validated on various benchmarks. Since the bottlenecks of student and teacher do not have the same shapes, an adaptation layer is added to match the shapes of mapped features and to provide a smooth transition between the feature spaces of the student and teacher models. We also found that a PTB module followed by a $Conv1\times1$ layer can be beneficial for the adaptation layer (discussed in Section \ref{subsubsec:adaptation}).

\subsubsection{Two-stage Training}

In an ill-posed vision task like dehazing, even a large network cannot achieve a perfect convergence, hence, fully distilling soft knowledge from the teacher possibly leads the student to a suboptimal convergence (weighted average solution of the teacher space and clean space). Therefore, soft knowledge should be aptly exploited through a two-stage training procedure, as illustrated in Fig. \ref{fig03as:featurespaces}. In Stage I, hint (soft knowledge) is combined with clean data for training, and the impact of hint is decreased gradually. In Stage II, the student learns the hazy-to-clean translation by itself when receiving adequate guidance from the teacher. This process aims to guide the student to a better position in feature space in the first phase of training before leaving it to learn independently, as illustrated in Fig. \ref{fig03:featurespaces}b. This two-stage training process can be managed by a loss balancing decay $\lambda$ which is calculated as follows:
\begin{equation}
\lambda = 
\begin{cases} 
1 - \frac{e}{\delta E}, & \mbox{if } e \le \delta E \\ 
0, & \mbox{if } e > \delta E 
\end{cases}
\label{eq5}
\end{equation}
where $e$ is the epoch index, $E$ is the total number of training epochs, and $\delta$ ($\delta \in (0, 1]$) is the parameter that controls the period of Stage I. For instance, with $\delta=0.5$, the impact of hint is 100\% at the beginning ($\lambda=1$) and is linearly dropped until the $(E/2)\textsuperscript{th}$ epoch ($\lambda=0$).

\begin{figure}
  \centering
   \includegraphics[width=1.0\linewidth]{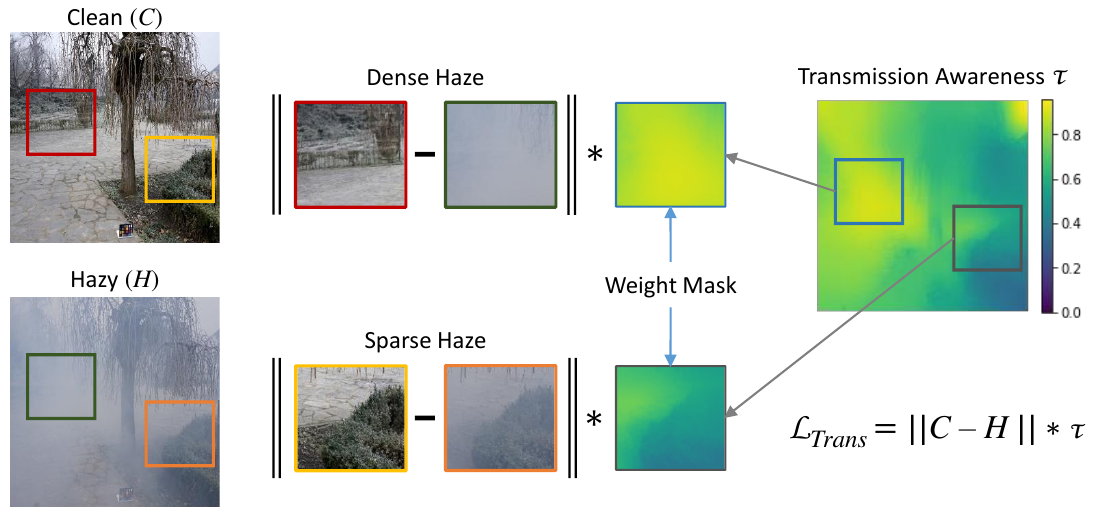}
  \caption{Transmission-aware loss leverages the transmission information in order to optimize the network's generalization based on different degrees of haze density.}
  \label{fig:trans-aware-loss}
\end{figure}

\begin{table*}
  \centering
  \caption{Quantitative comparisons on I-HAZE \cite{ancuti2018ihaze} and O-HAZE \cite{ancuti2018haze}. The \textbf{bold} numbers denote the best results under each category, and \underline{underline} numbers indicate that our results fall in the top three best results, while it is worth noting that most methods under comparison have more than 10$\times$ MACs compared to our model.}
  \resizebox{0.65\textwidth}{!}{
  \begin{tabular}{ccccccrr}
    \toprule
    \multirow{2}{*}{Type} & \multirow{2}{*}{Method} & \multicolumn{2}{c}{I-HAZE} & \multicolumn{2}{c}{O-HAZE} & \multirow{2}{*}{\#Params$\downarrow$} & \multirow{2}{*}{MACs$\downarrow$}\\
    \cmidrule{3-6}
           & & PSNR$\uparrow$ & SSIM$\uparrow$ & PSNR$\uparrow$ & SSIM$\uparrow$ & \\
    \midrule

    \multirow{5}{*}{Prior-based} & CAP \cite{zhu2015fast} & 12.24 & 0.6065 & 16.08 & 0.5965 & - & - \\
    & BCCR \cite{meng2013efficient} & 14.15 & 0.7046 & 14.07 & 0.5103 & - & - \\
    & DCP \cite{he2010single} & 14.43 & 0.7516 & 16.78 & 0.6532 & - & - \\
    & NLID \cite{berman2016non} & 14.12 & 0.6537 & 15.98 & 0.5849 & - & - \\
    & CEP \cite{bui2017single} & 11.10 & 0.5490 & 13.19 & 0.5430 & - & - \\
    
    \midrule
    
    \multirow{15}{*}{DL-based} & GFN \cite{ren2018gated} & 15.84 & 0.7510 & 18.16 & 0.6710 & \textbf{0.50M} & 60G \\
    & GCANet \cite{chen2019gated} & 16.50 & 0.7598 & 21.86 & 0.7304 & 0.70M & 74G \\
    & GridDehaze \cite{liu2019griddehazenet} & 16.62 & 0.7870 & 18.92 & 0.6720 & 0.95M & 86G \\
    & FFA-Net \cite{qin2020ffa} & 17.20 & 0.7943 & 22.12 & 0.7700 & 4.45M & 1151G \\
    & CycleGAN \cite{zhu2017unpaired} & 17.80 & 0.7500 & 18.92 & 0.5300 & 11.38M & 232G \\
    & Cycle-Dehaze \cite{engin2018cycle} & 18.03 & 0.8000 & 19.92 & 0.6400 & 11.38M & 232G \\
    & RefineDNet \cite{zhao2021refinednet} & 16.91 & 0.7533 & 17.06 & 0.7724 & 65.80M & 151G \\
    & DehazeFormer-S \cite{song2023vision} & 16.67 & 0.7570 & 16.26 & 0.6540 & 1.28M & 52G \\
    & PPD-Net \cite{zhang2018multi} & 22.53 & \textbf{0.8705} & 24.24 & 0.7205 & 31.28M & 204G \\
    & EDN-GTM \cite{tran2024encoder} & \textbf{22.90} & 0.8270 & 23.46 & \textbf{0.8198} & 49.30M & 308G \\
    & UHD \cite{zheng2021ultra} & - & - & 18.43 & 0.8138 & 34.55M & - \\
    & TransWeather \cite{valanarasu2022transweather} & - & - & 18.84 & 0.7133 & 31.00M & 23G \\
    & SGID-PFF \cite{bai2022self} & - & - & 20.96 & 0.7410 & 13.87M & 625G \\
    & Restormer \cite{zamir2022restormer} & - & - & 23.58 & 0.7680 & 26.10M & 564G \\
    & Dehamer \cite{guo2022image} & - & - & \textbf{25.11} & 0.7770 & 132.50M & 241G \\
    
    \midrule
    DL-based & DPTE-Net (ours) & \underline{21.68} & \underline{0.8164} & 22.02 & 0.6901 & 3.10M & \textbf{19G} \\
    \bottomrule
  \end{tabular}}
  \label{tab:results_4datasets}
\end{table*}

\begin{table*}
  \centering
  \caption{Quantitative comparisons on Dense-HAZE \cite{ancuti2019dense} and NH-HAZE \cite{ancuti2020nh}. The \textbf{bold} numbers denote the best results under each category, and \underline{underline} numbers indicate that our results fall in the top three best results, while it is worth noting that most methods under comparison have more than 10$\times$ MACs compared to our model.}
  \resizebox{0.65\textwidth}{!}{
  \begin{tabular}{ccccccrr}
    \toprule
    \multirow{2}{*}{Type} & \multirow{2}{*}{Method} & \multicolumn{2}{c}{Dense-HAZE} & \multicolumn{2}{c}{NH-HAZE} & \multirow{2}{*}{\#Params$\downarrow$} & \multirow{2}{*}{MACs$\downarrow$}\\
    \cmidrule{3-6}
           & & PSNR$\uparrow$ & SSIM$\uparrow$ & PSNR$\uparrow$ & SSIM$\uparrow$ &\\
    \midrule

    \multirow{5}{*}{Prior-based} & CAP \cite{zhu2015fast} & 11.01 & 0.4874 & 12.58 & 0.4231 & - & - \\
    & BCCR \cite{meng2013efficient} & 11.24 & 0.3514 & 12.48 & 0.4233 & - & - \\
    & DCP \cite{he2010single} & 10.06 & 0.3856 & 10.57 & 0.5196 & - & - \\
    & NLID \cite{berman2016non} & 11.62 & 0.3340 & 11.01 & 0.4850 & - & - \\
    & CEP \cite{bui2017single} & 11.60 & 0.3340 & 10.27 & 0.4100 & - & - \\
    
    \midrule
    
    \multirow{15}{*}{DL-based} & GCANet \cite{chen2019gated} & 10.71 & 0.3615 & 14.27 & 0.5839 & \textbf{0.70M} & 74G \\
    & GridDehaze \cite{liu2019griddehazenet} & 13.31 & 0.3681 & 13.80 & 0.5370 & 0.95M & 86G \\
    & FFA-Net \cite{qin2020ffa} & 14.39 & 0.4524 & 19.87 & 0.6915 & 4.45M & 1151G \\
    & AECR-Net \cite{wu2021contrastive} & 15.80 & 0.4660 & 19.88 & 0.7173 & 2.61M & 209G \\
    & KDDN \cite{hong2020distilling} & 14.28 & 0.4074 & 17.39 & 0.5897 & 5.99M & 41G \\
    & PPD-Net \cite{zhang2018multi} & 13.31 & 0.3681 & 18.23 & 0.6200 & 31.28M & 204G \\
    & UHD \cite{zheng2021ultra} & 15.04 & 0.4830 & 18.01 & 0.5435 & 34.55M & - \\
    & PSD \cite{chen2021psd} & 15.95 & 0.4767 & 19.27 & 0.5824 & 39.62M & 365G \\
    & EDN-GTM \cite{tran2024encoder} & 15.43 & 0.5200 & 20.24 & \textbf{0.7178} & 49.30M & 308G \\
    & SGID-PFF \cite{bai2022self} & 12.49 & 0.5170 & - & - & 13.87M & 625G \\
    & Restormer \cite{zamir2022restormer} & 15.78 & 0.5480 & - & - & 26.10M & 564G \\
    & TransWeather \cite{valanarasu2022transweather} & 15.75 & 0.4957 & - & - & 31.00M & 23G \\
    & DRN \cite{li2022single} & 15.73 & 0.4992 & 17.81 & 0.6219 & 12.60M & - \\
    & DehazeFormer-S \cite{song2023vision} & 16.29 & 0.5104 & 20.17 & 0.7312 & 1.28M & 52G \\
    & Dehamer \cite{guo2022image} & \textbf{16.62} & \textbf{0.5600} & \textbf{20.66} & 0.6840 & 132.50M & 241G \\

    \midrule
    DL-based & DPTE-Net (ours) & 15.59 & \underline{0.5248} & \underline{20.18} & 0.5623 & 3.10M & \textbf{19G} \\
    \bottomrule
  \end{tabular}}
  \label{tab:results_4datasets2}
\end{table*}

\begin{figure*}[t]
    \centering
    \includegraphics[width=1.0\linewidth]{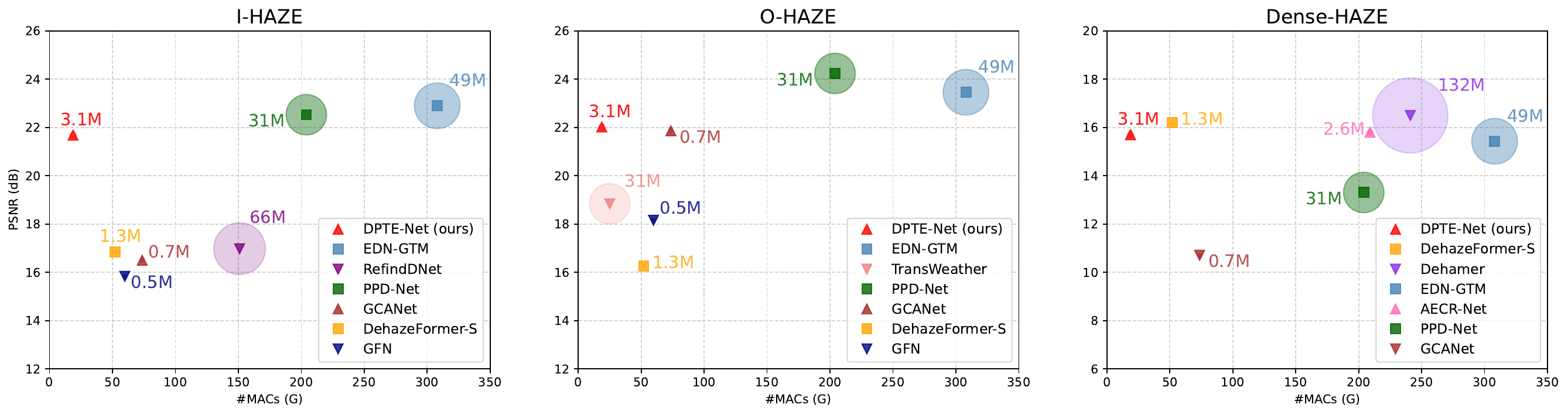}
    \caption{Comparisons of various methods in terms of the trade-off between effectiveness (PSNR) and computational cost (MACs) (the radius of each circle represents the model size (\#Params)).}
    \label{fig:propoassed-frameworkas}
\end{figure*}

\subsection{Loss Functions}
\label{subsec:lossfunctions}

In GAN frameworks, using an integral loss function is often more effective than relying on a single loss function, especially for complex tasks like dehazing \cite{tran2022anovel,li2022single}. Hence, we have adopted an integral loss that combines multiple components to balance the visual realism and structural fidelity of restored images.

{\bf Adversarial Loss.} The adversarial loss $\mathcal{L}_{Adv}$ is defined as follows \cite{kupyn2018deblurgan}: 
\begin{equation}
    \mathcal{L}_{Adv}(x) = -C(S(x)),
\end{equation} 
where $C(.)$ and $S(.)$ denote the critic and student models, respectively, while $x$ represents the hazy input.

{\bf Perceptual Loss.} The perceptual loss $\mathcal{L}_{Per}$ to optimize the perceptual similarity in feature space is defined as \cite{kupyn2018deblurgan}:
\begin{equation}
\mathcal{L}_{Per}(x,y) = || \phi(S(x)) - \phi(y) ||,
\end{equation} 
where $y$ is the clean image and $\phi(.)$ denotes the feature map at layer $Conv3-3$ of the VGG16 model \cite{kupyn2018deblurgan}.

{\bf Hint Loss.} The hint loss $\mathcal{L}_{Hint}$ to distill the knowledge from the teacher is defined as: 
\begin{equation}
\mathcal{L}_{Hint}(x) = || \theta(x) - \psi(x) ||,
\end{equation} 
where $\theta(.)$ and $\psi(.)$ denote the outputs of the adaptation and hint feature maps, respectively.

{\bf Transmission-aware Loss.} As illustrated in Fig. \ref{fig:trans-aware-loss}, the transmission-aware loss $\mathcal{L}_{Trans}$ leverages the transmission estimated by DCP \cite{he2010single} as weight to compute loss and is defined as:
\begin{equation}
    \mathcal{L}_{Trans}(x,y) = || S(x)*\tau - y*\tau ||,
\end{equation} 
where $\tau$ is the transmission weight and ($*$) denotes spatial-wise multiplication.

Ultimately, the integral loss function $\mathcal{L}_I$ used for training the proposed network is formulated by combining all the related loss functions:
\begin{equation}
\begin{split}
\mathcal{L}_{I}(x,y) & = \omega_A \mathcal{L}_{Adv}(x) + \omega_P \mathcal{L}_{Per}(x,y) + \\
      & \quad\quad\quad\quad \omega_T \mathcal{L}_{Trans}(x,y) + \lambda\omega_H \mathcal{L}_{Hint}(x),
\end{split}
\end{equation} 
while the critic loss function $\mathcal{L}_C$ for optimizing the critic model is defined as:
\begin{equation}
\mathcal{L}_{C}(x,y) = \omega_C [C(y) - C(S(x))],
\end{equation}
where $\omega_A$, $\omega_P$, $\omega_T$, $\omega_H$, $\omega_{C}$ are balancing weights. In the primary settings of our present framework, we set $\omega_A$ = 100, $\omega_P$ = 100, $\omega_H$ = 100, $\omega_T$ = 100, $\omega_C$ = 1, and $\delta$ = 0.5. 
In the integral loss $\mathcal{L}_I$, each component specifically targets a particular aspect of the dehazing process, thus enhancing the model's adaptability across diverse scenarios. Moreover, the weights of these loss components can be adjusted during training, allowing for emphasis on certain aspects as required. This flexibility promotes a more balanced learning process, ultimately increasing the robustness and applicability of the framework in practical scenarios.

\begin{figure*}
  \centering
  
  \resizebox{0.998\textwidth}{!}{
  \begin{subfigure}{0.171\linewidth}
    \centering
    {\small PSNR/SSIM}
    \includegraphics[width=1.0\linewidth]{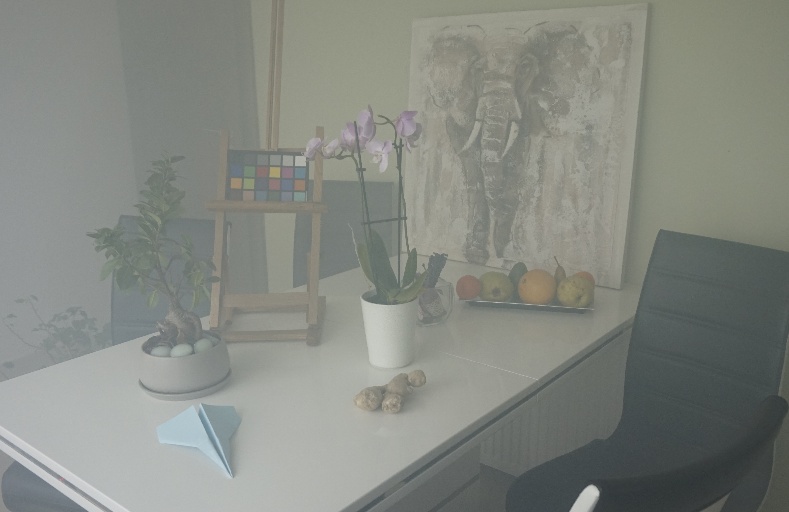} \\
    \captionsetup{font={small}}
    \caption{Hazy}
    \label{fig:ihaze-result-a}
  \end{subfigure}
  \hfill
  \begin{subfigure}{0.171\linewidth}
    \centering
    {\small 10.66/0.5745}
    \includegraphics[width=1.0\linewidth]{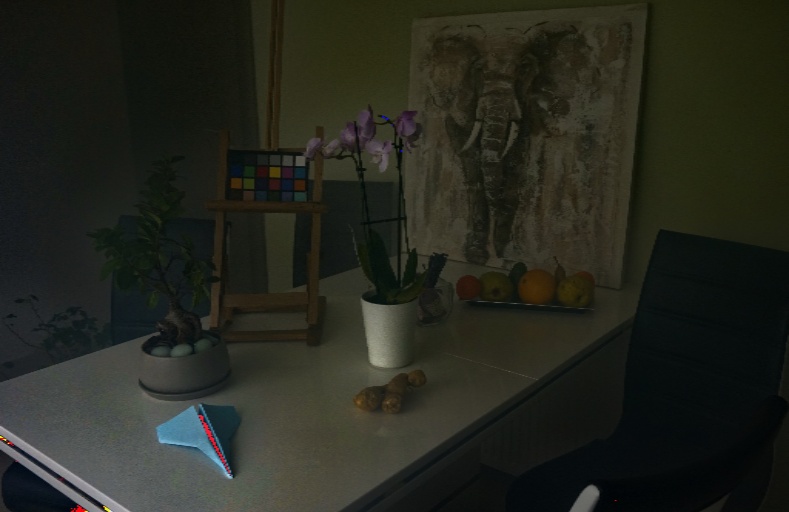} \\
    \captionsetup{font={small}}
    \caption{DCP \cite{he2010single}}
    \label{fig:ihaze-result-b}
  \end{subfigure}
  \hfill
  \begin{subfigure}{0.171\linewidth}
    \centering
    {\small 11.33/0.6035}
    \includegraphics[width=1.0\linewidth]{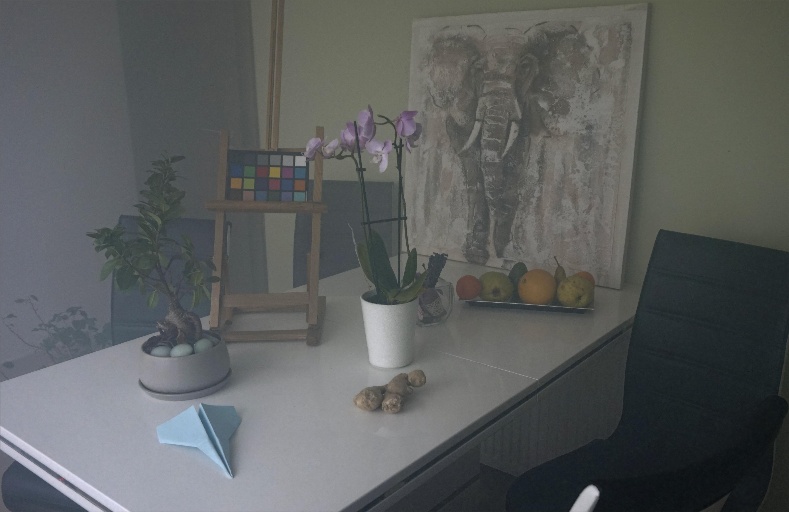} \\
    \captionsetup{font={small}}
    \caption{CAP \cite{zhu2015fast}}
    \label{fig:ihaze-result-c}
  \end{subfigure}
  \hfill
  \begin{subfigure}{0.171\linewidth}
    \centering
    {\small 14.81/0.6416}
    \includegraphics[width=1.0\linewidth]{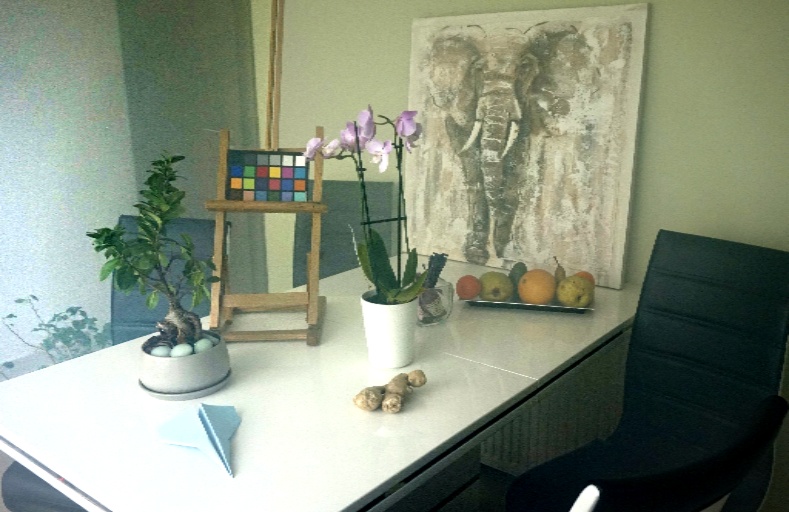} \\
    \captionsetup{font={small}}
    \caption{NLID \cite{berman2016non}}
    \label{fig:ihaze-result-d}
  \end{subfigure}
  \hfill
  \begin{subfigure}{0.171\linewidth}
    \centering
    {\small 13.58/0.6224}
    \includegraphics[width=1.0\linewidth]{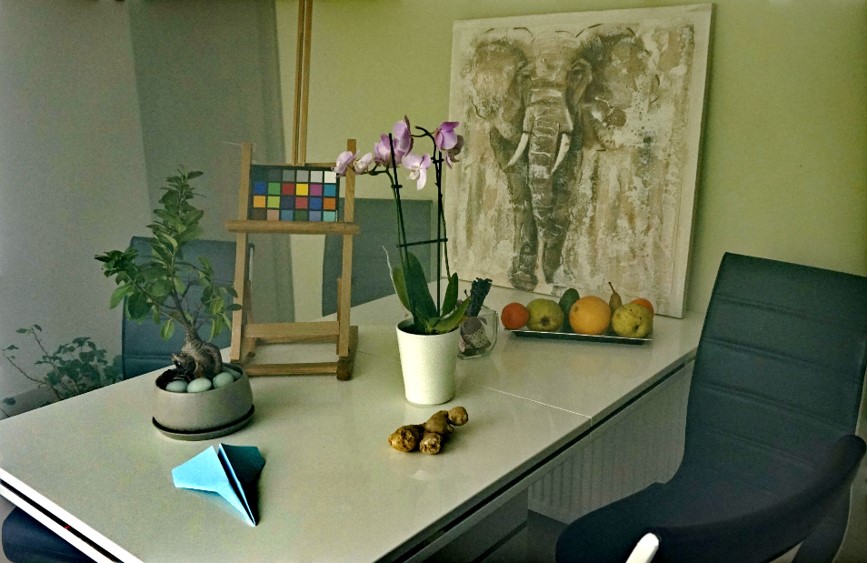} \\
    \captionsetup{font={small}}
    \caption{BCCR \cite{meng2013efficient}}
    \label{fig:ihaze-result-e}
  \end{subfigure}
  \hfill
  \begin{subfigure}{0.171\linewidth}
    \centering
    {\small 10.86/0.5585}
    \includegraphics[width=1.0\linewidth]{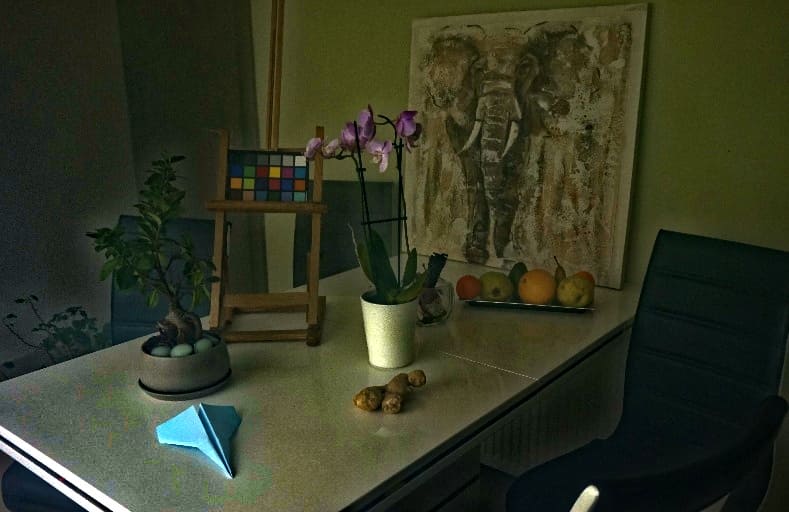} \\
    \captionsetup{font={small}}
    \caption{CEP \cite{bui2017single}}
    \label{fig:ihaze-result-f}
  \end{subfigure}}

  \vspace{0.1cm}

  \resizebox{0.998\textwidth}{!}{
  \begin{subfigure}{0.171\linewidth}
    \centering
    {\small 15.78/0.7502}
    \includegraphics[width=1.0\linewidth]{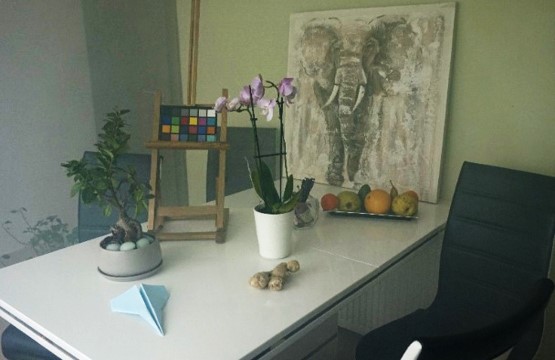} \\
    \captionsetup{font={small}}
    \caption{GridDehaze \cite{liu2019griddehazenet}}
    \label{fig:ihaze-result-h}
  \end{subfigure}
  \hfill
  \begin{subfigure}{0.171\linewidth}
    \centering
    {\small 15.64/0.7442}
    \includegraphics[width=1.0\linewidth]{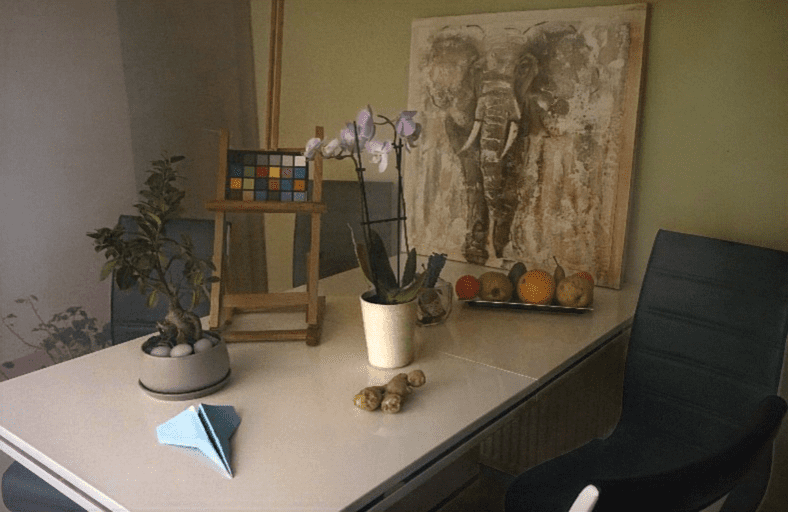} \\
    \captionsetup{font={small}}
    \caption{GCANet \cite{chen2019gated}}
    \label{fig:ihaze-result-g}
  \end{subfigure}
  \hfill
  \begin{subfigure}{0.171\linewidth}
    \centering
    {\small 19.82/0.8049}
    \includegraphics[width=1.0\linewidth]{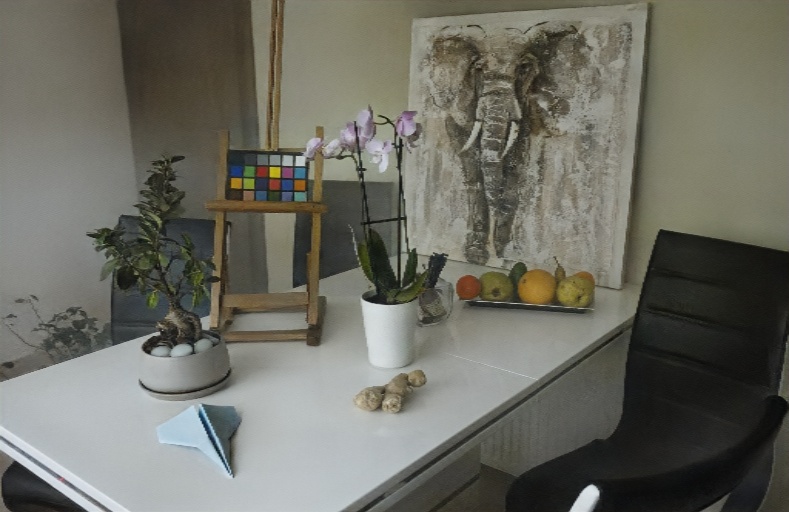} \\
    \captionsetup{font={small}}
    \caption{EDN-GTM \cite{tran2024encoder}}
    \label{fig:ihaze-result-i}
  \end{subfigure}
  \hfill
  \begin{subfigure}{0.171\linewidth}
    \centering
    {\small 19.56/0.8231}
    \includegraphics[width=1.0\linewidth]{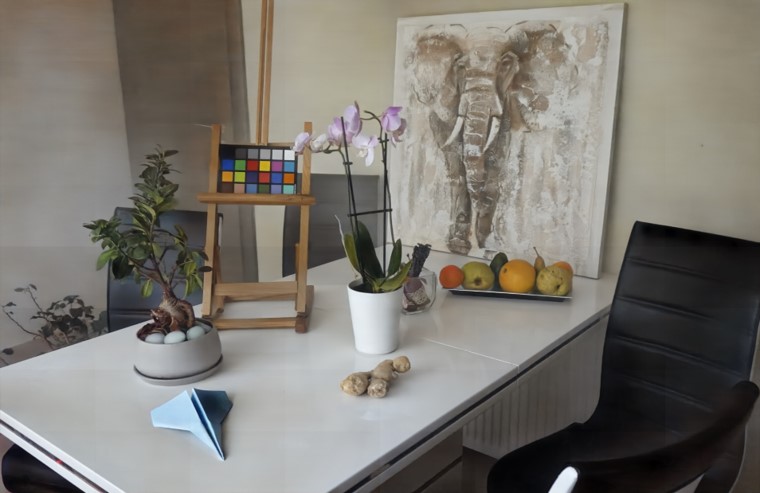} \\
    \captionsetup{font={small}}
    \caption{PDD-Net \cite{zhang2018multi}}
    \label{fig:ihaze-result-j}
  \end{subfigure}
  \hfill
  \begin{subfigure}{0.171\linewidth}
    \centering
    {\small 19.18/0.7917}
    \includegraphics[width=1.0\linewidth]{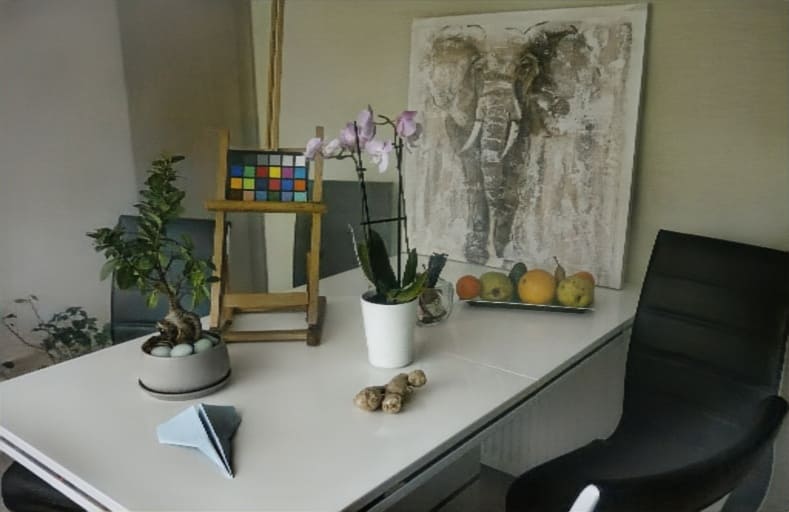} \\
    \captionsetup{font={small}}
    \caption{DPTE-Net (ours)}
    \label{fig:ihaze-result-k}
  \end{subfigure}
  \hfill
  \begin{subfigure}{0.171\linewidth}
    \centering
    {\small $\infty$/1.0}
    \includegraphics[width=1.0\linewidth]{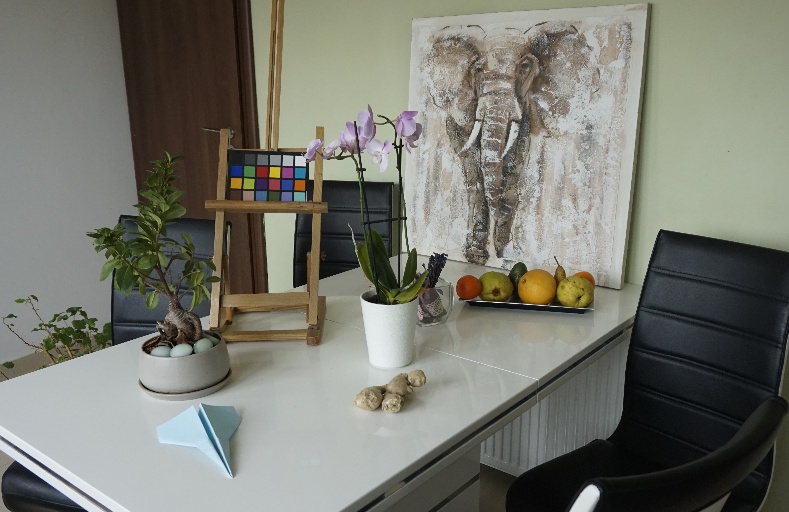} \\
    \captionsetup{font={small}}
    \caption{Clean}
    \label{fig:ihaze-result-l}
  \end{subfigure}}
  
  \caption{Representative results of different approaches on I-HAZE image data.}
  \label{fig:ihaze-result}
\end{figure*}

\begin{figure*}
  \centering
  
  \resizebox{0.998\textwidth}{!}{
  \begin{subfigure}{0.171\linewidth}
    \centering
    {\small PSNR/SSIM}
    \includegraphics[width=1.0\linewidth]{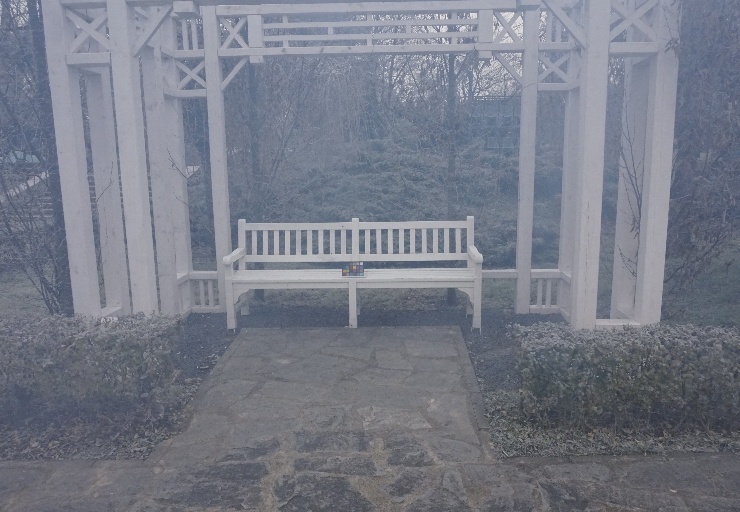} \\ 
    \captionsetup{font={small}}
    \caption{Hazy}
    \label{fig:ohaze-result-a}
  \end{subfigure}
  \hfill
  \begin{subfigure}{0.171\linewidth}
    \centering
    {\small 14.44/0.5580}
    \includegraphics[width=1.0\linewidth]{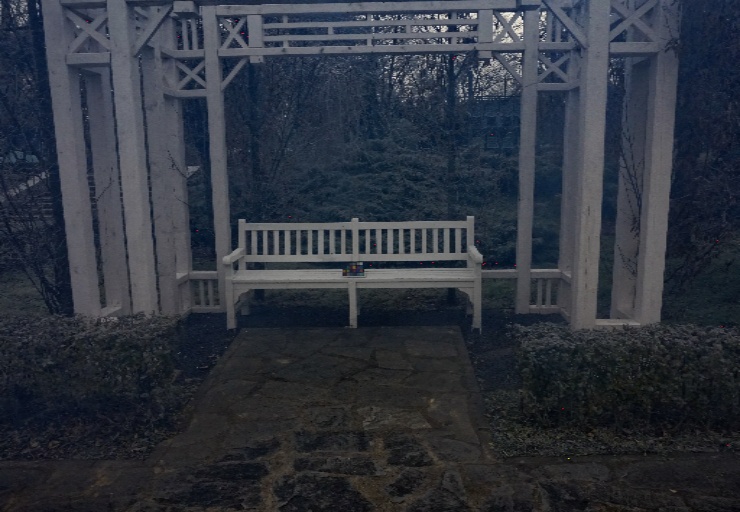} \\
    \captionsetup{font={small}}
    \caption{DCP \cite{he2010single}}
    \label{fig:ohaze-result-b}
  \end{subfigure}
  \hfill
  \begin{subfigure}{0.171\linewidth}
    \centering
    {\small 15.08/0.5682}
    \includegraphics[width=1.0\linewidth]{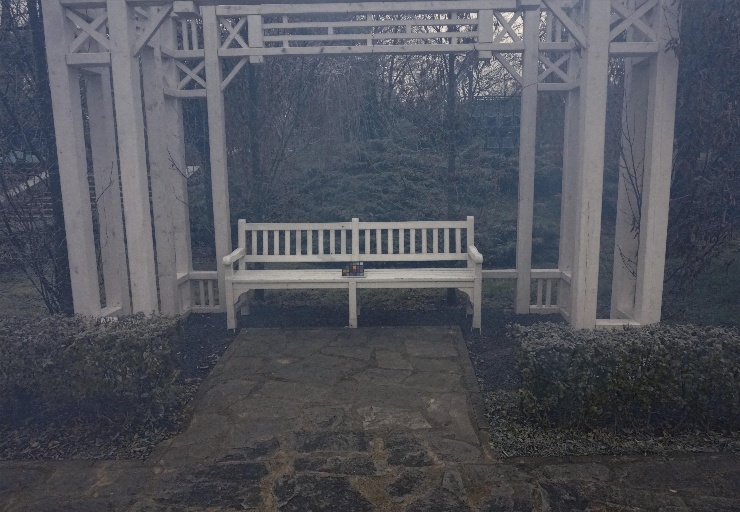} \\
    \captionsetup{font={small}}
    \caption{CAP \cite{zhu2015fast}}
    \label{fig:ohaze-result-c}
  \end{subfigure}
  \hfill
  \begin{subfigure}{0.171\linewidth}
    \centering
    {\small 15.32/0.5444}
    \includegraphics[width=1.0\linewidth]{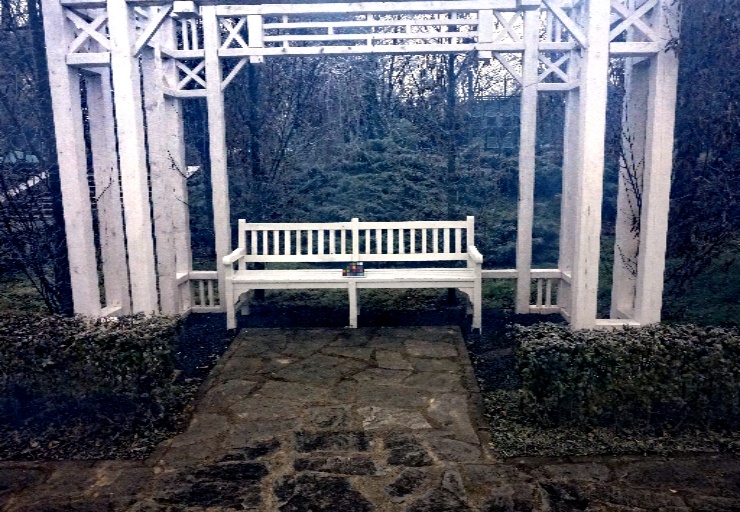} \\
    \captionsetup{font={small}}
    \caption{NLID \cite{berman2016non}}
    \label{fig:ohaze-result-d}
  \end{subfigure}
  \hfill
  \begin{subfigure}{0.171\linewidth}
    \centering
    {\small 13.86/0.5066}
    \includegraphics[width=1.0\linewidth]{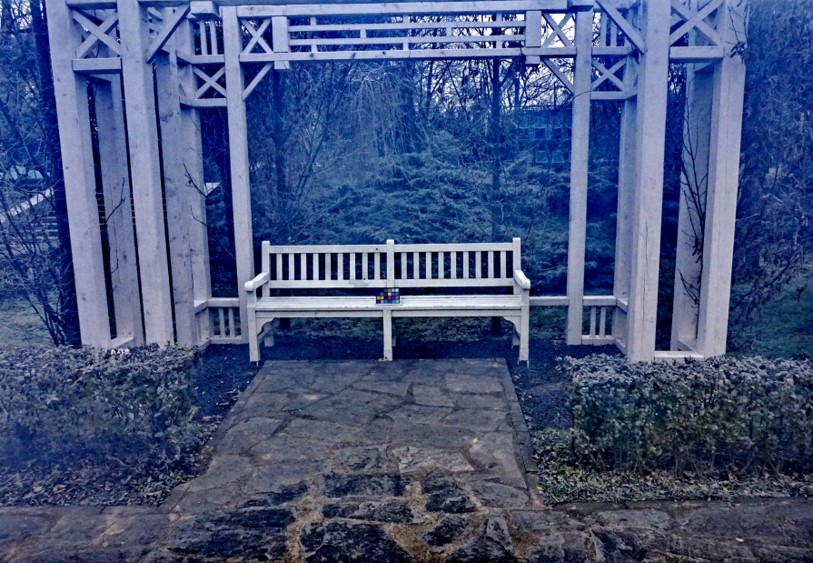} \\
    \captionsetup{font={small}}
    \caption{BCCR \cite{meng2013efficient}}
    \label{fig:ohaze-result-e}
  \end{subfigure}
  \hfill
  \begin{subfigure}{0.171\linewidth}
    \centering
    {\small 12.53/0.5099}
    \includegraphics[width=1.0\linewidth]{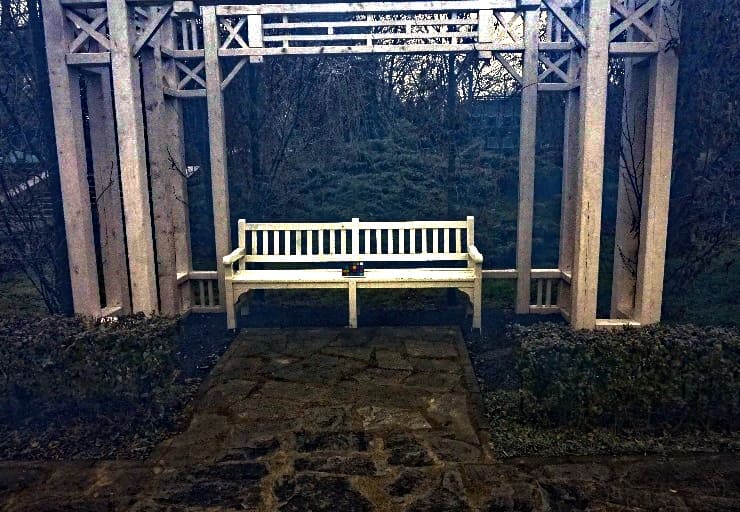} \\
    \captionsetup{font={small}}
    \caption{CEP \cite{bui2017single}}
    \label{fig:ohaze-result-f}
  \end{subfigure}}

  \vspace{0.1cm}

  \resizebox{0.998\textwidth}{!}{
  \begin{subfigure}{0.171\linewidth}
    \centering
    {\small 18.84/0.6085}
    \includegraphics[width=1.0\linewidth]{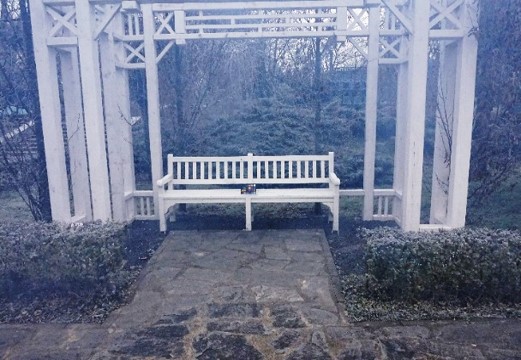} \\
    \captionsetup{font={small}}
    \caption{GridDehaze \cite{liu2019griddehazenet}}
    \label{fig:ohaze-result-h}
  \end{subfigure}
  \hfill
  \begin{subfigure}{0.171\linewidth}
    \centering
    {\small 17.68/0.6056}
    \includegraphics[width=1.0\linewidth]{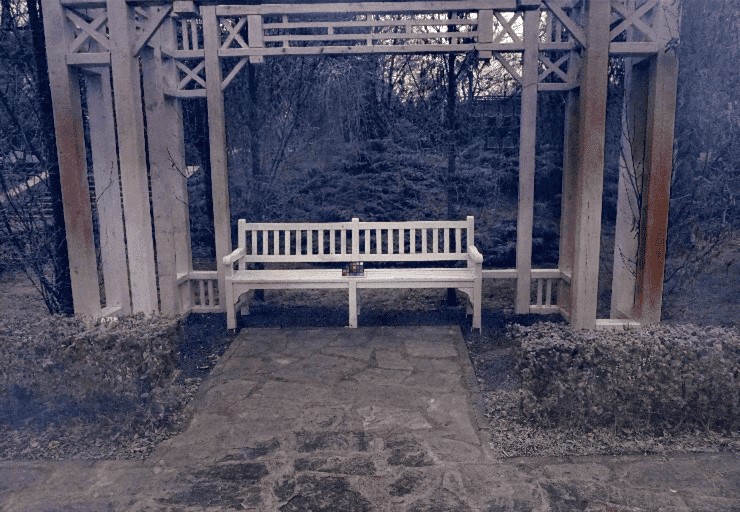} \\
    \captionsetup{font={small}}
    \caption{GCANet \cite{chen2019gated}}
    \label{fig:ohaze-result-g}
  \end{subfigure}
  \hfill
  \begin{subfigure}{0.171\linewidth}
    \centering
    {\small 24.24/0.6398}
    \includegraphics[width=1.0\linewidth]{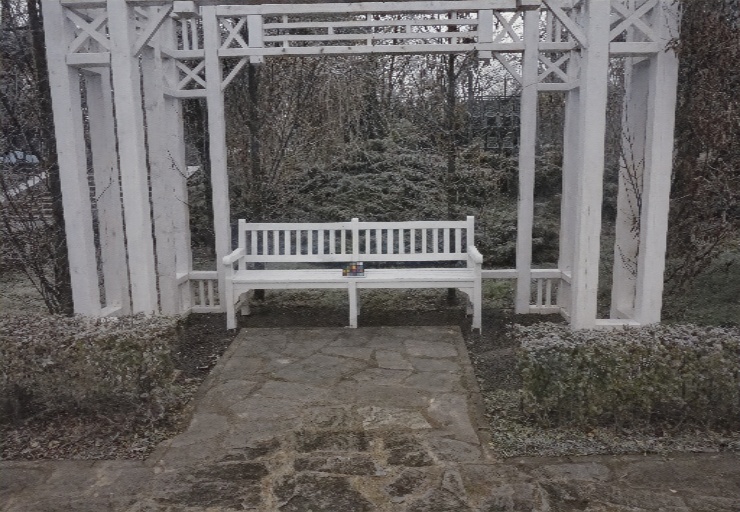} \\
    \captionsetup{font={small}}
    \caption{EDN-GTM \cite{tran2024encoder}}
    \label{fig:ohaze-result-i}
  \end{subfigure}
  \hfill
  \begin{subfigure}{0.171\linewidth}
    \centering
    {\small 24.32/0.6985}
    \includegraphics[width=1.0\linewidth]{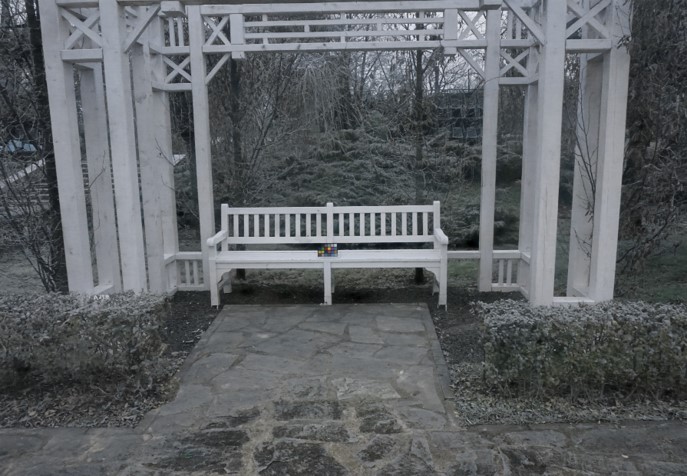} \\
    \captionsetup{font={small}}
    \caption{PDD-Net \cite{zhang2018multi}}
    \label{fig:ohaze-result-j}
  \end{subfigure}
  \hfill
  \begin{subfigure}{0.171\linewidth}
    \centering
    {\small 24.44/0.6524}
    \includegraphics[width=1.0\linewidth]{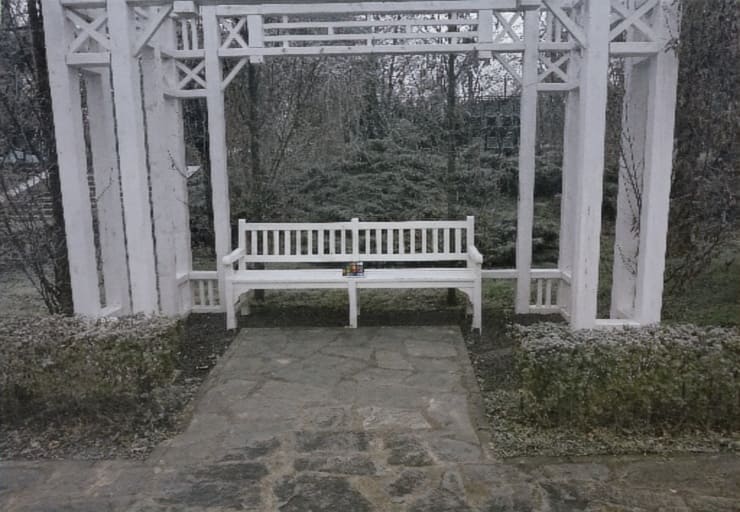} \\
    \captionsetup{font={small}}
    \caption{DPTE-Net (ours)}
    \label{fig:ohaze-result-k}
  \end{subfigure}
  \hfill
  \begin{subfigure}{0.171\linewidth}
    \centering
    {\small $\infty$/1.0}
    \includegraphics[width=1.0\linewidth]{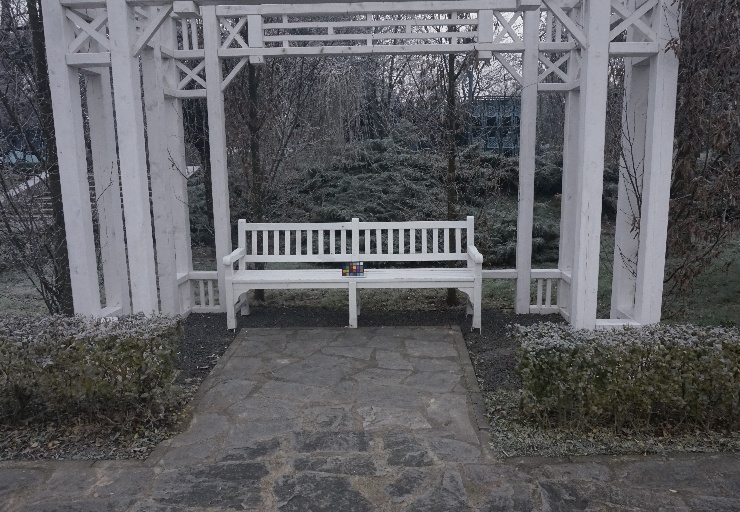} \\
    \captionsetup{font={small}}
    \caption{Clean}
    \label{fig:ohaze-result-l}
  \end{subfigure}}
  
  \caption{Representative results of different approaches on O-HAZE image data.}
  \label{fig:ohaze-result}
\end{figure*}

\section{Experiments and Results}
\label{sec:experiments}

\subsection{Experimental Setup}
\label{subsec:experimentsetup}

\subsubsection{Datasets and Competitors}

Given the focus of this research on real-world hazy scenes, four benchmark datasets including I-HAZE \cite{ancuti2018ihaze}, O-HAZE \cite{ancuti2018haze}, Dense-HAZE \cite{ancuti2019dense}, and NH-HAZE \cite{ancuti2020nh} were chosen to assess the proposed DPTE-Net's performance as these datasets are recognized as commonly used benchmarks in various realistic image dehazing studies. In addition, these datasets represent distinct categories of hazy scenes. Specifically, the I-HAZE and O-HAZE training datasets comprise 25 and 35 pairs of hazy-clean indoor and outdoor images with homogeneous haze conditions, respectively. On the other hand, the Dense-HAZE and NH-HAZE training datasets contain 50 image pairs each, representing densely hazy scenes and non-homogeneous haze distributions, respectively. Each of these four datasets includes 5 image pairs for validation.
This comprehensive range of hazy image data enables us to assess the proposed network's performance across different scenarios, thereby validating its robustness. The proposed DPTE-Net has been compared to state-of-the-art approaches which are typically categorized into two types: \textbf{\textit{Prior-based methods}} (DCP \cite{he2010single}, CAP \cite{zhu2015fast}, NLID \cite{berman2016non}, BCCR \cite{meng2013efficient}, CEP \cite{bui2017single}), and \textbf{\textit{Deep learning (DL)-based models}} (GFN \cite{ren2018gated}, GCANet \cite{chen2019gated}, GridDehaze \cite{liu2019griddehazenet}, FFA-Net \cite{qin2020ffa}, AECR-Net \cite{wu2021contrastive}, KDDN \cite{hong2020distilling}, CycleGAN \cite{zhu2017unpaired}, Cycle-Dehaze \cite{engin2018cycle}, PPD-Net \cite{zhang2018multi}, EDN-GTM \cite{tran2024encoder}, SGID-PFF \cite{bai2022self}, Restormer \cite{zamir2022restormer}, Dehamer \cite{guo2022image}, RefineDNet \cite{zhao2021refinednet}, UHD \cite{zheng2021ultra}, PSD \cite{chen2021psd}, DRN \cite{li2022single}, TransWeather \cite{valanarasu2022transweather}, DehazeFormer \cite{song2023vision}).

\subsubsection{Implementation Settings}

The experiments were conducted on a Linux system utilizing an Intel(R) Xeon(R) Gold 6134 @ 3.20GHz CPU and GeForce GTX TITAN X GPUs. The proposed framework was implemented using the TensorFlow library, with training conducted using the Adam optimizer \cite{kingma2014adam} and a batch size of 4. The learning rate was initially set to \(10^{-4}\) and was uniformly decreased to zero by the end of the training process. The networks were trained for a total of 400 epochs, with early stopping implemented to mitigate the risk of overfitting. To enhance the diversity of the training dataset, several data augmentation techniques, such as random cropping and horizontal flipping, were applied. The quantitative performances were evaluated using Peak Signal-to-Noise Ratio (PSNR) and Structural Similarity Index Measure (SSIM), while the computational complexity was assessed based on the number of parameters (\#Params) and multiply-accumulate (MAC) operations measured on an input size of \(512 \times 512\).

\begin{figure*}
  \centering
  
  \resizebox{0.998\textwidth}{!}{
  \begin{subfigure}{0.171\linewidth}
    \centering
    {\small PSNR/SSIM}
    \includegraphics[width=1.0\linewidth]{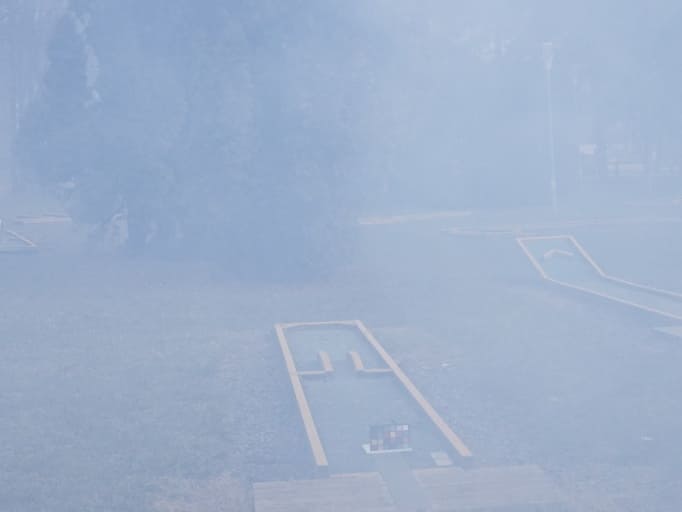} \\ 
    \captionsetup{font={small}}
    \caption{Hazy}
    \label{fig:densehaze-result-a}
  \end{subfigure}
  \hfill
  \begin{subfigure}{0.171\linewidth}
    \centering
    {\small 10.71/0.3234}
    \includegraphics[width=1.0\linewidth]{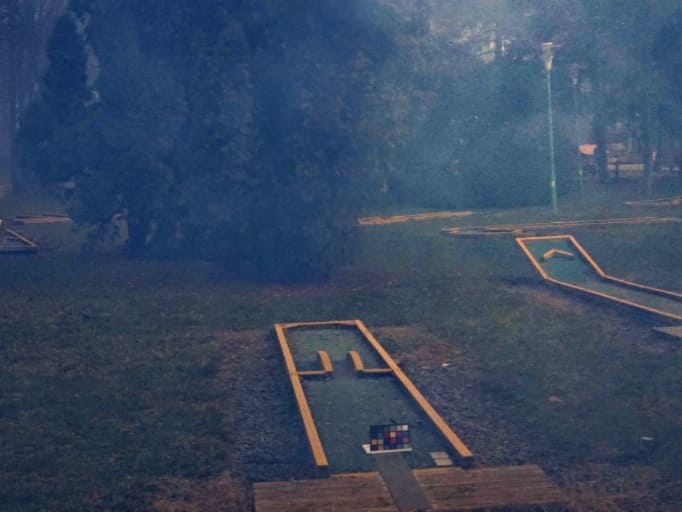} \\
    \captionsetup{font={small}}
    \caption{DCP \cite{he2010single}}
    \label{fig:densehaze-result-b}
  \end{subfigure}
  \hfill
  \begin{subfigure}{0.171\linewidth}
    \centering
    {\small 10.66/0.3328}
    \includegraphics[width=1.0\linewidth]{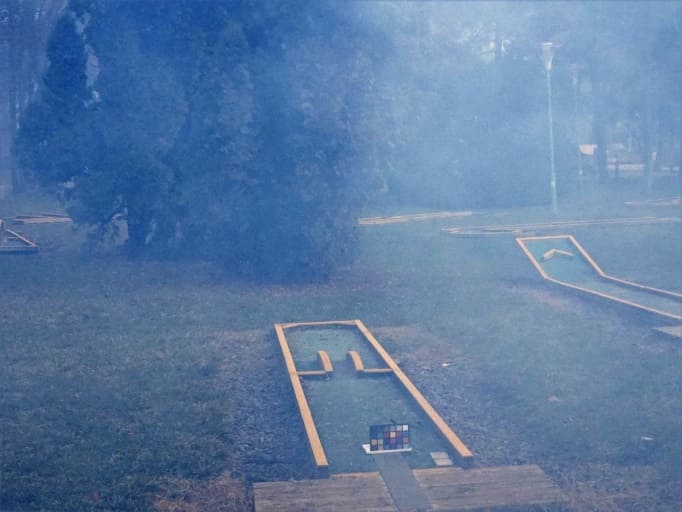} \\
    \captionsetup{font={small}}
    \caption{CAP \cite{zhu2015fast}}
    \label{fig:densehaze-result-c}
  \end{subfigure}
  \hfill
  \begin{subfigure}{0.171\linewidth}
    \centering
    {\small 11.61/0.3542}
    \includegraphics[width=1.0\linewidth]{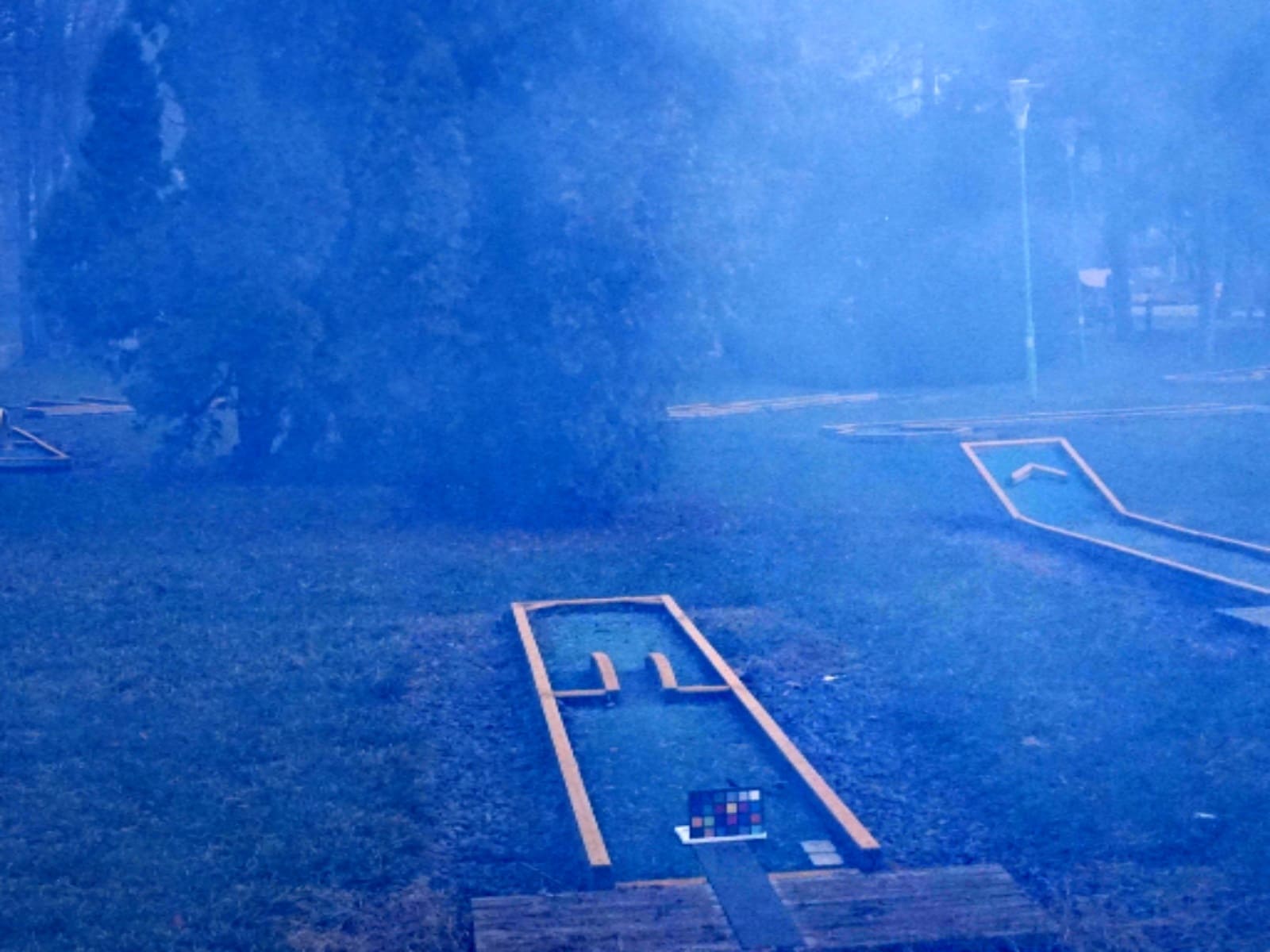} \\
    \captionsetup{font={small}}
    \caption{NLID \cite{berman2016non}}
    \label{fig:densehaze-result-d}
  \end{subfigure}
  \hfill
  \begin{subfigure}{0.171\linewidth}
    \centering
    {\small 11.01/0.3496}
    \includegraphics[width=1.0\linewidth]{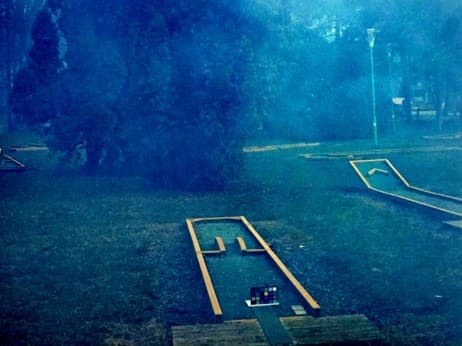} \\
    \captionsetup{font={small}}
    \caption{BCCR \cite{meng2013efficient}}
    \label{fig:densehaze-result-e}
  \end{subfigure}
  \hfill
  \begin{subfigure}{0.171\linewidth}
    \centering
    {\small 10.16/0.3285}
    \includegraphics[width=1.0\linewidth]{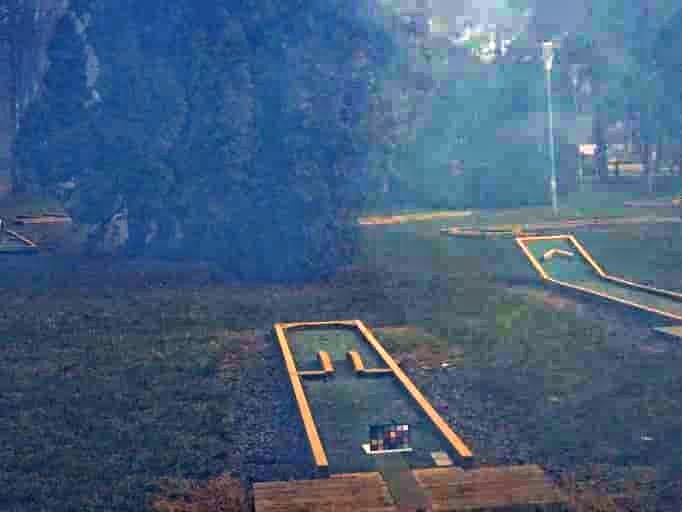} \\
    \captionsetup{font={small}}
    \caption{CEP \cite{bui2017single}}
    \label{fig:densehaze-result-f}
  \end{subfigure}}

  \vspace{0.1cm}

  \resizebox{0.998\textwidth}{!}{
  \begin{subfigure}{0.171\linewidth}
    \centering
    {\small 12.35/0.3649}
    \includegraphics[width=1.0\linewidth]{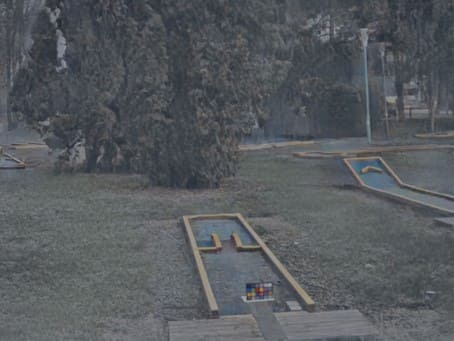} \\
    \captionsetup{font={small}}
    \caption{GridDehaze \cite{liu2019griddehazenet}}
    \label{fig:densehaze-result-h}
  \end{subfigure}
  \hfill
  \begin{subfigure}{0.171\linewidth}
    \centering
    {\small 11.56/0.3575}
    \includegraphics[width=1.0\linewidth]{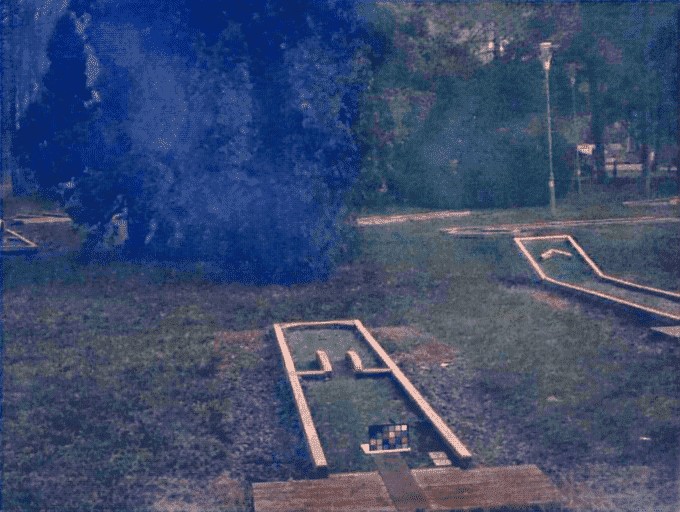} \\
    \captionsetup{font={small}}
    \caption{GCANet \cite{chen2019gated}}
    \label{fig:densehaze-result-g}
  \end{subfigure}
  \hfill
  \begin{subfigure}{0.171\linewidth}
    \centering
    {\small 18.06/0.5681}
    \includegraphics[width=1.0\linewidth]{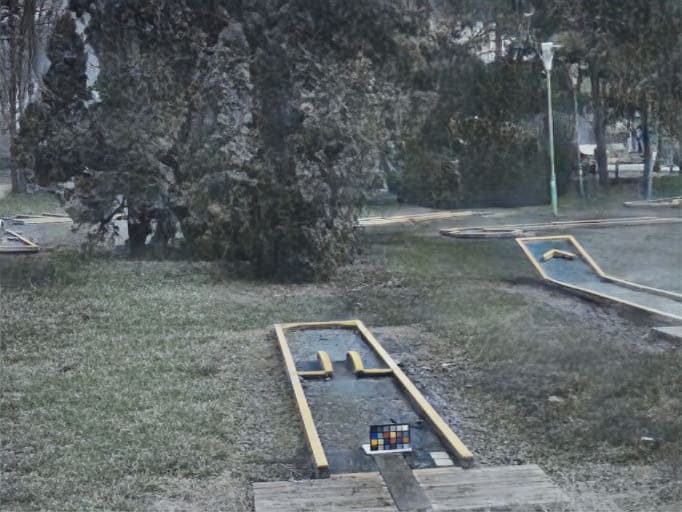} \\
    \captionsetup{font={small}}
    \caption{EDN-GTM \cite{tran2024encoder}}
    \label{fig:densehaze-result-i}
  \end{subfigure}
  \hfill
  \begin{subfigure}{0.171\linewidth}
    \centering
    {\small 21.83/0.6023}
    \includegraphics[width=1.0\linewidth]{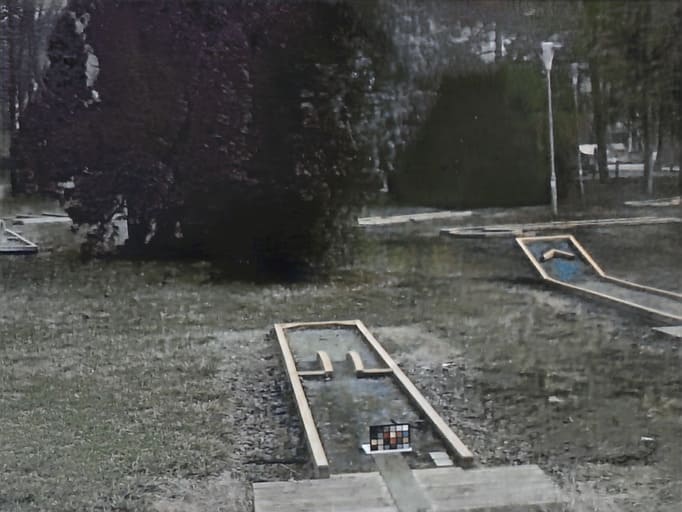} \\
    \captionsetup{font={small}}
    \caption{Dehamer \cite{guo2022image}}
    \label{fig:densehaze-result-j}
  \end{subfigure}
  \hfill
  \begin{subfigure}{0.171\linewidth}
    \centering
    {\small 19.85/0.5355}
    \includegraphics[width=1.0\linewidth]{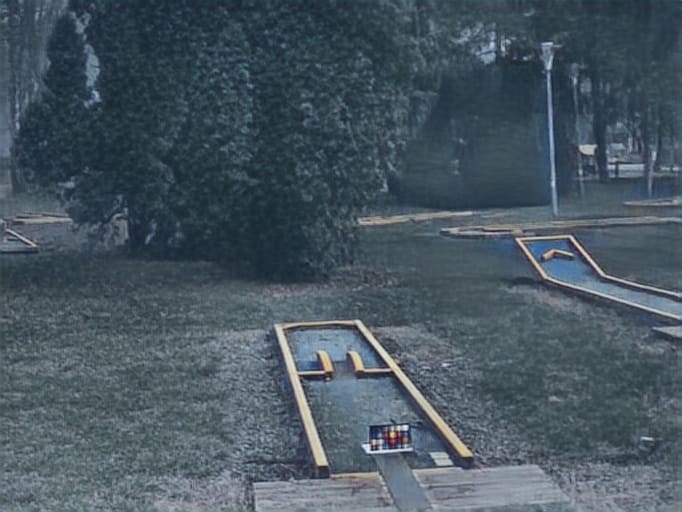} \\
    \captionsetup{font={small}}
    \caption{DPTE-Net (ours)}
    \label{fig:densehaze-result-k}
  \end{subfigure}
  \hfill
  \begin{subfigure}{0.171\linewidth}
    \centering
    {\small $\infty$/1.0}
    \includegraphics[width=1.0\linewidth]{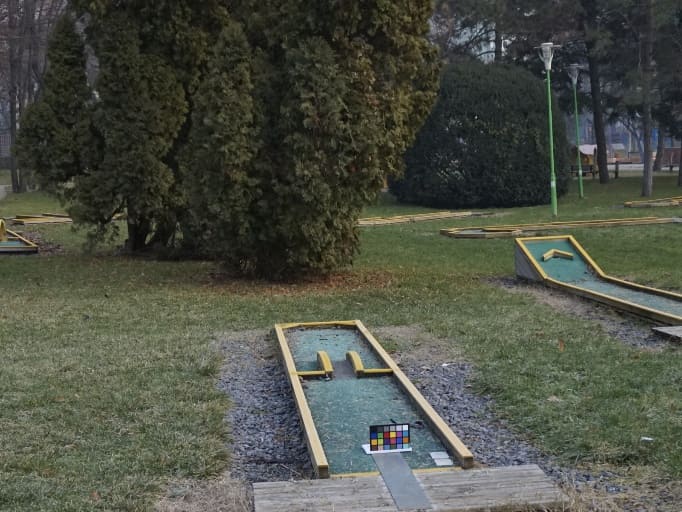} \\
    \captionsetup{font={small}}
    \caption{Clean}
    \label{fig:densehaze-result-l}
  \end{subfigure}}
  
  \caption{Representative results of different approaches on Dense-HAZE image data.}
  \label{fig:densehaze-result}
\end{figure*}

\begin{figure*}
  \centering
  \resizebox{0.998\textwidth}{!}{
  \begin{subfigure}{0.171\linewidth}
    \centering
    {\small PSNR/SSIM}
    \includegraphics[width=1.0\linewidth]{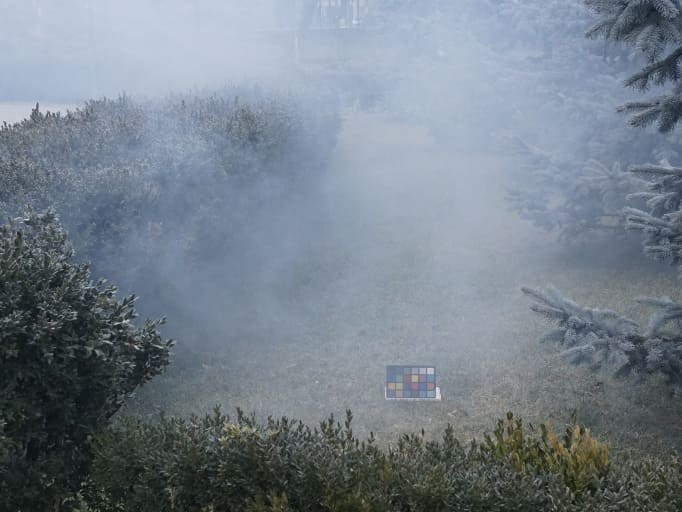} \\ 
    \captionsetup{font={small}}
    \caption{Hazy}
    \label{fig:nhhaze-result-a}
  \end{subfigure}
  \hfill
  \begin{subfigure}{0.171\linewidth}
    \centering
    {\small 11.12/0.3719}
    \includegraphics[width=1.0\linewidth]{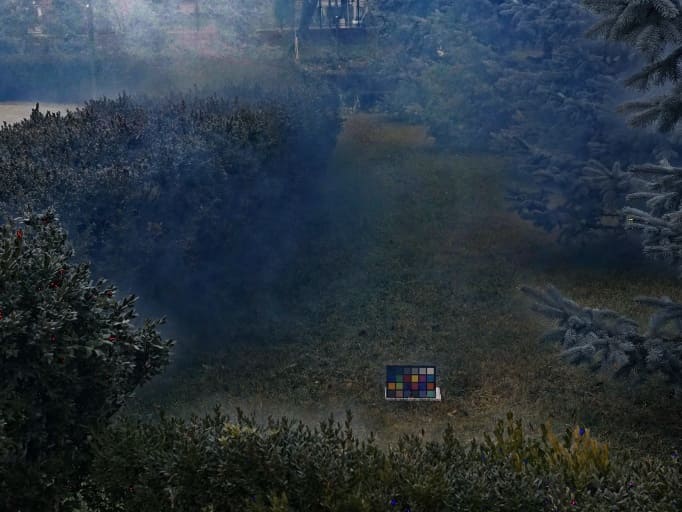} \\
    \captionsetup{font={small}}
    \caption{DCP \cite{he2010single}}
    \label{fig:nhhaze-result-b}
  \end{subfigure}
  \hfill
  \begin{subfigure}{0.171\linewidth}
    \centering
    {\small 13.36/0.4366}
    \includegraphics[width=1.0\linewidth]{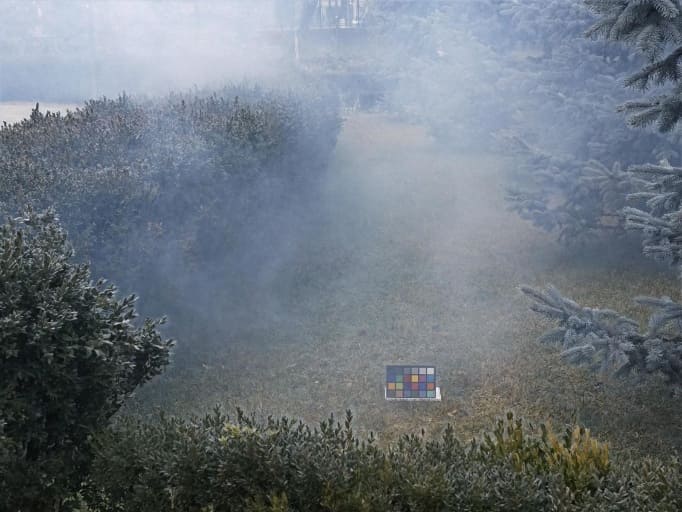} \\
    \captionsetup{font={small}}
    \caption{CAP \cite{zhu2015fast}}
    \label{fig:nhhaze-result-c}
  \end{subfigure}
  \hfill
  \begin{subfigure}{0.171\linewidth}
    \centering
    {\small 10.07/0.3215}
    \includegraphics[width=1.0\linewidth]{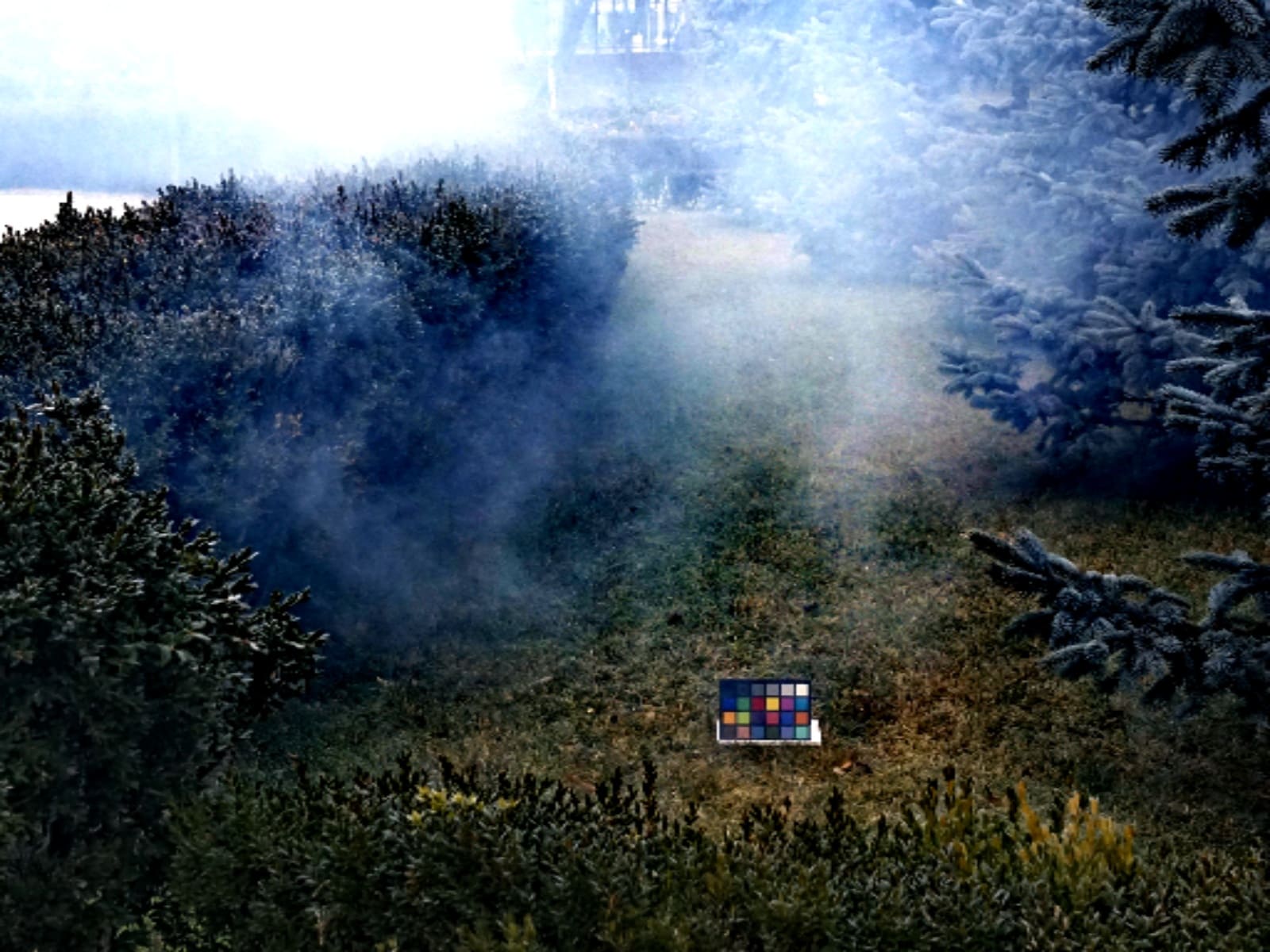} \\
    \captionsetup{font={small}}
    \caption{NLID \cite{berman2016non}}
    \label{fig:nhhaze-result-d}
  \end{subfigure}
  \hfill
  \begin{subfigure}{0.171\linewidth}
    \centering
    {\small 11.56/0.4165}
    \includegraphics[width=1.0\linewidth]{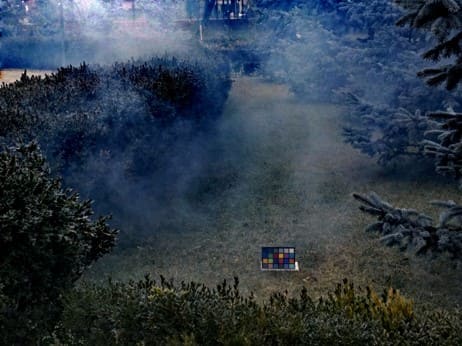} \\
    \captionsetup{font={small}}
    \caption{BCCR \cite{meng2013efficient}}
    \label{fig:nhhaze-result-e}
  \end{subfigure}
  \hfill
  \begin{subfigure}{0.171\linewidth}
    \centering
    {\small 10.59/0.3869}
    \includegraphics[width=1.0\linewidth]{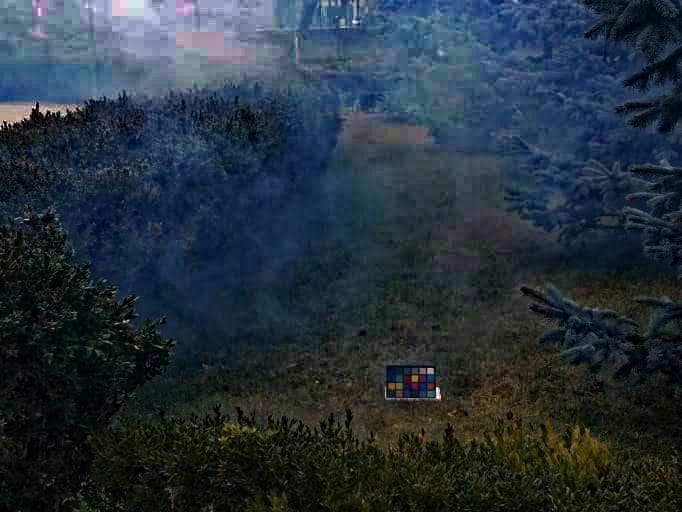} \\
    \captionsetup{font={small}}
    \caption{CEP \cite{bui2017single}}
    \label{fig:nhhaze-result-f}
  \end{subfigure}}

  \vspace{0.1cm}

  \resizebox{0.998\textwidth}{!}{
  \begin{subfigure}{0.171\linewidth}
    \centering
    {\small 12.62/0.4581}
    \includegraphics[width=1.0\linewidth]{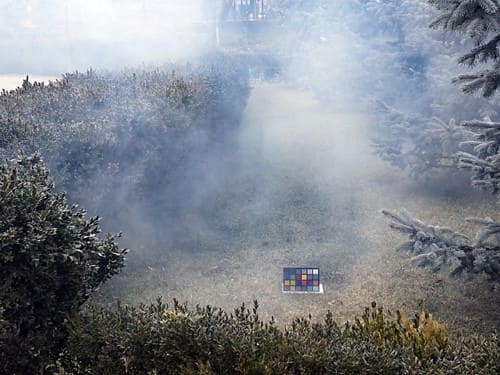} \\
    \captionsetup{font={small}}
    \caption{GridDehaze \cite{liu2019griddehazenet}}
    \label{fig:nhhaze-result-h}
  \end{subfigure}
  \hfill
  \begin{subfigure}{0.171\linewidth}
    \centering
    {\small 12.88/0.4633}
    \includegraphics[width=1.0\linewidth]{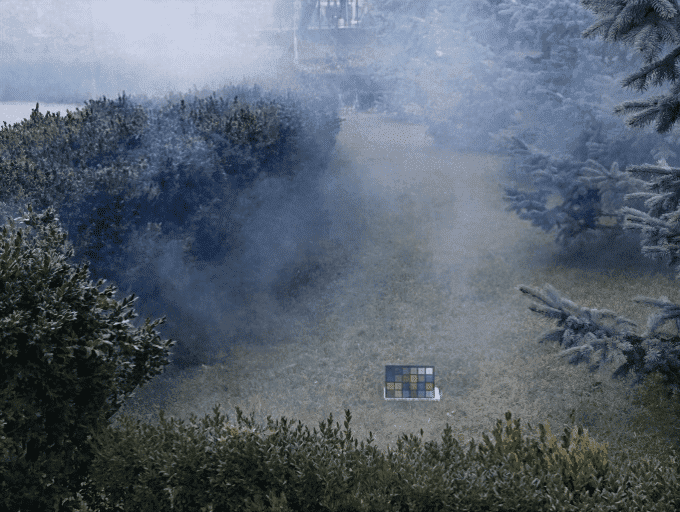} \\
    \captionsetup{font={small}}
    \caption{GCANet \cite{chen2019gated}}
    \label{fig:nhhaze-result-g}
  \end{subfigure}
  \hfill
  \begin{subfigure}{0.171\linewidth}
    \centering
    {\small 18.09/0.6410}
    \includegraphics[width=1.0\linewidth]{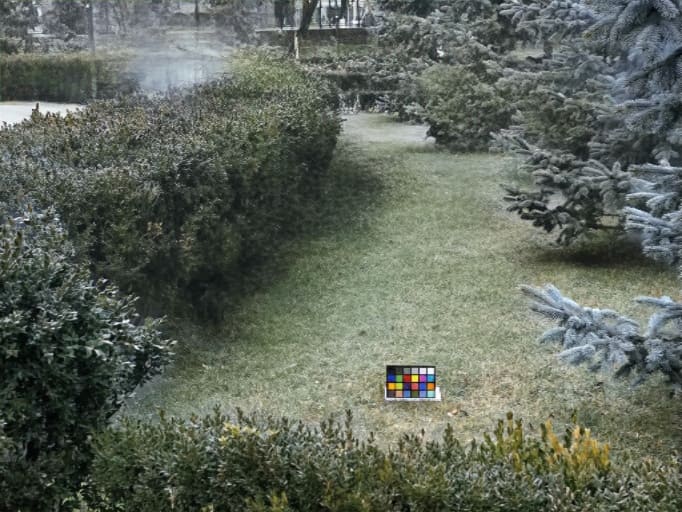} \\
    \captionsetup{font={small}}
    \caption{EDN-GTM \cite{tran2024encoder}}
    \label{fig:nhhaze-result-i}
  \end{subfigure}
  \hfill
  \begin{subfigure}{0.171\linewidth}
    \centering
    {\small 20.14/0.6321}
    \includegraphics[width=1.0\linewidth]{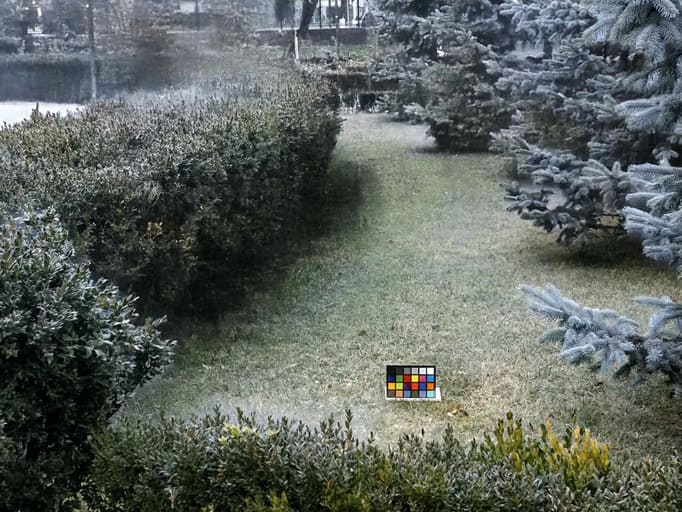} \\
    \captionsetup{font={small}}
    \caption{Dehamer \cite{guo2022image}}
    \label{fig:nhhaze-result-j}
  \end{subfigure}
  \hfill
  \begin{subfigure}{0.171\linewidth}
    \centering
    {\small 19.40/0.5224}
    \includegraphics[width=1.0\linewidth]{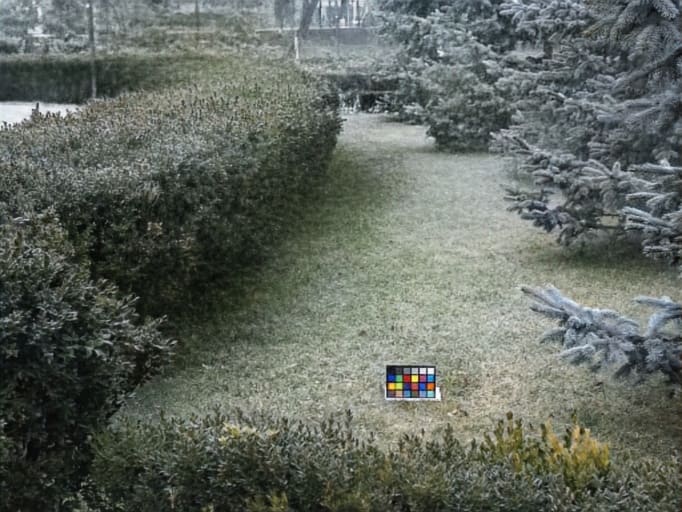} \\
    \captionsetup{font={small}}
    \caption{DPTE-Net (ours)}
    \label{fig:nhhaze-result-k}
  \end{subfigure}
  \hfill
  \begin{subfigure}{0.171\linewidth}
    \centering
    {\small $\infty$/1.0}
    \includegraphics[width=1.0\linewidth]{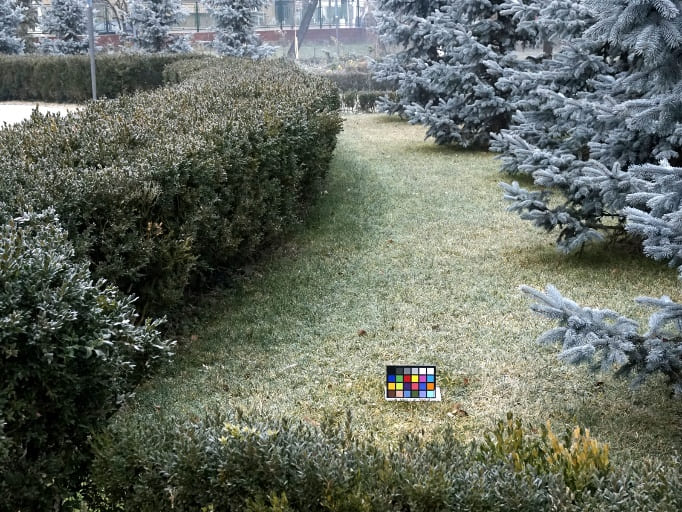} \\
    \captionsetup{font={small}}
    \caption{Clean}
    \label{fig:nhhaze-result-l}
  \end{subfigure}}
  
  \caption{Representative results of different approaches on NH-HAZE image data.}
  \label{fig:nhhaze-result}
\end{figure*}

\subsection{Results and Comparisons}
\label{subsec:resultsandcomparisons}

\subsubsection{Quantitative Evaluation}

Table \ref{tab:results_4datasets} summarizes the quantitative performances of DPTE-Net and various state-of-the-art dehazing methods on the I-HAZE and O-HAZE datasets. In general, DL-based methods considerably outperform prior-based algorithms, and DPTE-Net exhibits the fewest MACs among DL-based approaches under comparison. In terms of effectiveness, EDN-GTM, PPD-Net, and Dehamer produce the top results and outperform the remaining methods. Specifically, EDN-GTM gives the best PSNR on I-HAZE (22.90) and the highest SSIM on O-HAZE (0.8198), whereas Dehamer achieves the best PSNR on O-HAZE (25.11) and PPD-Net produces the highest SSIM on I-HAZE (0.8705). However, these methods rank among the most computationally expensive models within the comparison, with Dehamer particularly utilizing 132.5M parameters. Meanwhile, the proposed DPTE-Net demonstrates satisfactory outcomes, achieving PSNR/SSIM of 21.68/0.8164 on I-HAZE, and 22.02/0.6901 on O-HAZE, while it is worth noting that DPTE-Net has relatively low \#Params (3.1M) and MACs (19G), making it more competitive against other DL-based methods under comparison when considering the trade-off between effectiveness and efficiency.

The quantitative evaluations on the Dense-HAZE and NH-HAZE datasets, on the other hand, are summarized in Table \ref{tab:results_4datasets2}. The best quantitative measures are obtained by Dehamer, with the highest PSNR on both Dense-HAZE (16.62) and NH-HAZE (20.66), and the strongest SSIM on Dense-HAZE (0.5600). Meanwhile, the best SSIM score on NH-HAZE (0.7178) is achieved by EDN-GTM. However, those advantages come at the cost of efficiency. That is, as a standard ViT-based model, Dehamer needs 241G MACs to achieve its superior performance, while EDN-GTM also requires 308G MACs to deliver similarly impressive results. The proposed DPTE-Net, on the other hand, achieves PSNR/SSIM of 15.59/0.5248 on Dense-HAZE, and 20.18/0.5623 on NH-HAZE. These results indicate that DPTE-Net is highly competitive when compared to other DL-based approaches while necessitating only 3.1M parameters and 19G MACs. This implies that DPTE-Net can offer a favorable trade-off between dehazing quality and computational efficiency.

To further highlight the practical potential of DPTE-Net under parameter constraints, graphical summaries comparing various methods in terms of the trade-off between effectiveness (measured by PSNR) and computational cost (measured by MACs) across the four evaluated datasets are illustrated in Fig. \ref{fig01:chart} (for NH-HAZE) and Fig. \ref{fig:propoassed-frameworkas} (for I-HAZE, O-HAZE, and Dense-HAZE). As can be observed from these figures, although DPTE-Net may be outperformed slightly in terms of effectiveness by more computationally intensive models in specific scenarios, it achieves satisfactory dehazing performance on nearly all datasets while requiring fewer computational resources, making it well-suited for scenarios where efficiency is critical, such as applications requiring real-time processing or resource-constrained environments.

\begin{figure*}
  \centering  
  \resizebox{0.98\textwidth}{!}{
  \begin{subfigure}{0.171\linewidth}
    \centering 
    \includegraphics[width=1.0\linewidth]{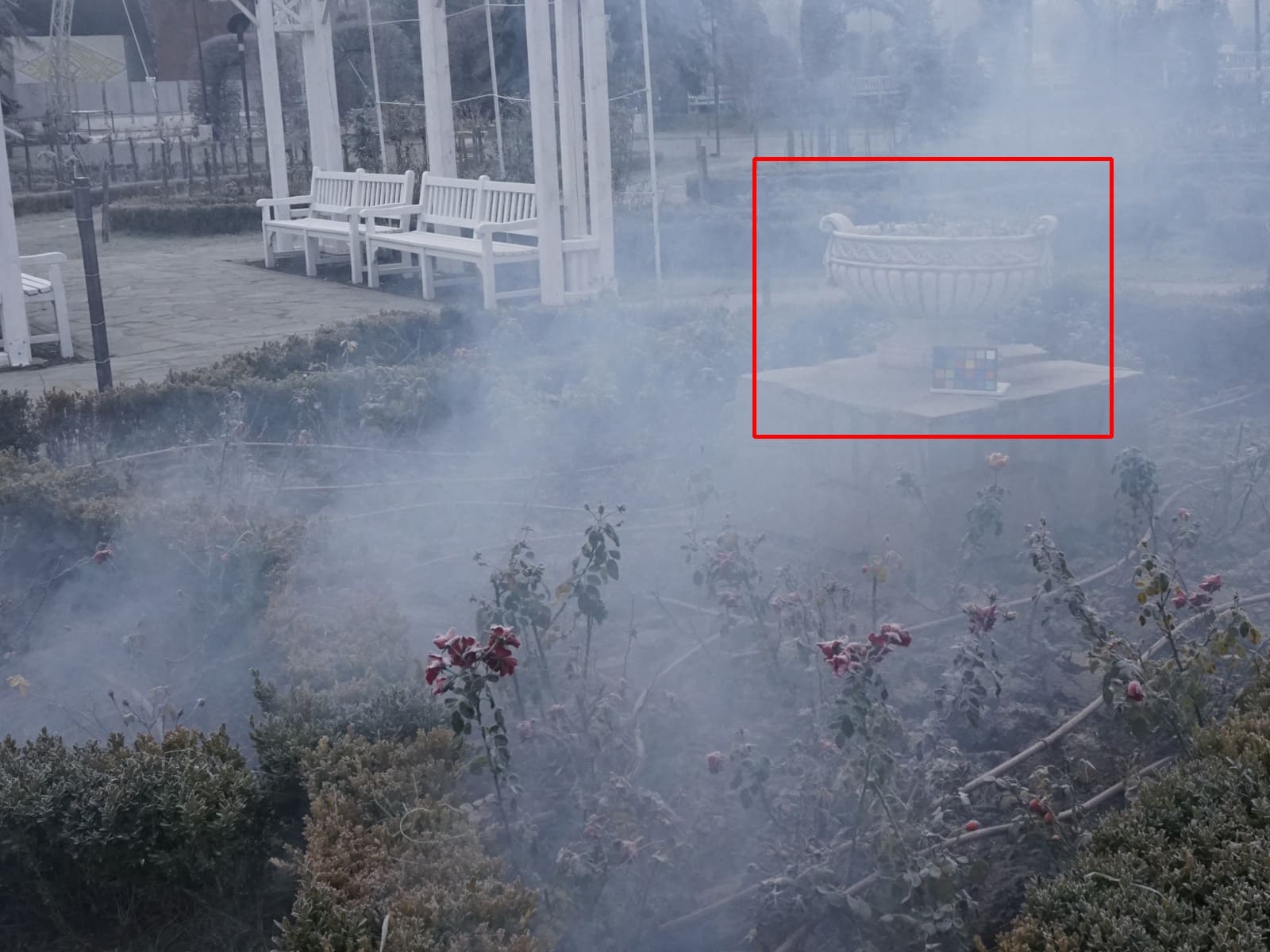}\\
    \vspace{0.05cm}
    \includegraphics[width=1.0\linewidth]{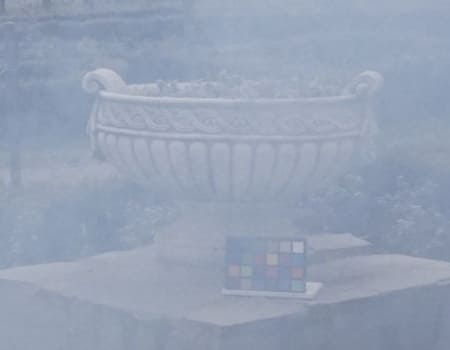}
    \captionsetup{font={small}}
    \caption{Hazy}
    \label{fig:ablation-a}
  \end{subfigure}
  \hfill
  \begin{subfigure}{0.171\linewidth}
    \centering
    \includegraphics[width=1.0\linewidth]{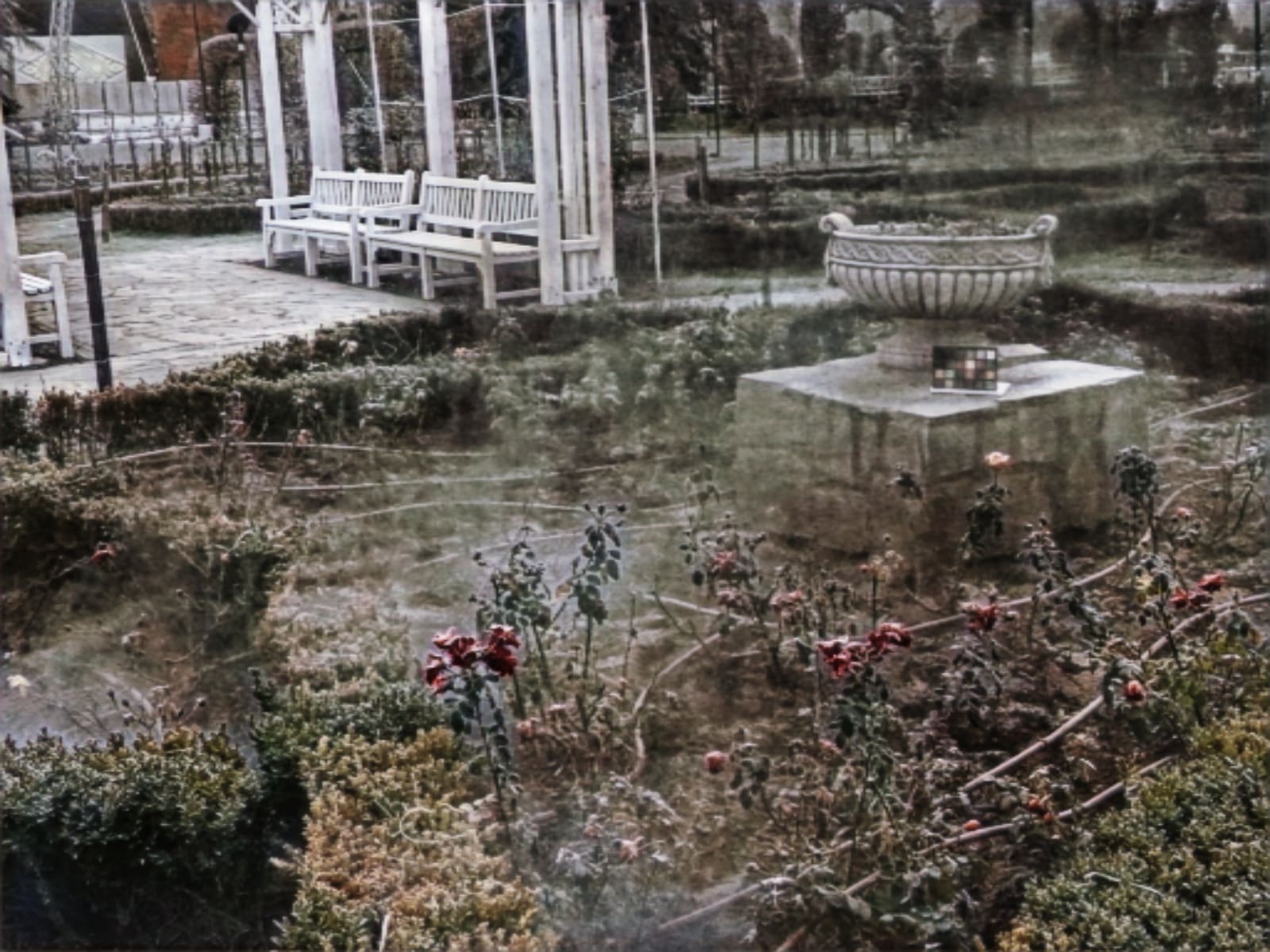}\\
    \vspace{0.05cm}
    \includegraphics[width=1.0\linewidth]{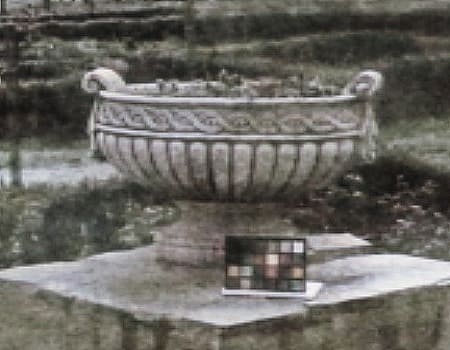}
    \captionsetup{font={small}}
    \caption{Baseline}
    \label{fig:ablation-b}
  \end{subfigure}
  \hfill
  \begin{subfigure}{0.171\linewidth}
    \centering
    \includegraphics[width=1.0\linewidth]{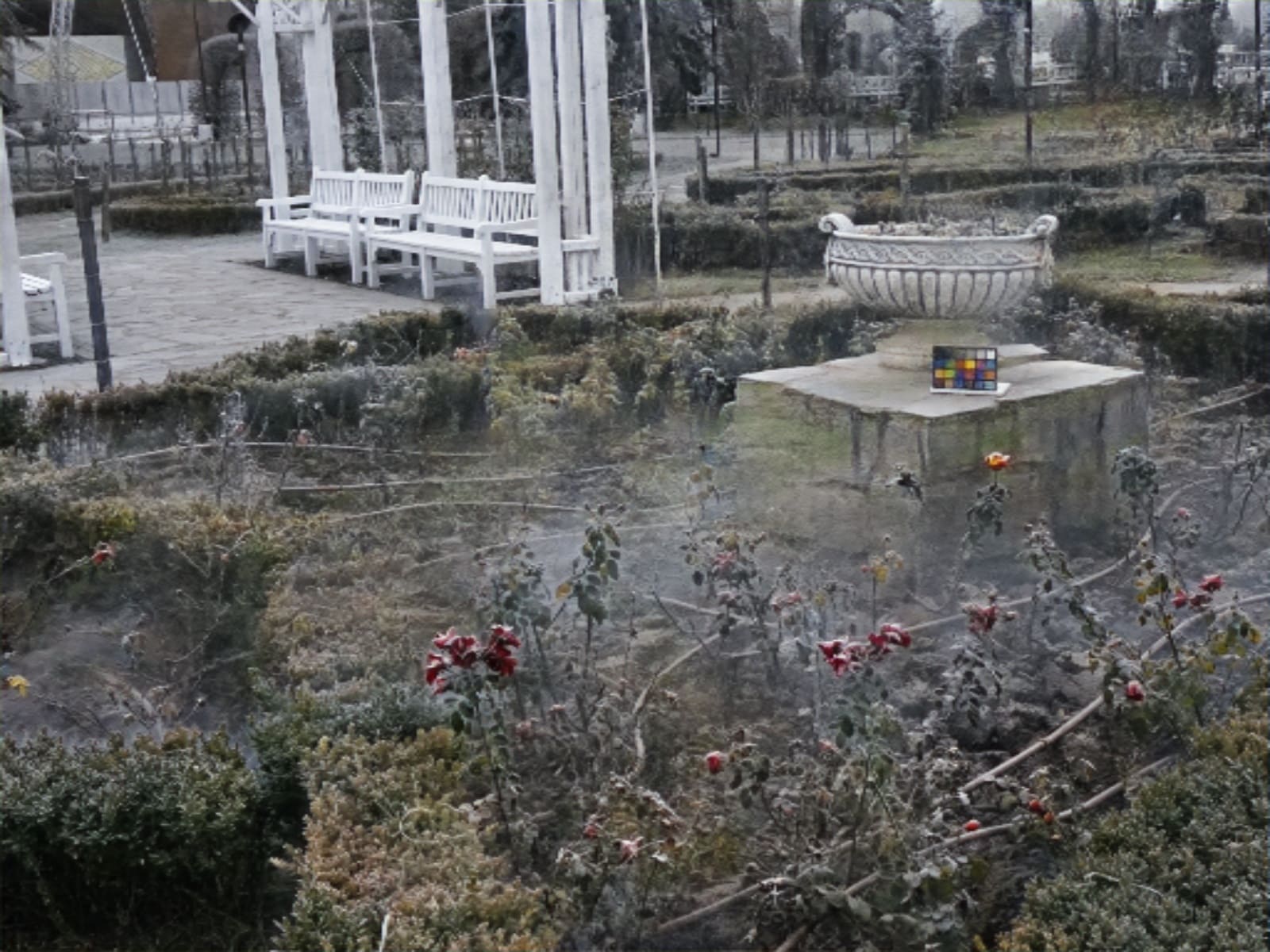}\\
    \vspace{0.05cm}
    \includegraphics[width=1.0\linewidth]{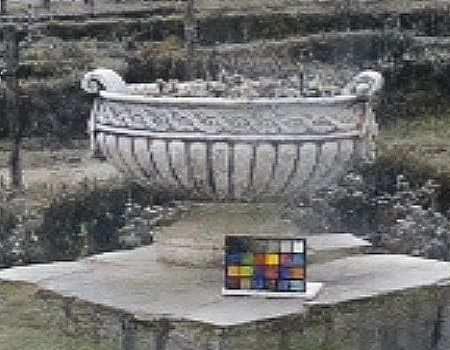}
    \captionsetup{font={small}}
    \caption{+\textbf{P}}
    \label{fig:ablation-c}
  \end{subfigure}
  \hfill
  \begin{subfigure}{0.171\linewidth}
    \centering
    \includegraphics[width=1.0\linewidth]{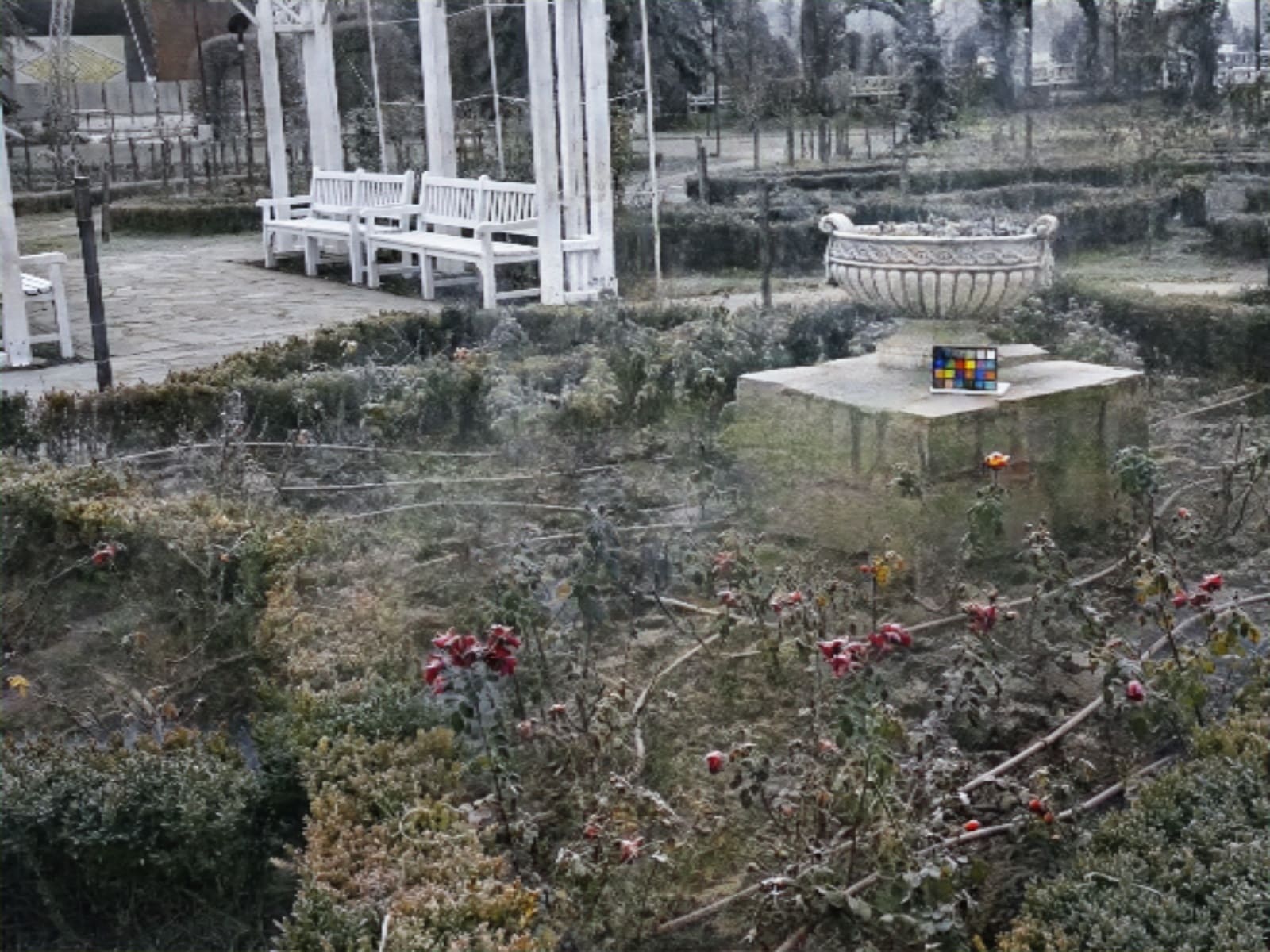}\\
    \vspace{0.05cm}
    \includegraphics[width=1.0\linewidth]{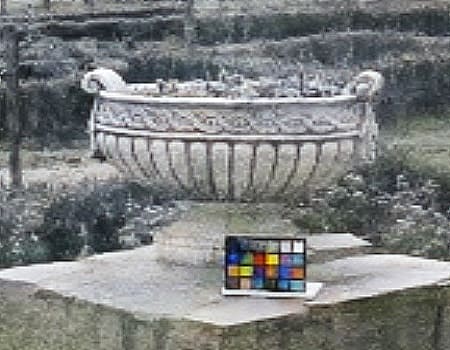}
    \captionsetup{font={small}}
    \caption{+\textbf{P} +\textbf{S} +\textbf{T}}
    \label{fig:ablation-d}
  \end{subfigure}
  \hfill
  \begin{subfigure}{0.171\linewidth}
    \centering
    \includegraphics[width=1.0\linewidth]{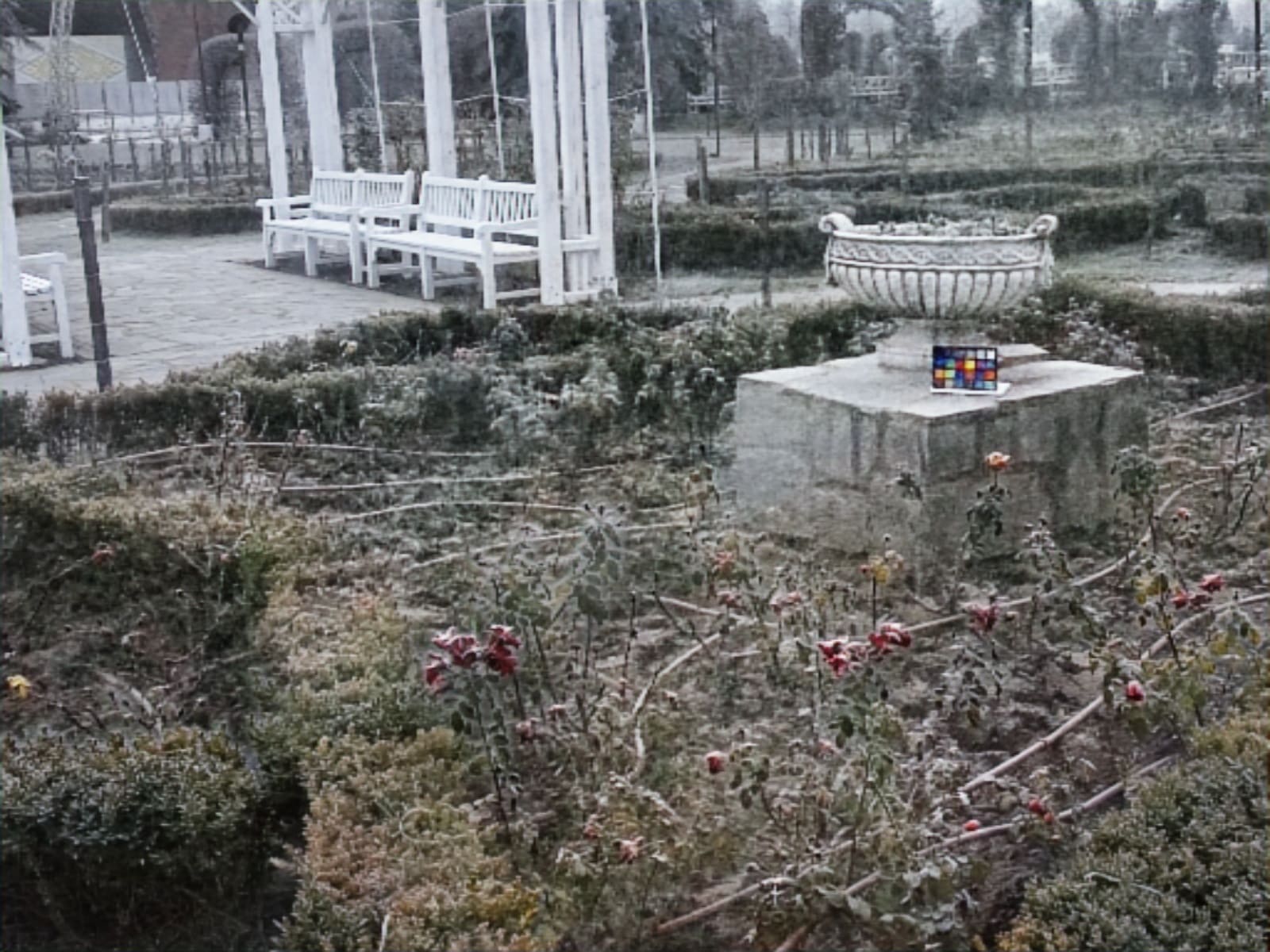}\\
    \vspace{0.05cm}
    \includegraphics[width=1.0\linewidth]{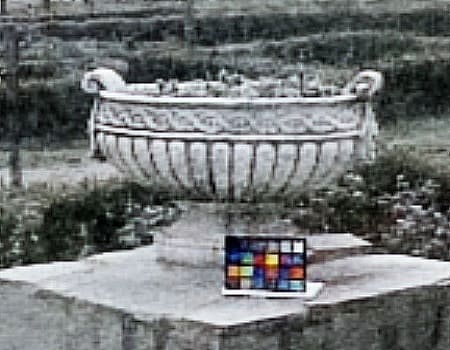}
    \captionsetup{font={small}}
    \caption{+\textbf{P} +\textbf{S} +\textbf{T} +\textbf{H}}
    \label{fig:ablation-f}
  \end{subfigure}
    \begin{subfigure}{0.171\linewidth}
    \centering
    \includegraphics[width=1.0\linewidth]{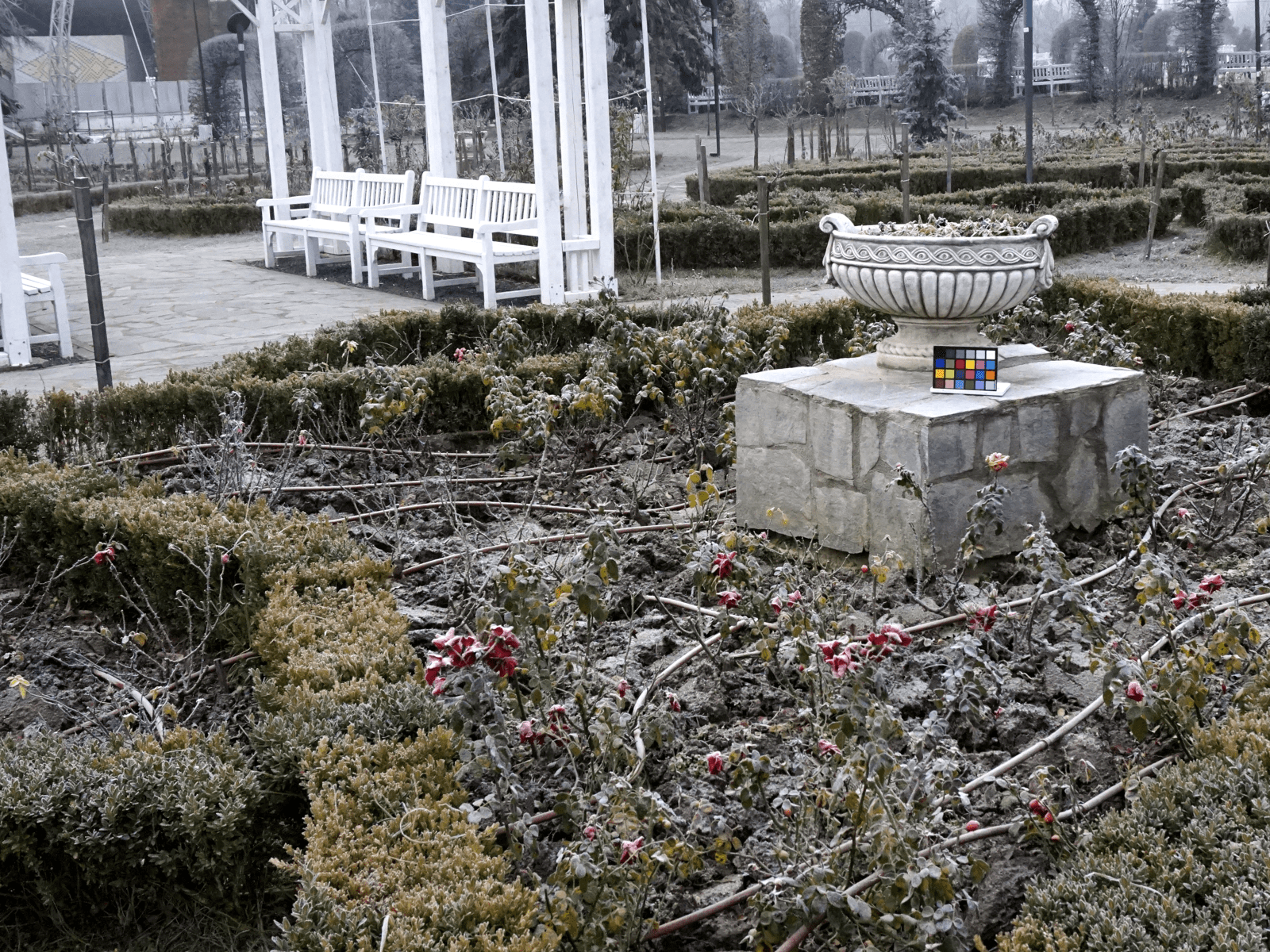}\\
    \vspace{0.05cm}
    \includegraphics[width=1.0\linewidth]{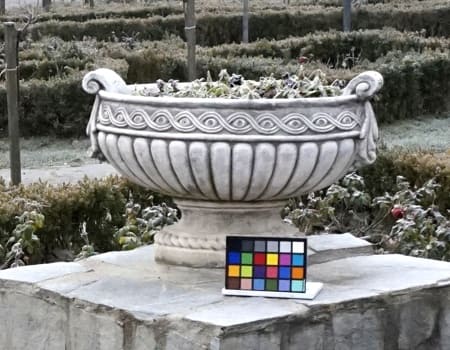}
    \captionsetup{font={small}}
    \caption{Clean}
    \label{fig:ablation-g}
  \end{subfigure}}
  
  \caption{Effects of PTB (\textbf{P}), SPP (\textbf{S}), $\mathcal{L}_{Trans}$ (\textbf{T}), and Hint (\textbf{H}) on visual outcomes.}
  \label{fig:ablation}
  
\end{figure*}

\subsubsection{Qualitative Evaluation}

Typical visual results produced by various approaches across four benchmark datasets are shown in Fig.s \ref{fig:ihaze-result}-\ref{fig:nhhaze-result}. From these figures, it is evident that prior-based methods often struggle with haze residue or unwanted color distortions, which negatively impact the clarity and natural appearance of the restored images. In contrast, DL-based approaches generally achieve visually superior outcomes, with fewer artifacts and more natural colors.

In particular, Fig. \ref{fig:ihaze-result} and Fig. \ref{fig:ohaze-result} illustrate representative results on the I-HAZE and O-HAZE datasets, which feature homogeneous haze conditions. In these cases, most DL-based methods can produce competitive results that effectively remove haze and enhance image clarity. On the other hand, Fig. \ref{fig:densehaze-result} and Fig. \ref{fig:nhhaze-result} depict visual results on the more challenging Dense-HAZE and NH-HAZE datasets, which contain dense or non-homogeneous haze. For these complex scenarios, it is observed that heavyweight DL-based models generally deliver more favorable restorations, showing a better ability to handle variations in haze density. However, the proposed DPTE-Net with a low complexity can still produce comparable visual quality across all datasets. Notably, DPTE-Net can manage to maintain natural color tones and minimize distortions, even in challenging conditions, often matching or surpassing the performance of more computationally expensive models like EDN-GTM and Dehamer.

\begin{table}
  \caption{Average runtime (in seconds) of various methods tested on GPU (GeForce GTX TITAN X) and CPU (Intel(R) Xeon(R) Gold 6134 @ 3.20GHz).}
  \centering
  \resizebox{0.35\textwidth}{!}{
  \begin{tabular}{cccc}
    \toprule
    \multirow{2}{*}{Methods} & \multirow{2}{*}{Framework} & \multicolumn{2}{c}{Runtime (sec.)} \\
    \cmidrule{3-4}
     & & GPU & CPU \\
    \midrule
    GCANet \cite{chen2019gated} & PyTorch & 0.052 & 1.87 \\
    GridDehaze \cite{liu2019griddehazenet} & PyTorch & 0.064 & 0.58 \\
    FFA-Net \cite{qin2020ffa} & PyTorch & 0.982 & 10.40 \\
    Cycle-GAN \cite{zhu2017unpaired} & TensorFlow & 0.074 & 0.84 \\
    EDN-GTM \cite{tran2024encoder} & TensorFlow & 0.102 & 0.98 \\
    Restormer \cite{zamir2022restormer} & PyTorch & 0.756 & 8.07 \\
    Dehamer \cite{guo2022image} & PyTorch & 0.128 & 1.96 \\
    RefineDNet \cite{zhao2021refinednet} & PyTorch & 0.069 & 0.75 \\
    DehazeFormer \cite{song2023vision} & HuggingFace & - & 2.14 \\
    \midrule
    DPTE-Net (ours) & TensorFlow & \textbf{0.048} & \textbf{0.38} \\
    \bottomrule
  \end{tabular}}
  \label{tab2:runtime}
\end{table}

\subsection{Runtime Comparison}
\label{subsec:runtime}

While \#Params and MACs can offer some insight into the model complexity to a certain extent, they are not comprehensive indicators of inference latency since they fail to consider several inference-related factors such as memory access, degree of parallelism, and platform characteristics \cite{mehta2021mobilevit}. Hence, we have conducted a comparison of various DL-based models in terms of average inference time on CPU (Intel(R) Xeon(R) Gold 6134 @ 3.20GHz) and GPU (GeForce GTX TITAN X). As summarized in Table \ref{tab2:runtime}, the proposed DPTE-Net can be considered the most efficient approach with the fastest inference speeds on both CPU and GPU platforms as compared to the other DL-based methods under consideration. Note that DPTE-Net does not utilize the full pipeline of DCP but only two steps including patch-level min and smoothing filtering, which consume only 0.015 seconds on modern computers.

\begin{table}
  \centering
  \caption{Effects of replacing SA with pooling.}
  \resizebox{0.48\textwidth}{!}{%
  \begin{tabular}{cccccrr}
    \toprule
    \multirow{2}{*}{Encoder} & \multicolumn{2}{c}{Dense-HAZE} & \multicolumn{2}{c}{NH-HAZE} & \multirow{2}{*}{\#Params$\downarrow$} & \multirow{2}{*}{MACs$\downarrow$}  \\
    \cmidrule{2-5}
        & PSNR & SSIM & PSNR & SSIM \\
    \midrule
    (baseline) & 14.59 & 0.5146 & 17.98 & 0.5088 & 2.56M & 16.0G \\
    +SA & 15.35 & 0.5207 & 19.22 & 0.5345 & 8.46M & 76.4G \\
    +PTB & 15.26 & 0.5168 & 18.96 & 0.5237 & \textbf{3.06M} & \textbf{18.6G} \\
    \bottomrule
  \end{tabular}}
  \label{tab:savspooling}
\end{table}

\subsection{Ablation Study}
\label{subsec:ablationstudy}

\subsubsection{Effects of Integrated Blocks}
\label{subsubsec:integrated_blocks}

Ablation studies have been carried out on the Dense-HAZE and NH-HAZE datasets in order to assess the impacts of integrated components including PTB, SPP, $\mathcal{L}_{Trans}$, and Hint on the overall network performance. To this end, a baseline network was designed using a plain U-Net backbone without any integration of the aforementioned modules. Subsequently, the performance of the network was analyzed by incrementally introducing each component.

First, to investigate the effects of substituting SA with pooling as a token mixing method, we conducted experiments starting from the baseline model. Specifically, we incorporated both SA and pooling separately to evaluate their impact on performance and computational efficiency, with quantitative results shown in Table \ref{tab:savspooling}. As illustrated in Table \ref{tab:savspooling}, replacing SA with pooling yields substantial reductions in model complexity while only slightly affecting the model's effectiveness. Specifically, \#Params and MACs are reduced by 64\% (8.46M $\rightarrow$ 3.06M) and 76\% (76.4G $\rightarrow$ 18.6G), respectively. These results indicate that pooling can serve as an efficient token-mixing method, facilitating a lightweight model architecture without a significant compromise in performance.

\begin{table}
  \centering
  \caption{Effects of PTB, SPP, $\mathcal{L}_{Trans}$, and Hint.}
  \resizebox{0.48\textwidth}{!}{%
  \begin{tabular}{cccccccc}
    \toprule
    \multirow{2}{*}{+PTB} & \multirow{2}{*}{+SPP} & \multirow{2}{*}{+$\mathcal{L}_{Trans}$} & \multirow{2}{*}{+Hint} & \multicolumn{2}{c}{Dense-HAZE} & \multicolumn{2}{c}{NH-HAZE} \\
    \cmidrule{5-8}
     & & & & PSNR & SSIM & PSNR & SSIM \\
    \midrule
    \multicolumn{4}{c}{(baseline)} & 14.59 & 0.5146 & 17.98 & 0.5088 \\
    \midrule
    \checkmark & & & & 15.26 & 0.5168 & 18.96 & 0.5237 \\
    \checkmark & \checkmark & & & 15.27 & 0.5175 & 19.43 & 0.5486 \\
    \checkmark & \checkmark & \checkmark & & 15.31 & 0.5169 & 19.54 & 0.5489 \\
    \midrule
    \checkmark & \checkmark & \checkmark & \checkmark & \textbf{15.59} & \textbf{0.5248} & \textbf{20.18} & \textbf{0.5623} \\
    \bottomrule
  \end{tabular}}
  \label{tab:ablation_study}
\end{table}

\begin{table}
  \centering
  \caption{Effects of two-stage training with different settings of $\delta$.}
  \resizebox{0.30\textwidth}{!}{%
  \begin{tabular}{ccccc}
    \toprule
    \multirow{2}{*}{\quad$\delta$\quad} & \multicolumn{2}{c}{Dense-HAZE} & \multicolumn{2}{c}{NH-HAZE} \\
    \cmidrule{2-5}
        & PSNR & SSIM & PSNR & SSIM \\
    \midrule
    0.01 & 15.28 & 0.5155 & 19.58 & 0.5472 \\
    0.5 & \textbf{15.59} & \textbf{0.5248} & \textbf{20.18} & \textbf{0.5623} \\
    1.0 & 15.38 & 0.5197 & 19.75 & 0.5586 \\
    \bottomrule
  \end{tabular}}
  \label{tab:lamda}
\end{table}

The effects of PTB, SPP, $\mathcal{L}_{Trans}$, and Hint on the network performance in terms of PSNR and SSIM are summarized in Table \ref{tab:ablation_study}. It can be observed from Table \ref{tab:ablation_study} that all the components have positive influences on the dehazing performance. While PTB, SPP, and Hint considerably improve the performance over the baseline, $\mathcal{L}_{Trans}$ only results in a slight performance gain. This may stem from the consequence of the integral loss when training the student network. That is, several loss terms are adopted and the contributions of those loss functions are averaged using balancing weights. Despite that, any performance improvement is considered important. An illustration of the effects of those components on the visual dehazing outcomes is shown in Fig. \ref{fig:ablation}. Specifically, Fig. \ref{fig:ablation}b shows the baseline model's result (without PTB, SPP, $\mathcal{L}_{Trans}$, and Hint). Fig. \ref{fig:ablation}c implies that the visual outcome can be improved when conventional convolution blocks in the encoder are replaced with the PTB module. Fig. \ref{fig:ablation}d shows the result when SPP is appended to the bottleneck and $\mathcal{L}_{Trans}$ is adopted in training, which can encourage the network to focus on significant regions. Eventually, it can be seen from Fig. \ref{fig:ablation}e that the combination of all four components can result in an improved visual quality of the restored image.

\begin{figure}
  \centering  
  \resizebox{0.98\linewidth}{!}{
  \begin{subfigure}{0.34\linewidth}
    \centering 
    \includegraphics[width=1.0\linewidth]{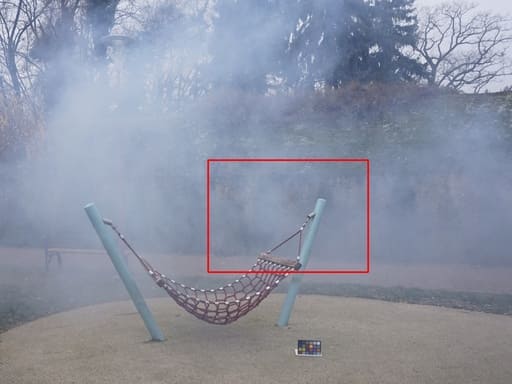}\\
    \captionsetup{font={footnotesize}}
    \text{Hazy}
    \label{fig:ablation2-a}
  \end{subfigure}
  \hfill
  \begin{subfigure}{0.34\linewidth}
    \centering
    \includegraphics[width=1.0\linewidth]{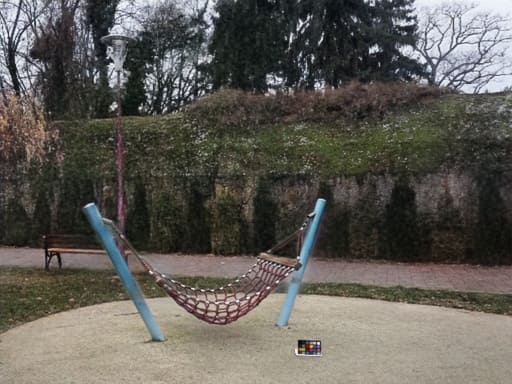}\\
    \captionsetup{font={footnotesize}}
    \text{DPTE-Net ($\delta=0.5)$}
    \label{fig:ablation2-b}
  \end{subfigure}
  \hfill
  \begin{subfigure}{0.34\linewidth}
    \centering
    \includegraphics[width=1.0\linewidth]{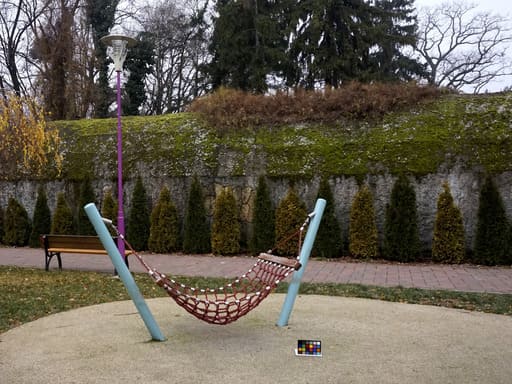}\\
    \captionsetup{font={footnotesize}}
    \text{Clean}
    \label{fig:ablation2-c}
  \end{subfigure}}

  \vspace{0.05cm}

  \resizebox{0.98\linewidth}{!}{
  \begin{subfigure}{0.20\linewidth}
    \centering 
    \includegraphics[width=1.0\linewidth]{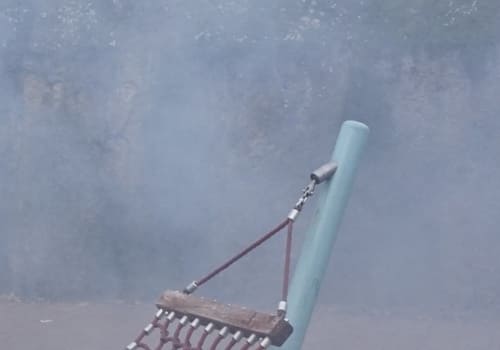}\\
    \captionsetup{font={footnotesize}}
    \text{Hazy crop}
    \label{fig:ablation2-d}
  \end{subfigure}
  \hfill
  \begin{subfigure}{0.20\linewidth}
    \centering
    \includegraphics[width=1.0\linewidth]{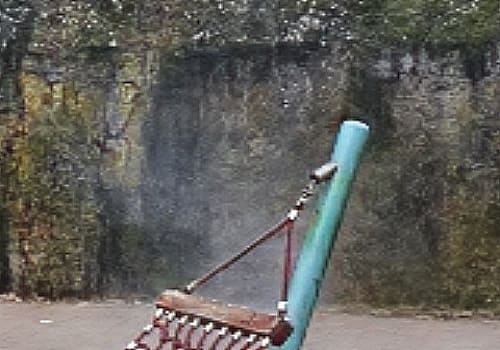}\\
    \captionsetup{font={footnotesize}}
    \text{$\delta=0.01$}
    \label{fig:ablation2-e}
  \end{subfigure}
  \hfill
  \begin{subfigure}{0.20\linewidth}
    \centering
    \includegraphics[width=1.0\linewidth]{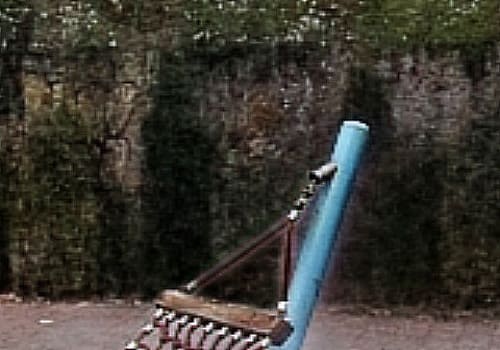}\\
    \captionsetup{font={footnotesize}}
    \text{$\delta=0.5$}
    \label{fig:ablation2-f}
  \end{subfigure}
  \hfill
  \begin{subfigure}{0.20\linewidth}
    \centering
    \includegraphics[width=1.0\linewidth]{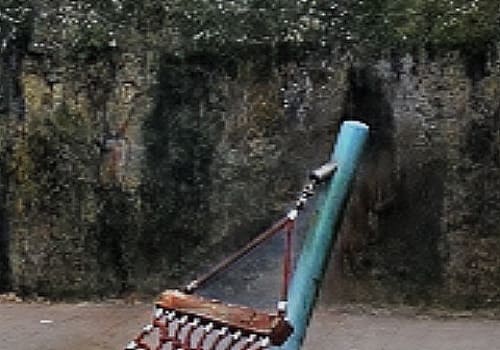}\\
    \captionsetup{font={footnotesize}}
    \text{$\delta=1.0$}
    \label{fig:ablation2-g}
  \end{subfigure}
  \hfill
  \begin{subfigure}{0.20\linewidth}
    \centering
    \includegraphics[width=1.0\linewidth]{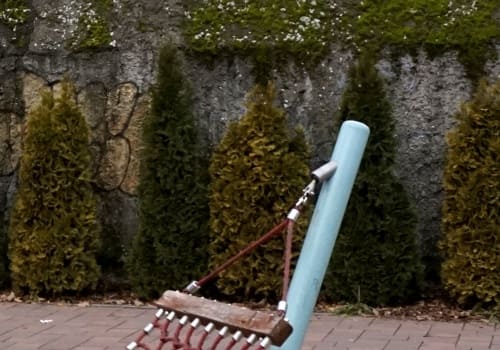}\\
    \captionsetup{font={footnotesize}}
    \text{Clean crop}
    \label{fig:ablation2-h}
  \end{subfigure}}
  
  \caption{Effects of two-stage training with different configurations of $\delta$ value.}
  \label{fig:ablation2}
  
\end{figure}

\begin{figure}
  \centering
  \includegraphics[width=1.0\linewidth]{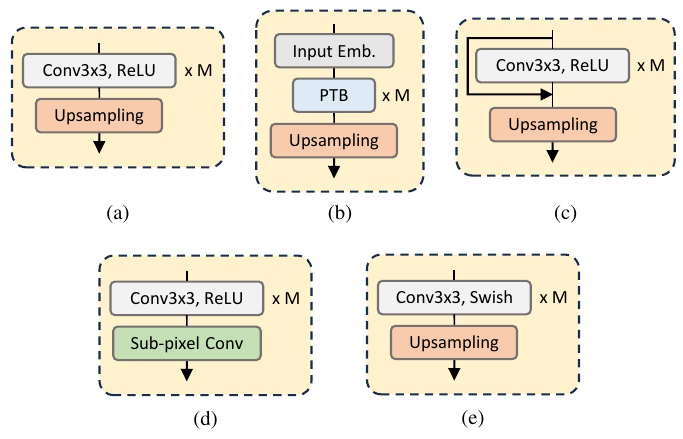}
  \caption{Candidate decoder structures: (a,e) U-Net's decoder with ReLU and Swish activations, respectively, and (b,c,d) other decoder structures with PTB module, residual connection, and sub-pixel convolution, respectively. Each structure denotes one level of the decoder.}
  \label{fig:decoder_structures}
\end{figure}

\begin{figure*}[t!]
  \centering  
  \resizebox{0.78\textwidth}{!}{
  \begin{subfigure}{0.201\linewidth}
    \centering 
    {\small PSNR/SSIM}
    \includegraphics[width=1.0\linewidth]{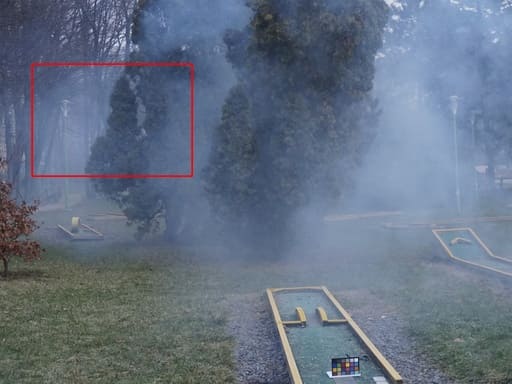}\\
    \vspace{0.05cm}
    \includegraphics[width=1.0\linewidth]{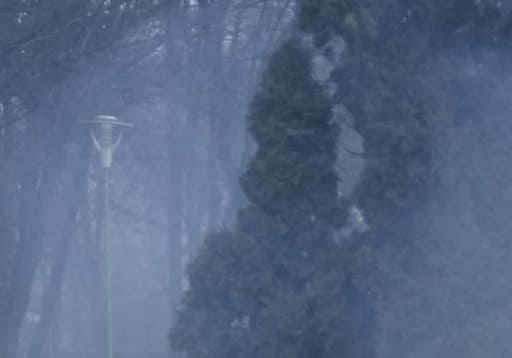}
    \captionsetup{font={small}}
    \caption{Hazy}
    \label{fig:ablation_decoder-a}
  \end{subfigure}
  \hfill
  \begin{subfigure}{0.201\linewidth}
    \centering
    {\small 18.83/0.5635}
    \includegraphics[width=1.0\linewidth]{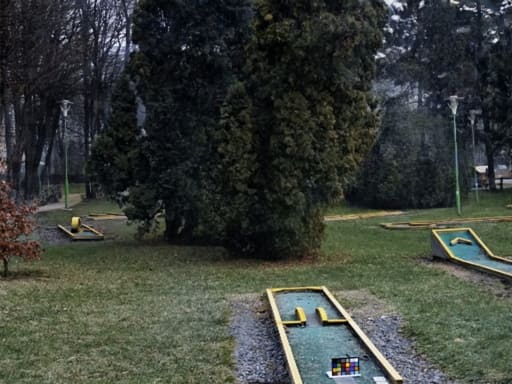}\\
    \vspace{0.05cm}
    \includegraphics[width=1.0\linewidth]{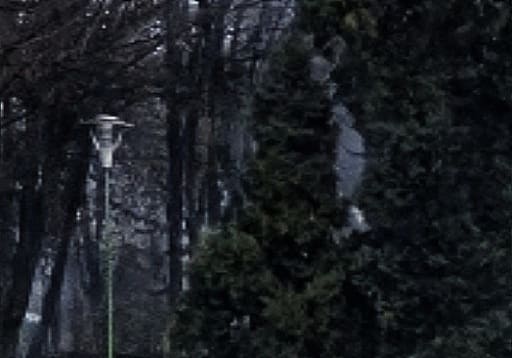}
    \captionsetup{font={small}}
    \caption{\textbf{ReLU-D}}
    \label{fig:ablation_decoder-b}
  \end{subfigure}
  \hfill
  \begin{subfigure}{0.201\linewidth}
    \centering
    {\small 18.67/0.5887}
    \includegraphics[width=1.0\linewidth]{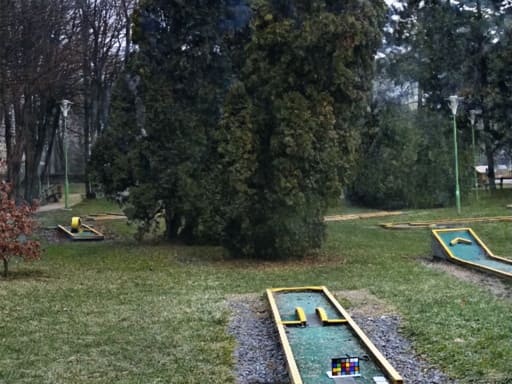}\\
    \vspace{0.05cm}
    \includegraphics[width=1.0\linewidth]{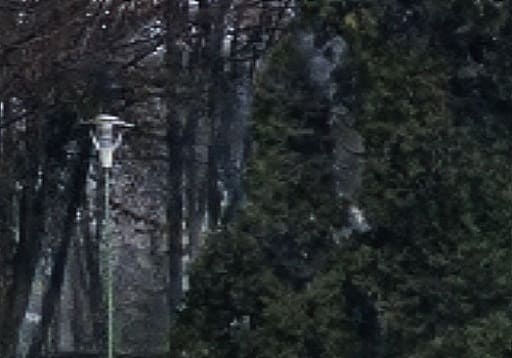}
    \captionsetup{font={small}}
    \caption{\textbf{PTB-D}}
    \label{fig:ablation_decoder-c}
  \end{subfigure}
  \hfill
  \begin{subfigure}{0.201\linewidth}
    \centering
    {\small 18.89/0.5961}
    \includegraphics[width=1.0\linewidth]{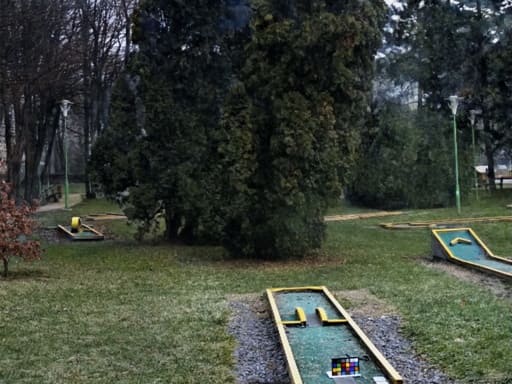}\\
    \vspace{0.05cm}
    \includegraphics[width=1.0\linewidth]{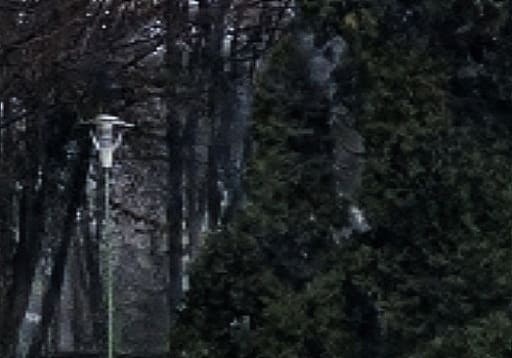}
    \captionsetup{font={small}}
    \caption{\textbf{Res-D}}
    \label{fig:ablation_decoder-d}
  \end{subfigure}}

  \vspace{0.10cm}

  \resizebox{0.78\textwidth}{!}{
  \begin{subfigure}{0.201\linewidth}
    \centering
    {\small 18.32/0.5458}
    \includegraphics[width=1.0\linewidth]{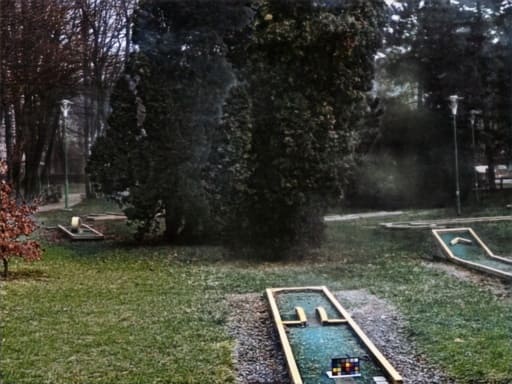}\\
    \vspace{0.05cm}
    \includegraphics[width=1.0\linewidth]{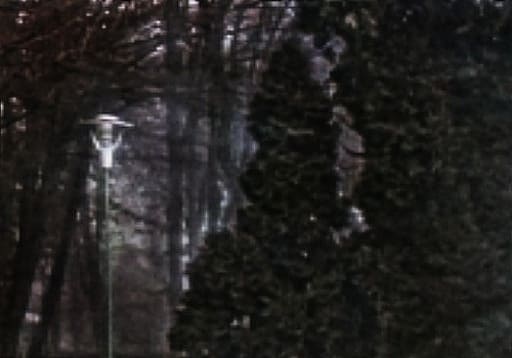}
    \captionsetup{font={small}}
    \caption{\textbf{Sub-D}}
    \label{fig:ablation_decoder-e}
  \end{subfigure}
  \begin{subfigure}{0.201\linewidth}
    \centering
    {\small 21.23/0.6014}
    \includegraphics[width=1.0\linewidth]{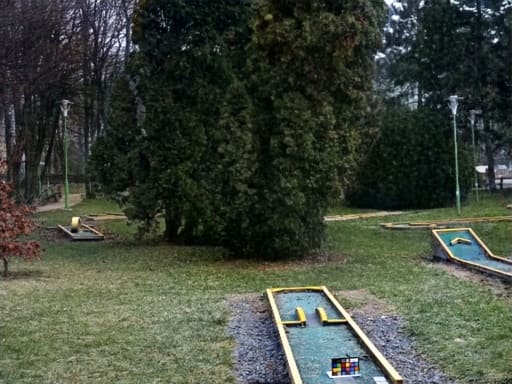}\\
    \vspace{0.05cm}
    \includegraphics[width=1.0\linewidth]{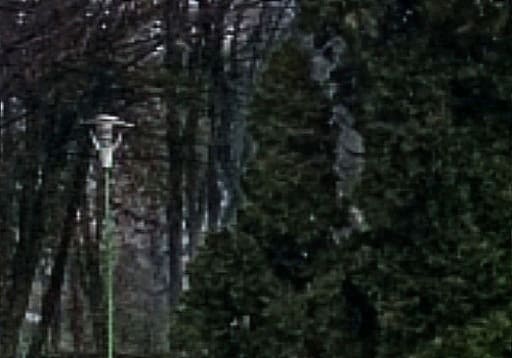}
    \captionsetup{font={small}}
    \caption{\textbf{w/o PTB-A}}
    \label{fig:ablation_decoder-f}
  \end{subfigure}
  \hfill
  \begin{subfigure}{0.201\linewidth}
    \centering
    {\small 21.79/0.6155}
    \includegraphics[width=1.0\linewidth]{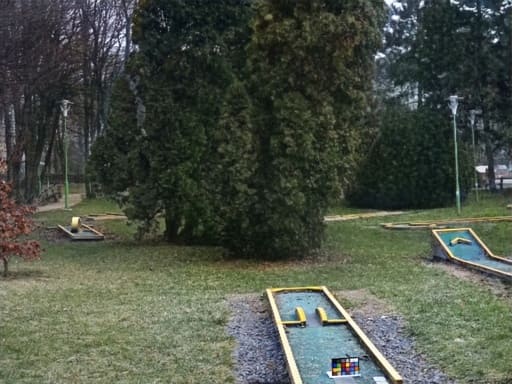}\\
    \vspace{0.05cm}
    \includegraphics[width=1.0\linewidth]{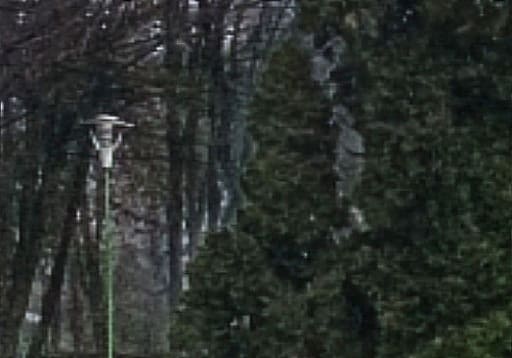}
    \captionsetup{font={small}}
    \caption{\textbf{DPTE-Net}}
    \label{fig:ablation_decoder-g}
  \end{subfigure}
  \hfill
  \begin{subfigure}{0.201\linewidth}
    \centering
    {\small $\infty$/1.0}
    \includegraphics[width=1.0\linewidth]{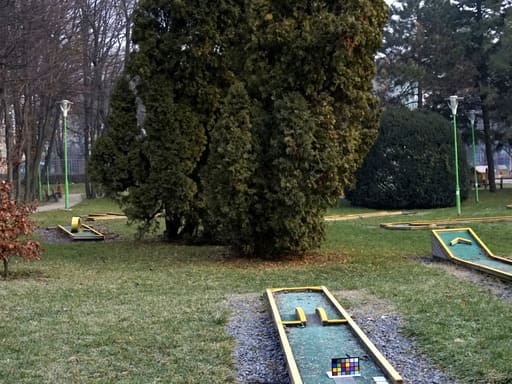}\\
    \vspace{0.05cm}
    \includegraphics[width=1.0\linewidth]{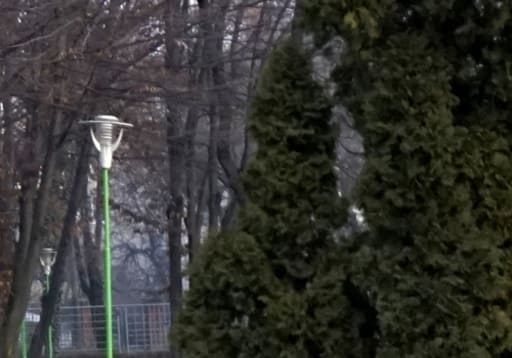}
    \captionsetup{font={small}}
    \caption{Clean}
    \label{fig:ablation_decoder-h}
  \end{subfigure}}
  
  \caption{Effects of different decoder types and PTB-based adaptation layer on visual outcomes: (b) ReLU-Decoder (\textbf{ReLU-D}), (c) PTB-Decoder (\textbf{PTB-D}), (d) Residual-Decoder (\textbf{Res-D}), (e) Subpixel-Decoder (\textbf{Sub-D}), (f) without PTB in adaptation (\textbf{w/o PTB-A}), and (g) Swish-Decoder | PTB in adaptation (\textbf{DPTE-Net}).}
  \label{fig:ablation_decoder}
  
\end{figure*}

\begin{table*}
  \centering
  \caption{Effects of different decoder types.}
  \resizebox{0.9\textwidth}{!}{%
  \begin{tabular}{cccccccccrr}
    \toprule
    \multirow{2}{*}{Type} & \multicolumn{2}{c}{I-HAZE} & \multicolumn{2}{c}{O-HAZE} & \multicolumn{2}{c}{Dense-HAZE} & \multicolumn{2}{c}{NH-HAZE} & \multirow{2}{*}{\#Params$\downarrow$} & \multirow{2}{*}{MACs$\downarrow$}  \\
    \cmidrule{2-9}
        & PSNR$\uparrow$ & SSIM$\uparrow$ & PSNR$\uparrow$ & SSIM$\uparrow$ & PSNR$\uparrow$ & SSIM$\uparrow$ & PSNR$\uparrow$ & SSIM$\uparrow$ \\
    \midrule
    ReLU-Decoder & 20.45 & 0.7835 & 21.23 & 0.5846 & 15.11 & 0.5165 & 19.61 & 0.5086 & 3.10M & 19.2G \\
    PTB-Decoder & 19.91 & 0.7651 & 20.77 & 0.5763 & 14.82 & 0.4917 & 18.89 & 0.4975 & 3.08M & 18.8G \\
    Residual-Decoder & 20.87 & 0.7888 & 21.78 & 0.5910 & 15.16 & 0.5166 & 19.63 & 0.5098 & 3.10M & 19.2G \\
    Subpixel-Decoder & 18.36 & 0.7346 & 19.85 & 0.5468 & 14.56 & 0.4852 & 18.06 & 0.4968 & \textbf{2.99M} & \textbf{17.6G} \\
    Swish-Decoder & \textbf{21.68} & \textbf{0.8164} & \textbf{22.02} & \textbf{0.6901} & \textbf{15.59} & \textbf{0.5248} & \textbf{20.18} & \textbf{0.5623} & 3.10M & 19.2G \\
    \bottomrule
  \end{tabular}}
  \label{tab:decoder}
\end{table*}

\subsubsection{Effects of Two-stage Training}
\label{subsubsec:twostagetraining}

The effects of the two-stage training process which is influenced by the loss balancing decay $\lambda$ are also investigated. As in Eq. (\ref{eq5}), $\lambda$ is calculated based on $\delta$ which controls the period of Stage I. Hence, different configurations of $\delta$ result in changes in the training process. To present a thorough examination, we examine three different settings of $\delta$, with $\delta \in \{0.01,0.5,1.0\}$. Note that we chose $0.01$ as the minimum value for $\delta$ since $\delta > 0$. When $\delta = 0.01$, as in Eq. (\ref{eq5}), the network is trained mostly without Stage I (without Hint), which represents a single-stage end-to-end training process. When $\delta = 1.0$, then $\lambda > 0$ during all training time, meaning that the student may converge to a weighted average solution between the teacher's convergence space and clean feature space, which can be considered a suboptimal solution. When $\delta = 0.5$, the periods for Stage I and Stage II are equal, representing the proposed two-phase training scheme. As shown in Table \ref{tab:lamda} and Fig. \ref{fig:ablation2}, $\delta = 0.5$ results in the most favorable dehazing effectiveness, this implies that balancing the periods of the two stages tends to achieve a more desired performance.

\begin{figure*}
  \centering

  \resizebox{0.9988\textwidth}{!}{
  \begin{subfigure}{0.151\linewidth}
    \centering
    \includegraphics[width=1.0\linewidth]{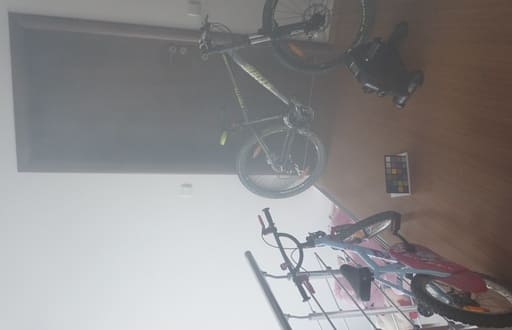}\\
    \vspace{0.05cm}
    \includegraphics[width=1.0\linewidth]{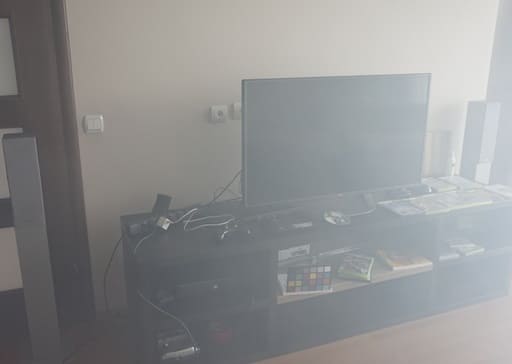}
    \captionsetup{font={small}}
    \text{Hazy}
  \end{subfigure}
  \begin{subfigure}{0.151\linewidth}
    \centering
    \includegraphics[width=1.0\linewidth]{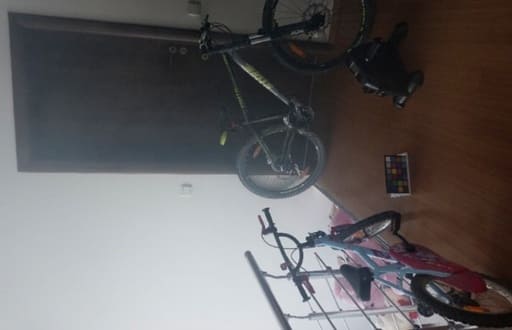}\\
    \vspace{0.05cm}
    \includegraphics[width=1.0\linewidth]{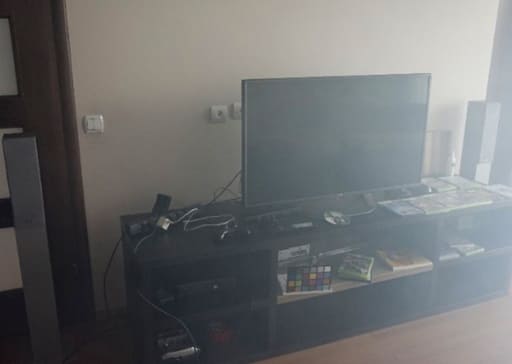}
    \captionsetup{font={small}}
    \text{GridDehaze \cite{liu2019griddehazenet}}
  \end{subfigure}
  \begin{subfigure}{0.151\linewidth}
    \centering
    \includegraphics[width=1.0\linewidth]{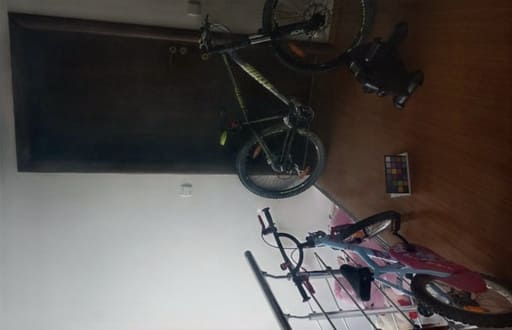}\\
    \vspace{0.05cm}
    \includegraphics[width=1.0\linewidth]{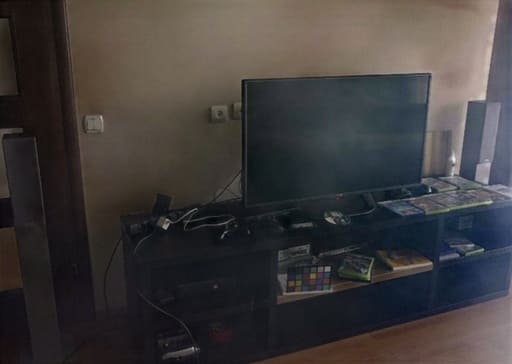}
    \captionsetup{font={small}}
    \text{GCANet \cite{chen2019gated}}
  \end{subfigure}
  \hfill
  \begin{subfigure}{0.151\linewidth}
    \centering
    \includegraphics[width=1.0\linewidth]{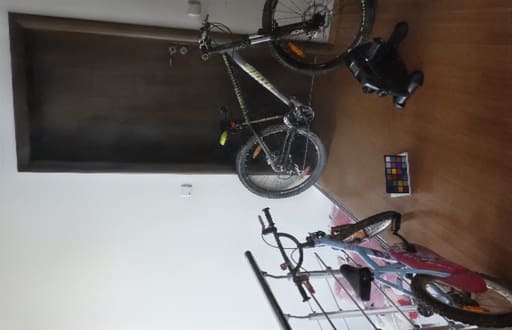}\\
    \vspace{0.05cm}
    \includegraphics[width=1.0\linewidth]{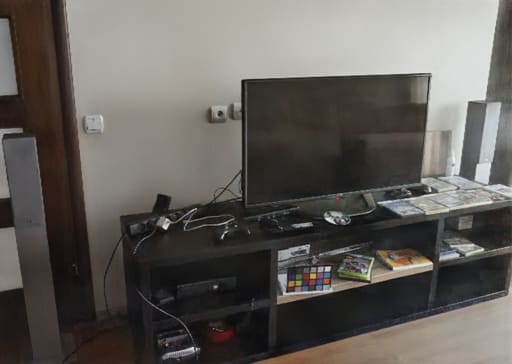}
    \captionsetup{font={small}}
    \text{EDN-GTM \cite{tran2024encoder}}
  \end{subfigure}
  \hfill
  \begin{subfigure}{0.151\linewidth}
    \centering
    \includegraphics[width=1.0\linewidth]{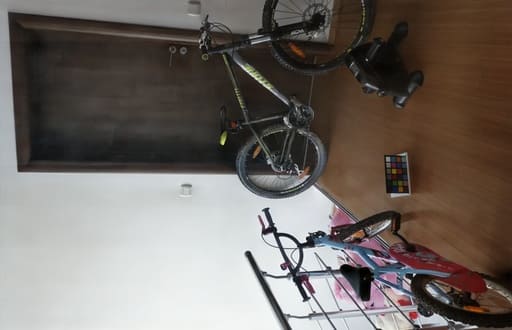}\\
    \vspace{0.05cm}
    \includegraphics[width=1.0\linewidth]{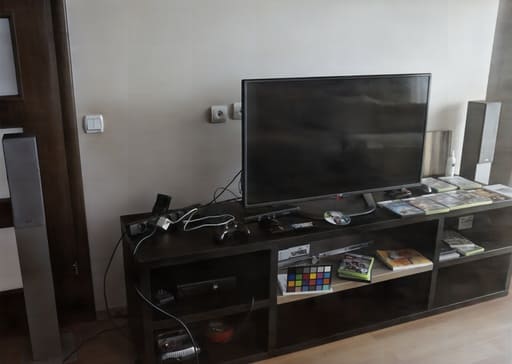}
    \captionsetup{font={small}}
    \text{PDD-Net \cite{zhang2018multi}}
  \end{subfigure}
  \hfill
  \begin{subfigure}{0.151\linewidth}
    \centering
    \includegraphics[width=1.0\linewidth]{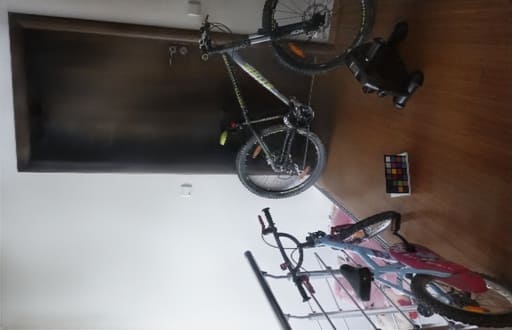}\\
    \vspace{0.05cm}
    \includegraphics[width=1.0\linewidth]{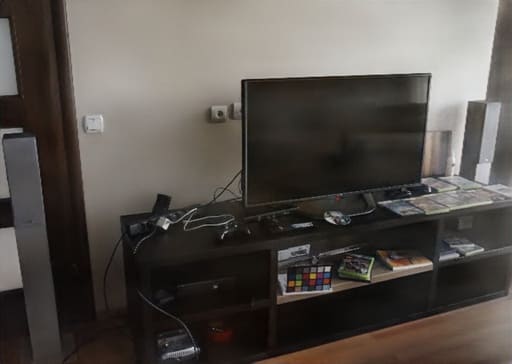}
    \captionsetup{font={small}}
    \text{DPTE-Net (ours)}
  \end{subfigure}
  \hfill
  \begin{subfigure}{0.151\linewidth}
    \centering
    \includegraphics[width=1.0\linewidth]{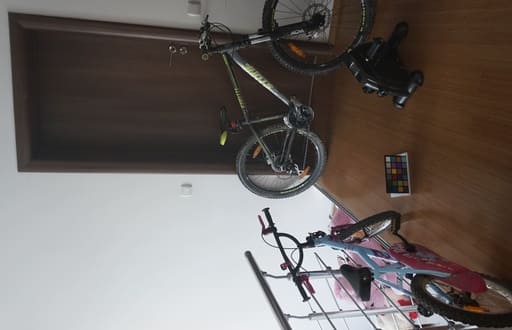}\\
    \vspace{0.05cm}
    \includegraphics[width=1.0\linewidth]{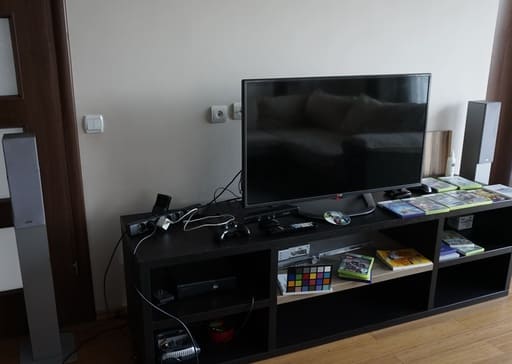}
    \captionsetup{font={small}}
    \text{Clean}
  \end{subfigure}}

  \vspace{0.1cm}
  
  \resizebox{0.9988\textwidth}{!}{
  \begin{subfigure}{0.151\linewidth}
    \centering
    \includegraphics[width=1.0\linewidth]{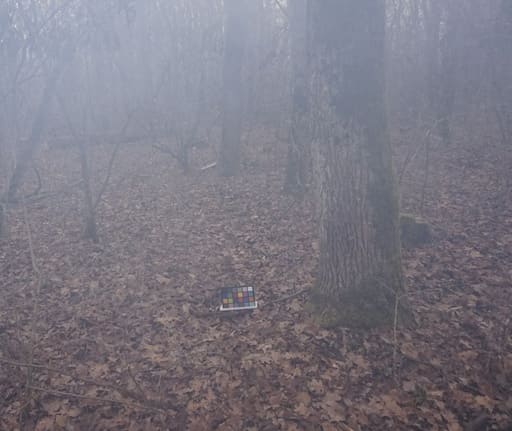}\\
    \vspace{0.05cm}
    \includegraphics[width=1.0\linewidth]{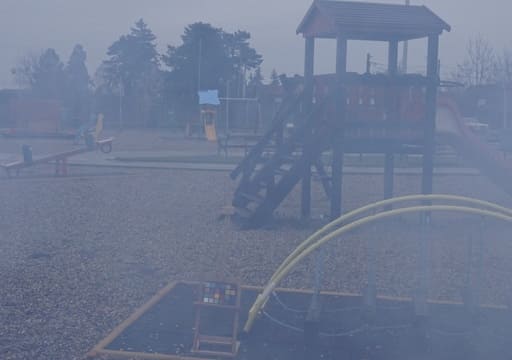}
    \captionsetup{font={small}}
    \text{Hazy}
  \end{subfigure}
  \begin{subfigure}{0.151\linewidth}
    \centering
    \includegraphics[width=1.0\linewidth]{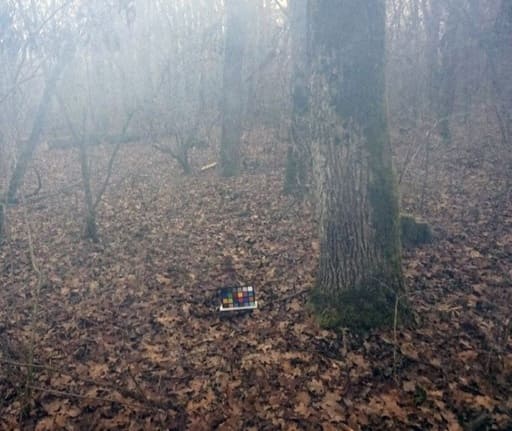}\\
    \vspace{0.05cm}
    \includegraphics[width=1.0\linewidth]{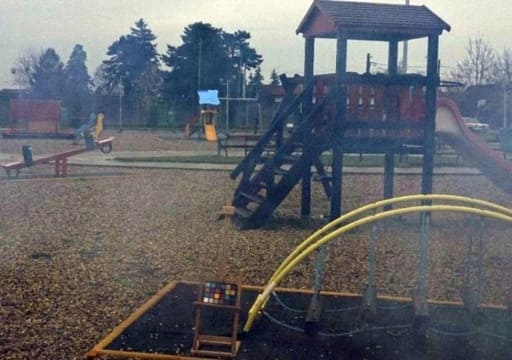}
    \captionsetup{font={small}}
    \text{GridDehaze \cite{liu2019griddehazenet}}
  \end{subfigure}
  \begin{subfigure}{0.151\linewidth}
    \centering
    \includegraphics[width=1.0\linewidth]{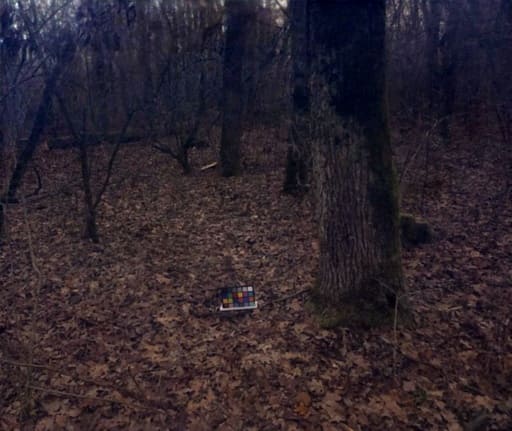}\\
    \vspace{0.05cm}
    \includegraphics[width=1.0\linewidth]{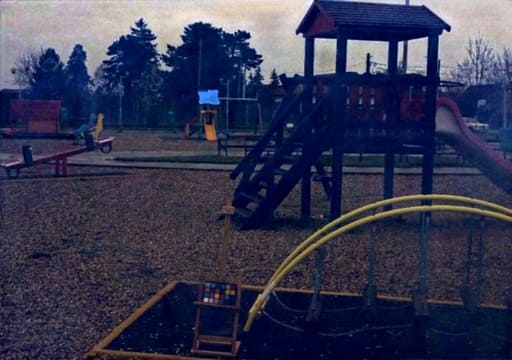}
    \captionsetup{font={small}}
    \text{GCANet \cite{chen2019gated}}
  \end{subfigure}
  \hfill
  \begin{subfigure}{0.151\linewidth}
    \centering
    \includegraphics[width=1.0\linewidth]{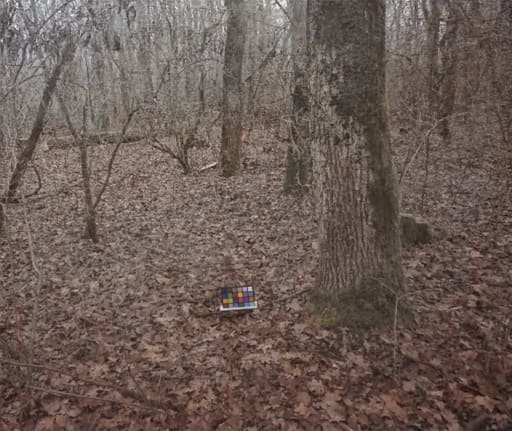}\\
    \vspace{0.05cm}
    \includegraphics[width=1.0\linewidth]{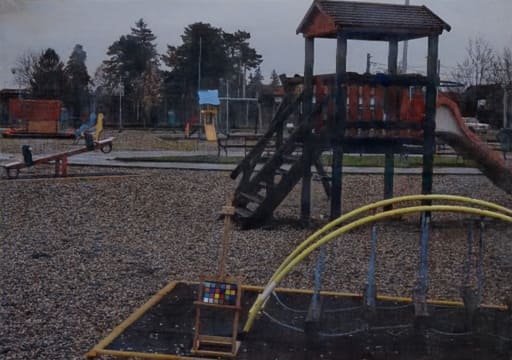}
    \captionsetup{font={small}}
    \text{EDN-GTM \cite{tran2024encoder}}
  \end{subfigure}
  \hfill
  \begin{subfigure}{0.151\linewidth}
    \centering
    \includegraphics[width=1.0\linewidth]{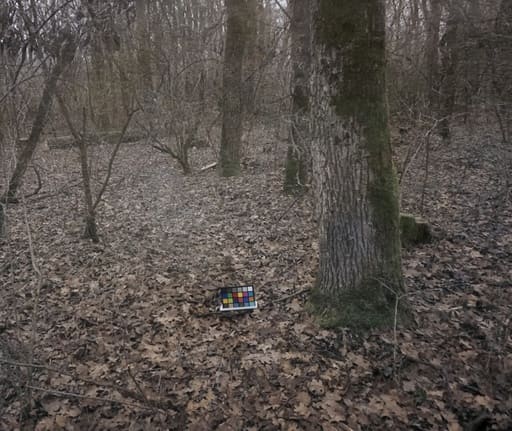}\\
    \vspace{0.05cm}
    \includegraphics[width=1.0\linewidth]{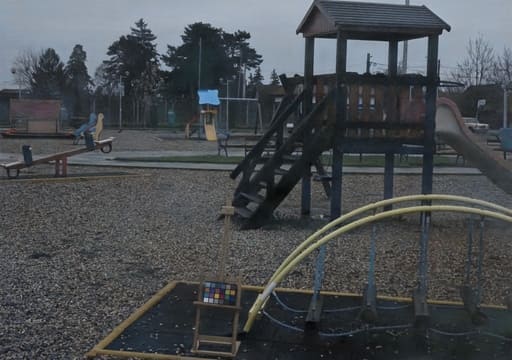}
    \captionsetup{font={small}}
    \text{PDD-Net \cite{zhang2018multi}}
  \end{subfigure}
  \hfill
  \begin{subfigure}{0.151\linewidth}
    \centering
    \includegraphics[width=1.0\linewidth]{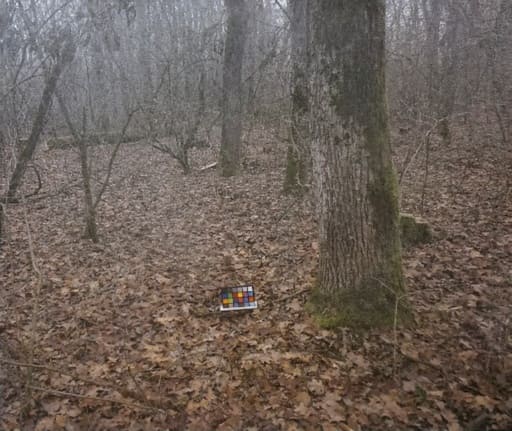}\\
    \vspace{0.05cm}
    \includegraphics[width=1.0\linewidth]{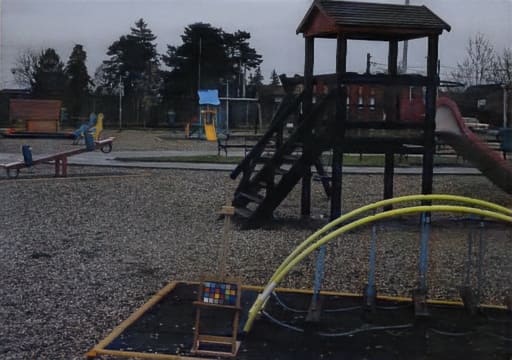}
    \captionsetup{font={small}}
    \text{DPTE-Net (ours)}
  \end{subfigure}
  \hfill
  \begin{subfigure}{0.151\linewidth}
    \centering
    \includegraphics[width=1.0\linewidth]{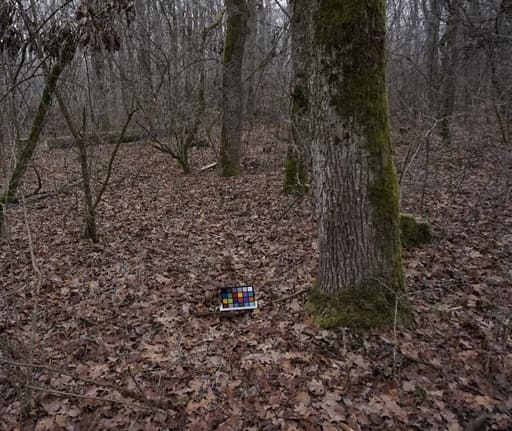}\\
    \vspace{0.05cm}
    \includegraphics[width=1.0\linewidth]{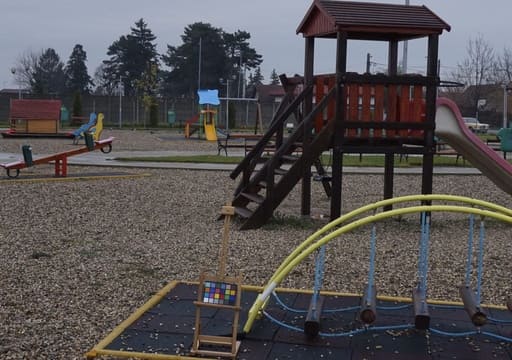}
    \captionsetup{font={small}}
    \text{Clean}
  \end{subfigure}}

  \vspace{0.1cm}

  \resizebox{0.9988\textwidth}{!}{
  \begin{subfigure}{0.151\linewidth}
    \centering
    \includegraphics[width=1.0\linewidth]{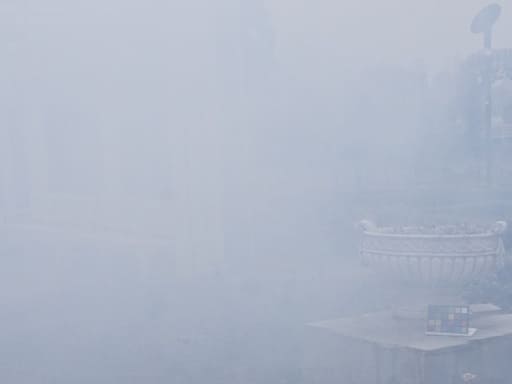}\\
    \vspace{0.05cm}
    \includegraphics[width=1.0\linewidth]{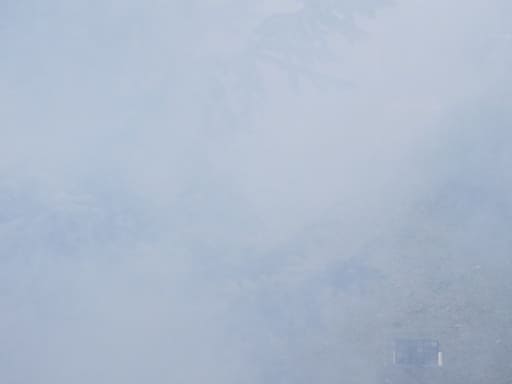}
    \captionsetup{font={small}}
    \text{Hazy}
  \end{subfigure}
  \begin{subfigure}{0.151\linewidth}
    \centering
    \includegraphics[width=1.0\linewidth]{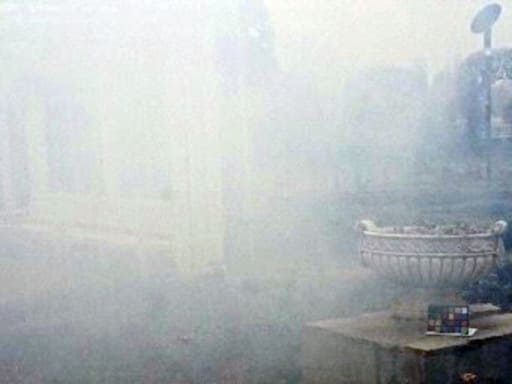}\\
    \vspace{0.05cm}
    \includegraphics[width=1.0\linewidth]{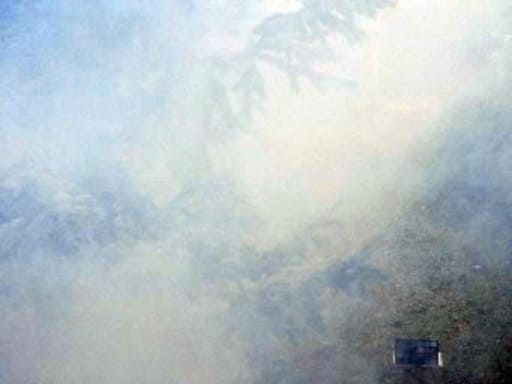}
    \captionsetup{font={small}}
    \text{GridDehaze \cite{liu2019griddehazenet}}
  \end{subfigure}
  \begin{subfigure}{0.151\linewidth}
    \centering
    \includegraphics[width=1.0\linewidth]{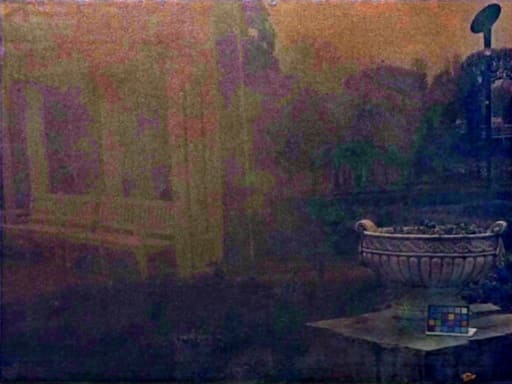}\\
    \vspace{0.05cm}
    \includegraphics[width=1.0\linewidth]{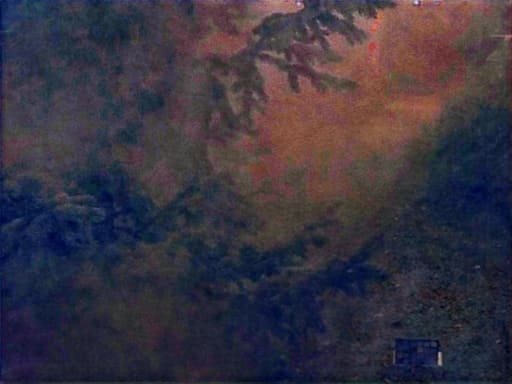}
    \captionsetup{font={small}}
    \text{GCANet \cite{chen2019gated}}
  \end{subfigure}
  \hfill
  \begin{subfigure}{0.151\linewidth}
    \centering
    \includegraphics[width=1.0\linewidth]{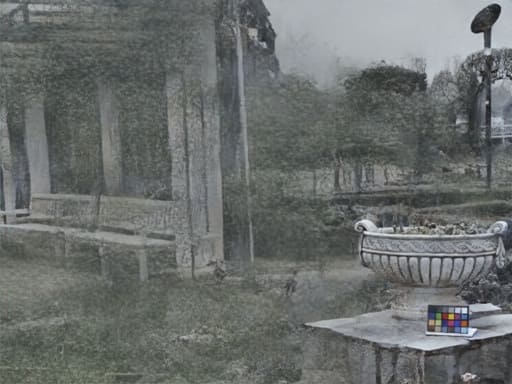}\\
    \vspace{0.05cm}
    \includegraphics[width=1.0\linewidth]{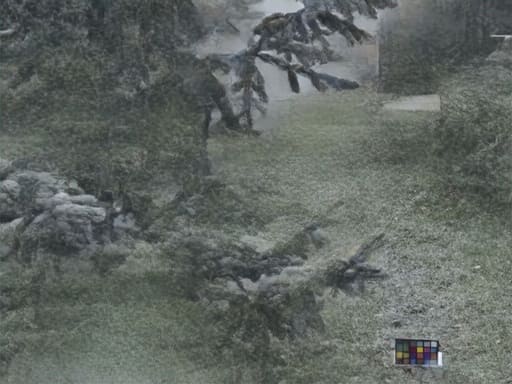}
    \captionsetup{font={small}}
    \text{EDN-GTM \cite{tran2024encoder}}
  \end{subfigure}
  \hfill
  \begin{subfigure}{0.151\linewidth}
    \centering
    \includegraphics[width=1.0\linewidth]{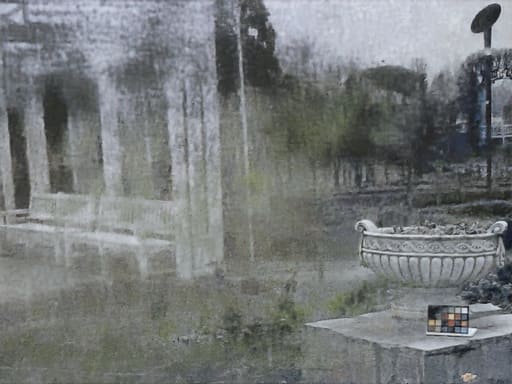}\\
    \vspace{0.05cm}
    \includegraphics[width=1.0\linewidth]{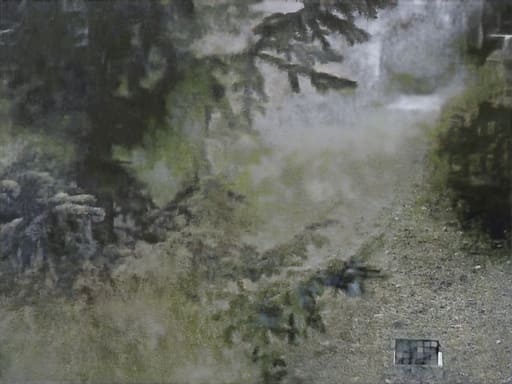}
    \captionsetup{font={small}}
    \text{Dehamer \cite{guo2022image}}
  \end{subfigure}
  \hfill
  \begin{subfigure}{0.151\linewidth}
    \centering
    \includegraphics[width=1.0\linewidth]{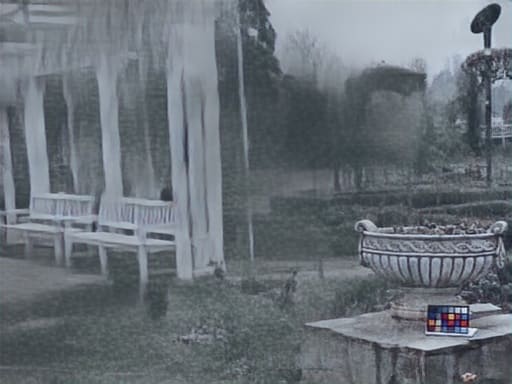}\\
    \vspace{0.05cm}
    \includegraphics[width=1.0\linewidth]{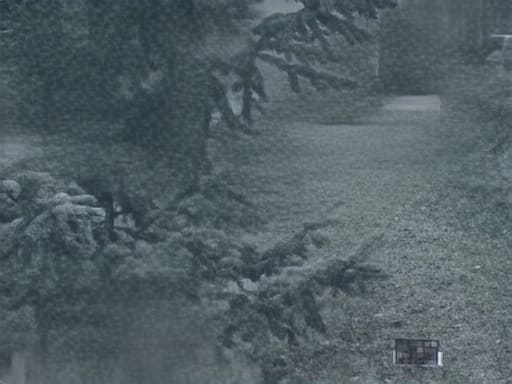}
    \captionsetup{font={small}}
    \text{DPTE-Net (ours)}
  \end{subfigure}
  \hfill
  \begin{subfigure}{0.151\linewidth}
    \centering
    \includegraphics[width=1.0\linewidth]{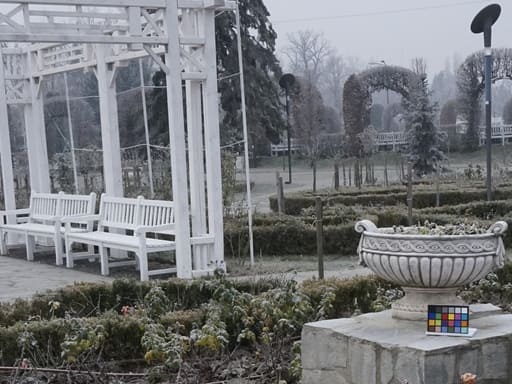}\\
    \vspace{0.05cm}
    \includegraphics[width=1.0\linewidth]{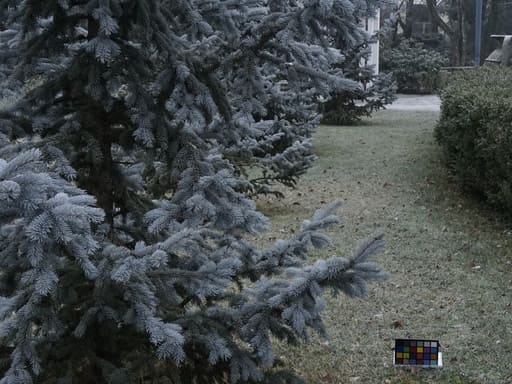}
    \captionsetup{font={small}}
    \text{Clean}
  \end{subfigure}}

  \vspace{0.1cm}

  \resizebox{0.9988\textwidth}{!}{
  \begin{subfigure}{0.151\linewidth}
    \centering
    \includegraphics[width=1.0\linewidth]{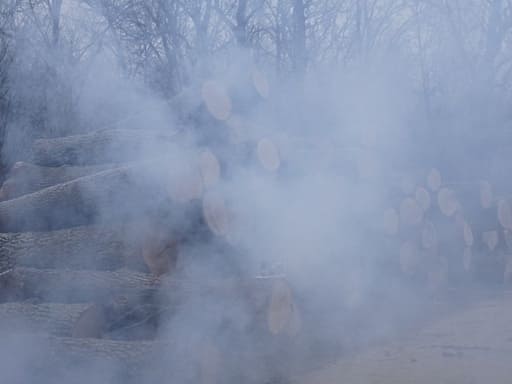}\\
    \vspace{0.05cm}
    \includegraphics[width=1.0\linewidth]{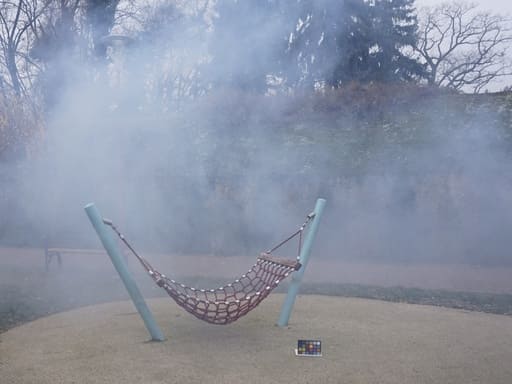}
    \captionsetup{font={small}}
    \text{Hazy}
  \end{subfigure}
  \begin{subfigure}{0.151\linewidth}
    \centering
    \includegraphics[width=1.0\linewidth]{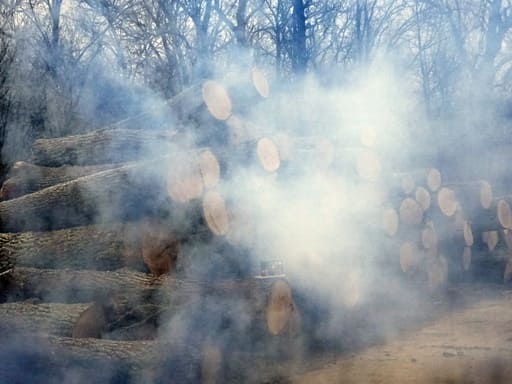}\\
    \vspace{0.05cm}
    \includegraphics[width=1.0\linewidth]{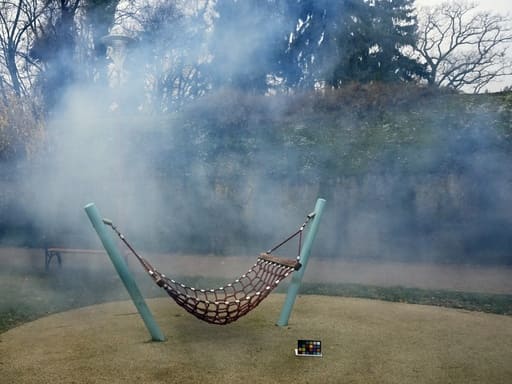}
    \captionsetup{font={small}}
    \text{GridDehaze \cite{liu2019griddehazenet}}
  \end{subfigure}
  \begin{subfigure}{0.151\linewidth}
    \centering
    \includegraphics[width=1.0\linewidth]{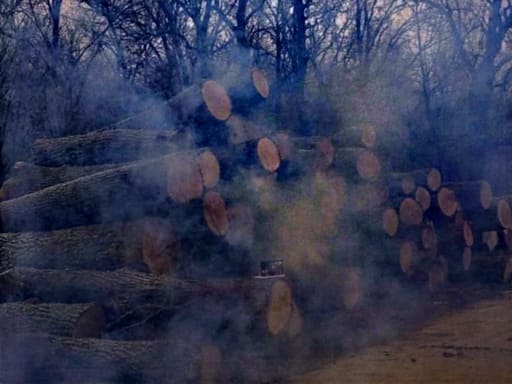}\\
    \vspace{0.05cm}
    \includegraphics[width=1.0\linewidth]{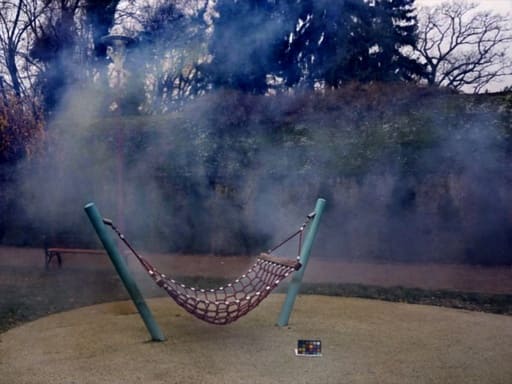}
    \captionsetup{font={small}}
    \text{GCANet \cite{chen2019gated}}
  \end{subfigure}
  \hfill
  \begin{subfigure}{0.151\linewidth}
    \centering
    \includegraphics[width=1.0\linewidth]{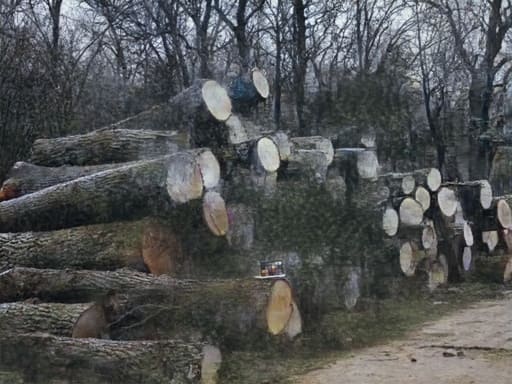}\\
    \vspace{0.05cm}
    \includegraphics[width=1.0\linewidth]{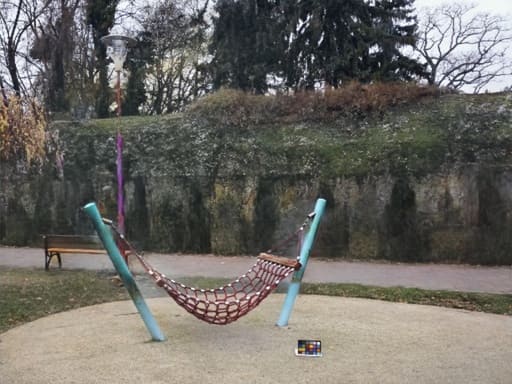}
    \captionsetup{font={small}}
    \text{EDN-GTM \cite{tran2024encoder}}
  \end{subfigure}
  \hfill
  \begin{subfigure}{0.151\linewidth}
    \centering
    \includegraphics[width=1.0\linewidth]{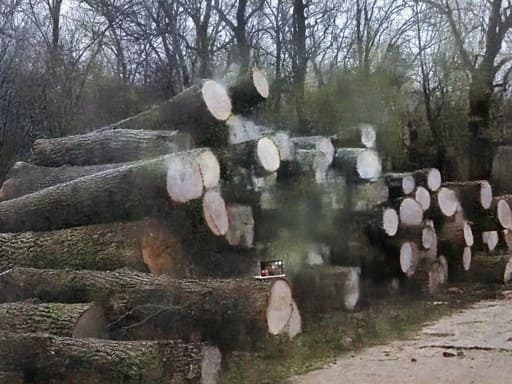}\\
    \vspace{0.05cm}
    \includegraphics[width=1.0\linewidth]{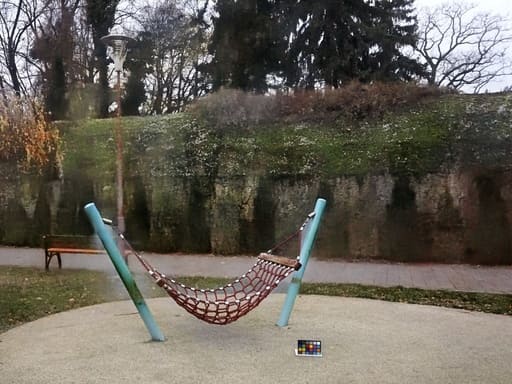}
    \captionsetup{font={small}}
    \text{Dehamer \cite{guo2022image}}
  \end{subfigure}
  \hfill
  \begin{subfigure}{0.151\linewidth}
    \centering
    \includegraphics[width=1.0\linewidth]{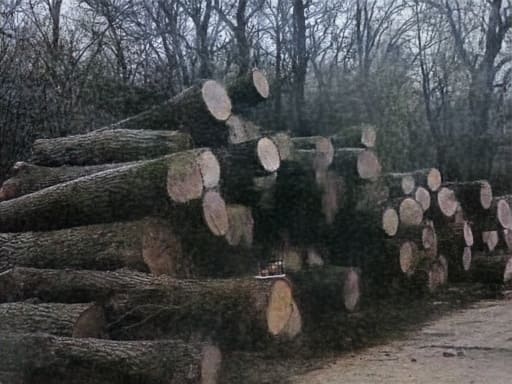}\\
    \vspace{0.05cm}
    \includegraphics[width=1.0\linewidth]{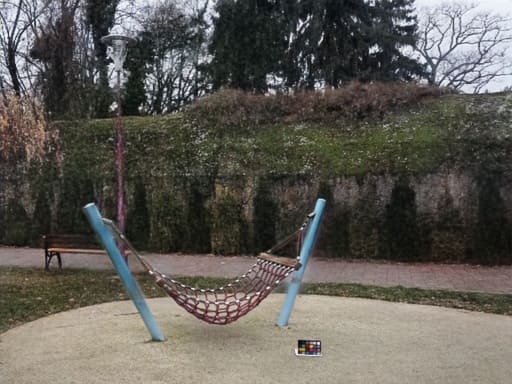}
    \captionsetup{font={small}}
    \text{DPTE-Net (ours)}
  \end{subfigure}
  \hfill
  \begin{subfigure}{0.151\linewidth}
    \centering
    \includegraphics[width=1.0\linewidth]{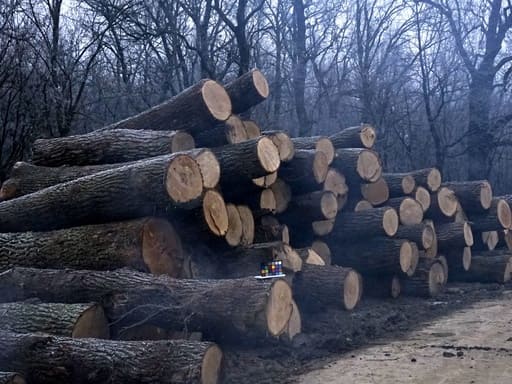}\\
    \vspace{0.05cm}
    \includegraphics[width=1.0\linewidth]{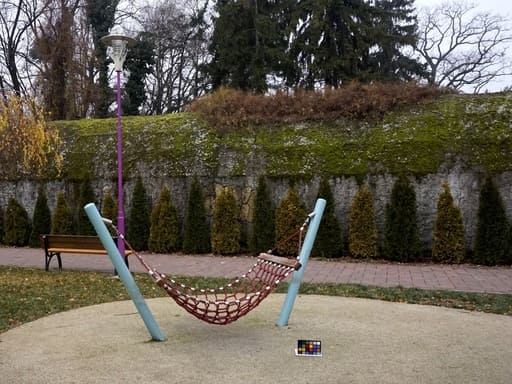}
    \captionsetup{font={small}}
    \text{Clean}
  \end{subfigure}}

  \caption{From top to bottom, visual dehazing results of different DL-based approaches on validation images from the I-HAZE, O-HAZE, Dense-HAZE, and NH-HAZE datasets, respectively.}
  \label{fig:ihaze-result-suppl}
\end{figure*}

\subsubsection{Selections of Decoder Design}
\label{subsubsec:dptenet_decoder}

In this section, we elaborate on the selection of the proposed DPTE-Net's decoder. In fact, we have examined several advanced structures integrated with the PTB module, sub-pixel convolution \cite{shi2016real}, and residual connection \cite{he2016deep} for the decoding part. However, these configurations tend to diminish the dehazing performance in our case. Meanwhile, when we merely adopt a simple U-Net decoder with a replacement of activation, a more favorable performance can be witnessed.

Several decoder variants that have been examined are described as follows. The PTB module was adopted as the core block of the decoder (PTB-Decoder) to examine the performance of a unified PTB-based dehazing architecture. On the other hand, sub-pixel convolution was selected to be a candidate since it is an efficient up-sampling operation that can diminish the grid artifact at the output \cite{wu2020knowledge} (Subpixel-Decoder). The residual connection was investigated because of its simplicity and effectiveness (Residual-Decoder). An ordinary U-Net decoder with ReLU activation (ReLU-Decoder) was utilized as the baseline for assessment and comparison. An illustration of those decoder structures is shown in Fig. \ref{fig:decoder_structures}, in which, $M$ denotes the number of convolution layers or PTB modules in every decoding level. We set $M=3$ for most cases since the teacher network has three main convolution layers in each backbone level. This can ensure that the student network has the same depth (number of principal layers) as the teacher network thereby facilitating the knowledge distillation process. However, since PTB-Decoder already has the $\mathrm{InputEmb}$ layer as a $Conv3\times3$, we set $M=2$ for this case. In the same spirit, we set $N=2$ for the encoder, where $N$ indicates the number of PTB modules in one encoding level, as shown in Fig. \ref{fig:proposed-framework}.

\begin{figure*}
  \centering  
  
  \resizebox{0.9988\textwidth}{!}{
  \begin{subfigure}{0.151\linewidth}
    \centering
    \includegraphics[width=1.0\linewidth]{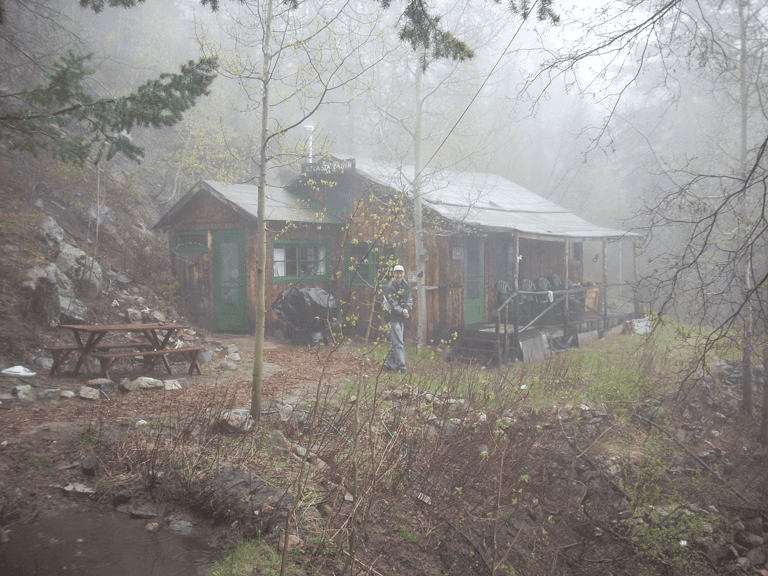}\\
    \vspace{0.05cm}
    \includegraphics[width=1.0\linewidth]{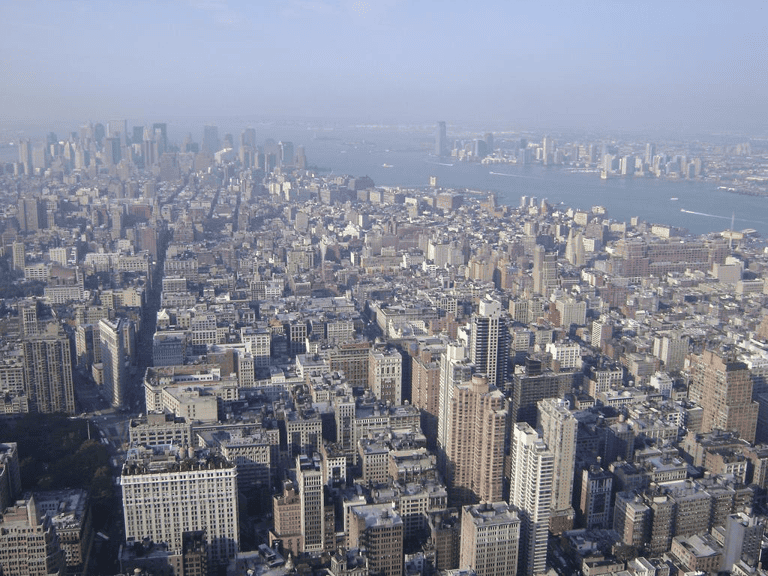}\\
    \vspace{0.05cm}
    \includegraphics[width=1.0\linewidth]{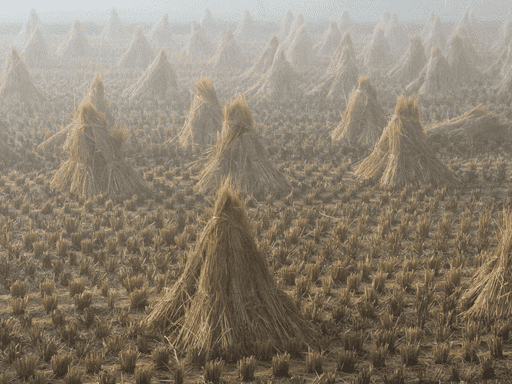}\\
    \vspace{0.05cm}
    \includegraphics[width=1.0\linewidth]{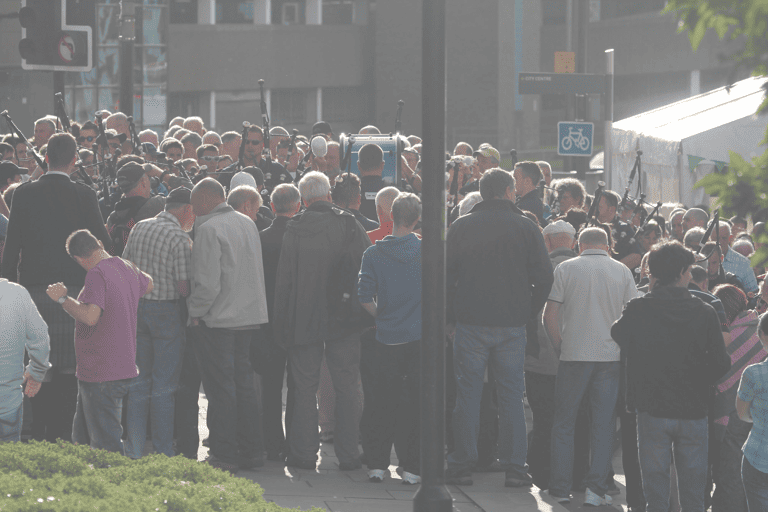}
    \captionsetup{font={small}}
    \text{Hazy}
  \end{subfigure}
  \begin{subfigure}{0.151\linewidth}
    \centering
    \includegraphics[width=1.0\linewidth]{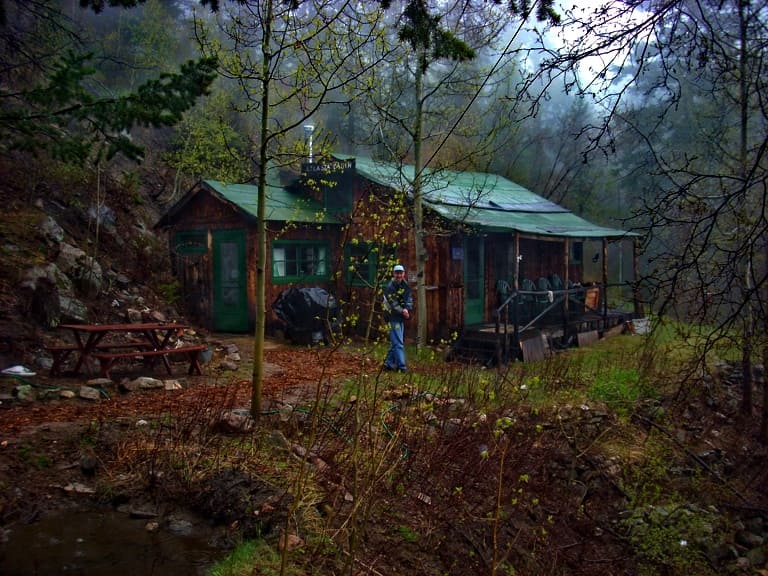}\\
    \vspace{0.05cm}
    \includegraphics[width=1.0\linewidth]{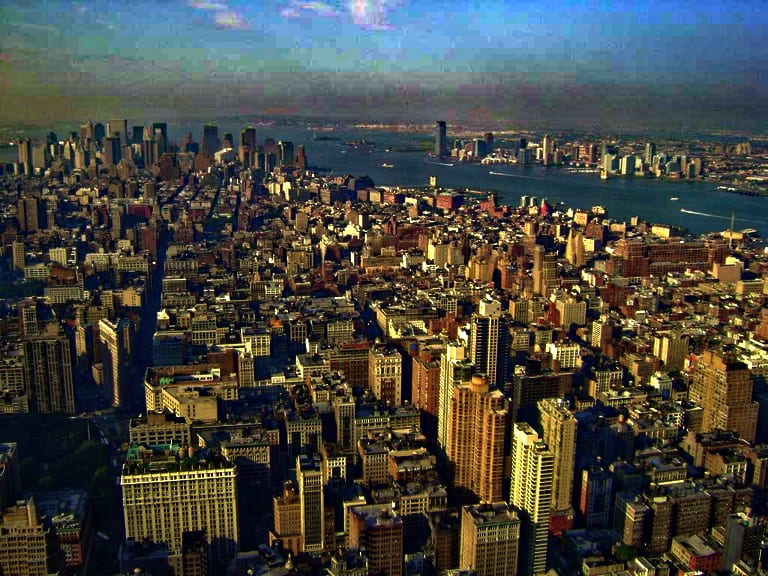}\\
    \vspace{0.05cm}
    \includegraphics[width=1.0\linewidth]{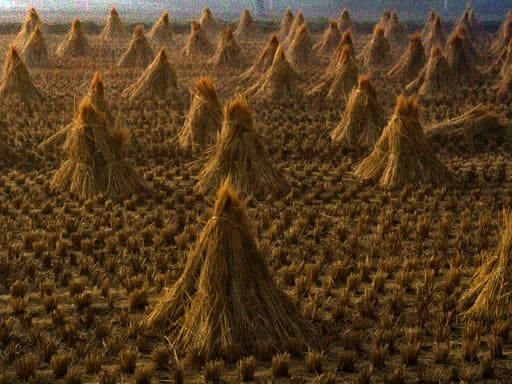}\\
    \vspace{0.05cm}
    \includegraphics[width=1.0\linewidth]{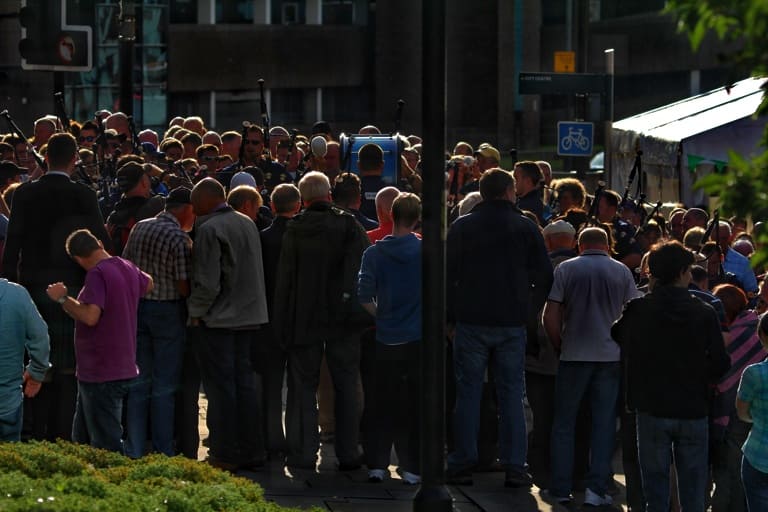}
    \captionsetup{font={small}}
    \text{CEP \cite{bui2017single}}
  \end{subfigure}
  \begin{subfigure}{0.151\linewidth}
    \centering
    \includegraphics[width=1.0\linewidth]{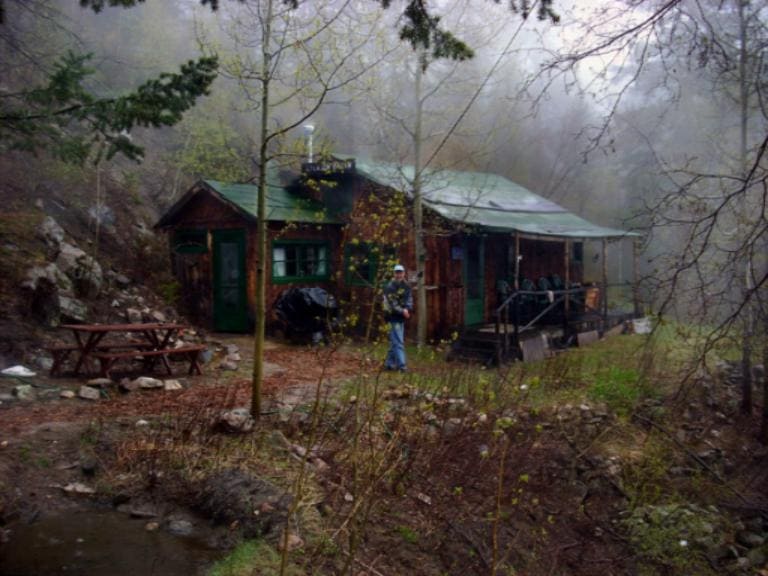}\\
    \vspace{0.05cm}
    \includegraphics[width=1.0\linewidth]{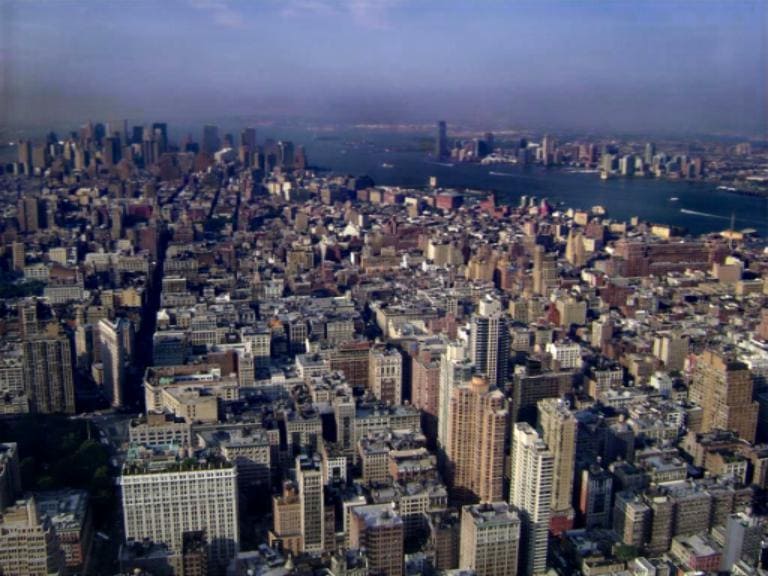}\\
    \vspace{0.05cm}
    \includegraphics[width=1.0\linewidth]{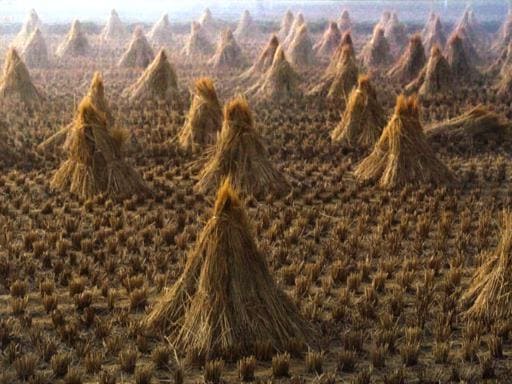}\\
    \vspace{0.05cm}
    \includegraphics[width=1.0\linewidth]{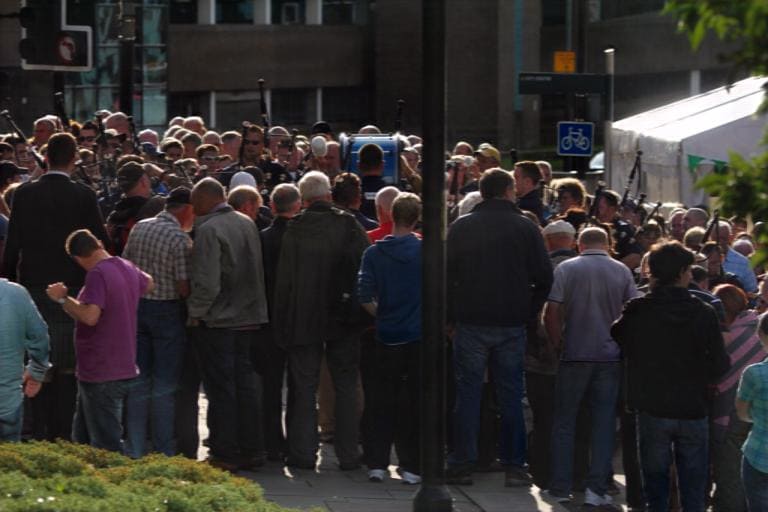}
    \captionsetup{font={small}}
    \text{GCANet \cite{chen2019gated}}
  \end{subfigure}
  \hfill
  \begin{subfigure}{0.151\linewidth}
    \centering
    \includegraphics[width=1.0\linewidth]{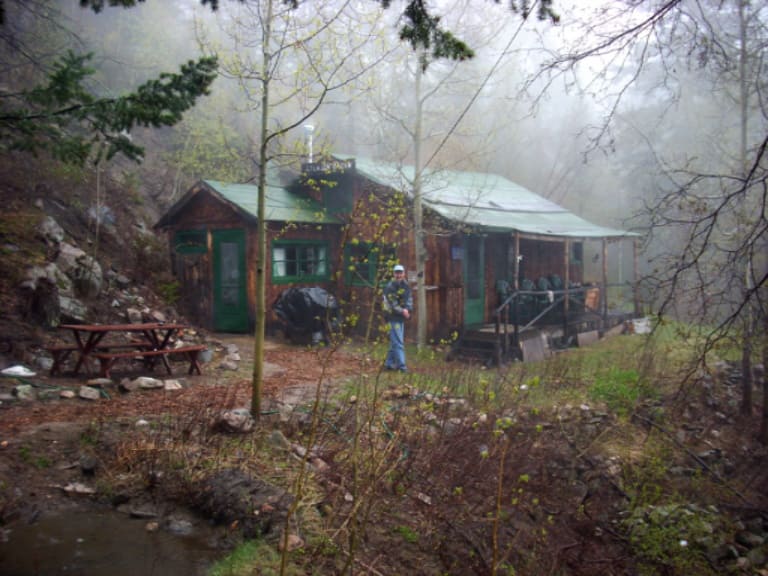}\\
    \vspace{0.05cm}
    \includegraphics[width=1.0\linewidth]{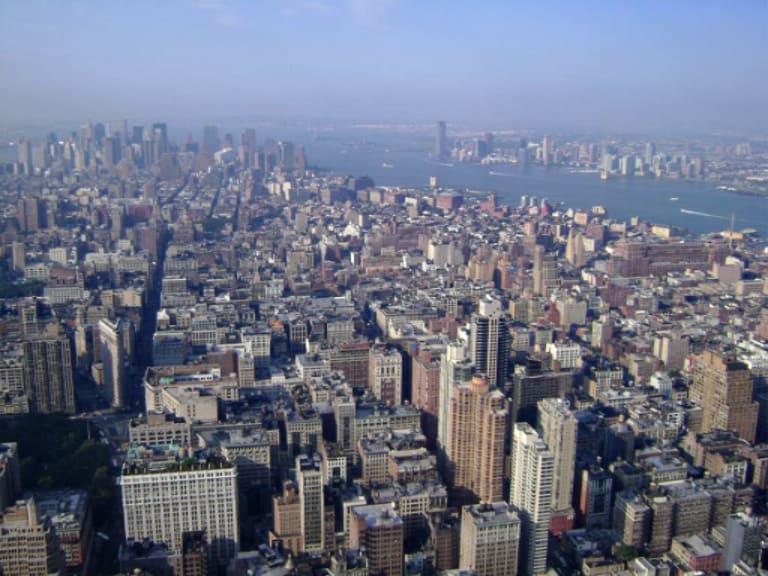}\\
    \vspace{0.05cm}
    \includegraphics[width=1.0\linewidth]{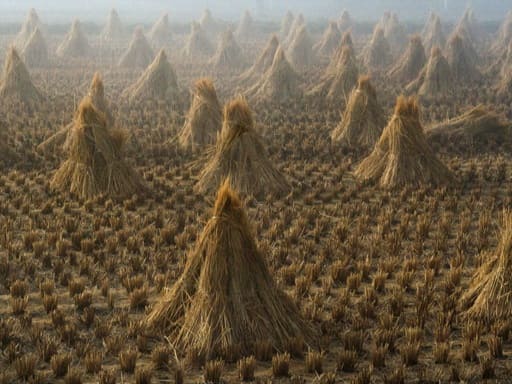}\\
    \vspace{0.05cm}
    \includegraphics[width=1.0\linewidth]{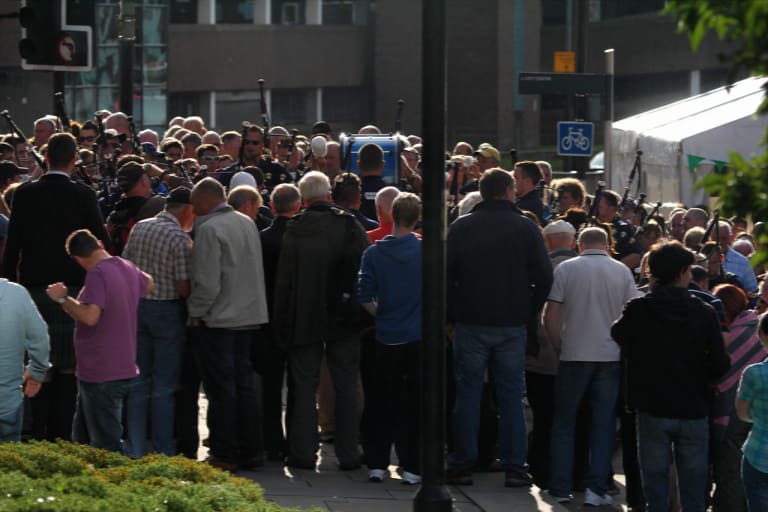}
    \captionsetup{font={small}}
    \text{EDN-GTM \cite{tran2024encoder}}
  \end{subfigure}
  \hfill
  \begin{subfigure}{0.151\linewidth}
    \centering
    \includegraphics[width=1.0\linewidth]{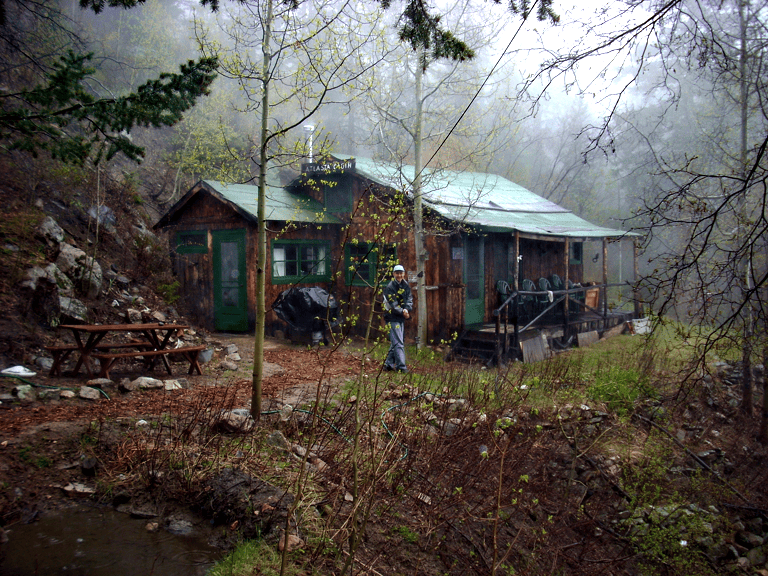}\\
    \vspace{0.05cm}
    \includegraphics[width=1.0\linewidth]{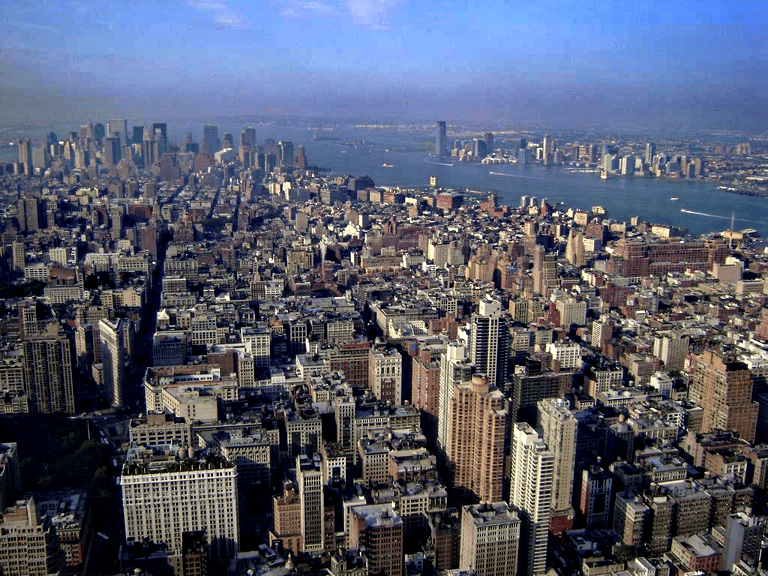}\\
    \vspace{0.05cm}
    \includegraphics[width=1.0\linewidth]{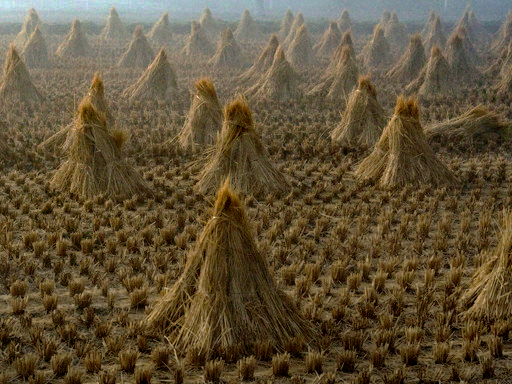}\\
    \vspace{0.05cm}
    \includegraphics[width=1.0\linewidth]{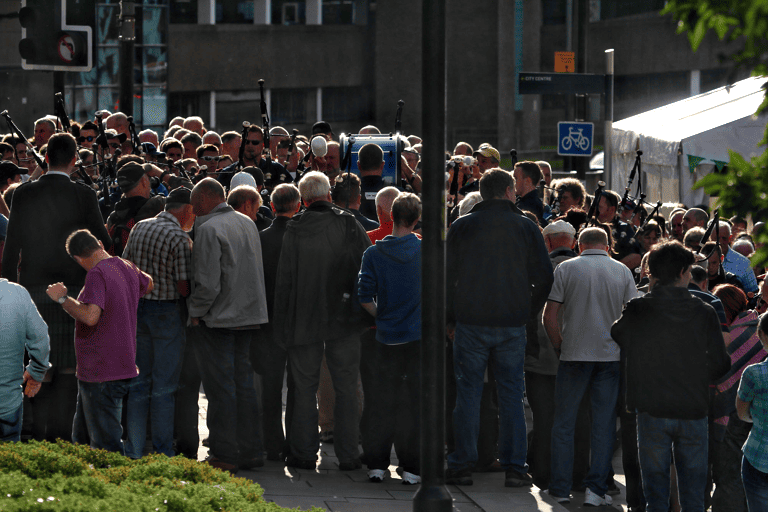}
    \captionsetup{font={small}}
    \text{Dehamer \cite{guo2022image}}
  \end{subfigure}
  \hfill
  \begin{subfigure}{0.151\linewidth}
    \centering
    \includegraphics[width=1.0\linewidth]{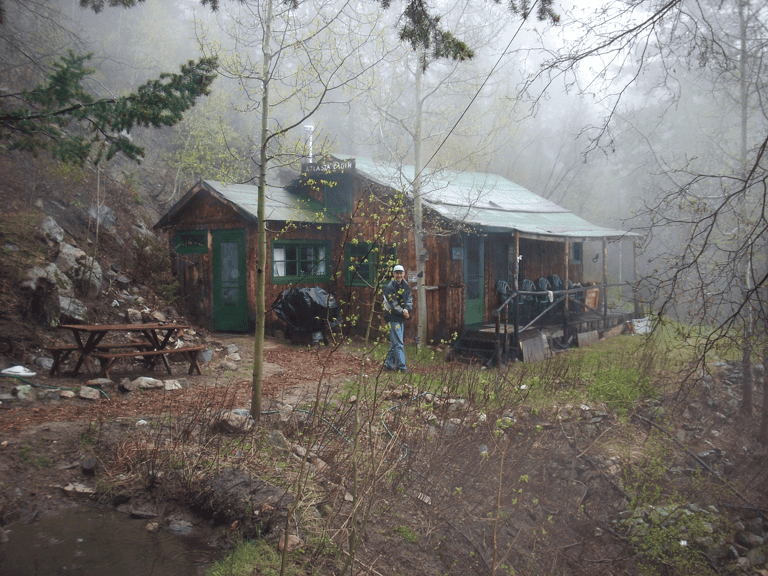}\\
    \vspace{0.05cm}
    \includegraphics[width=1.0\linewidth]{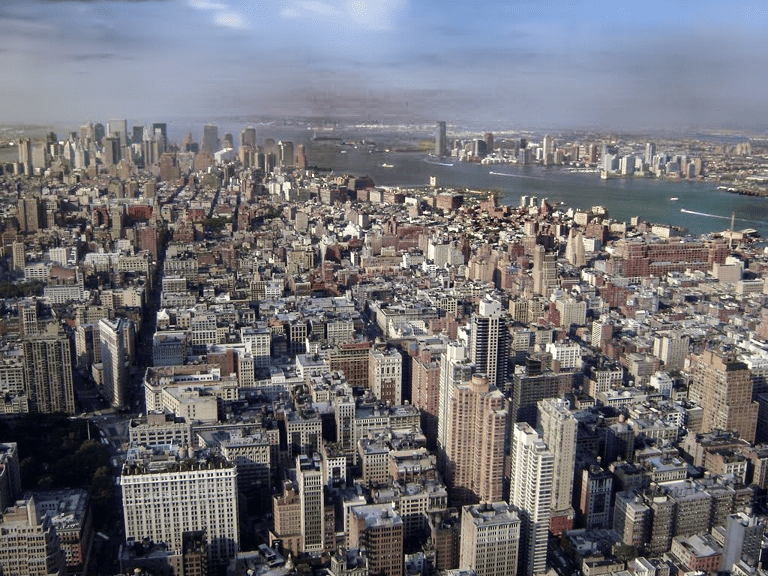}\\
    \vspace{0.05cm}
    \includegraphics[width=1.0\linewidth]{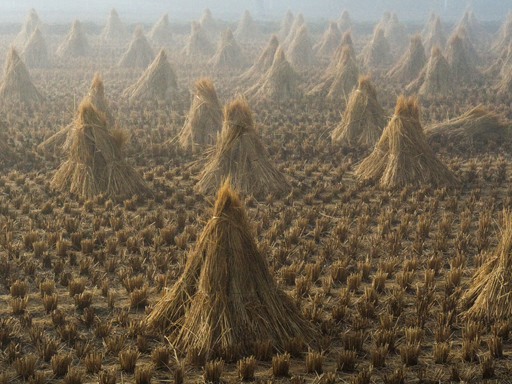}\\
    \vspace{0.05cm}
    \includegraphics[width=1.0\linewidth]{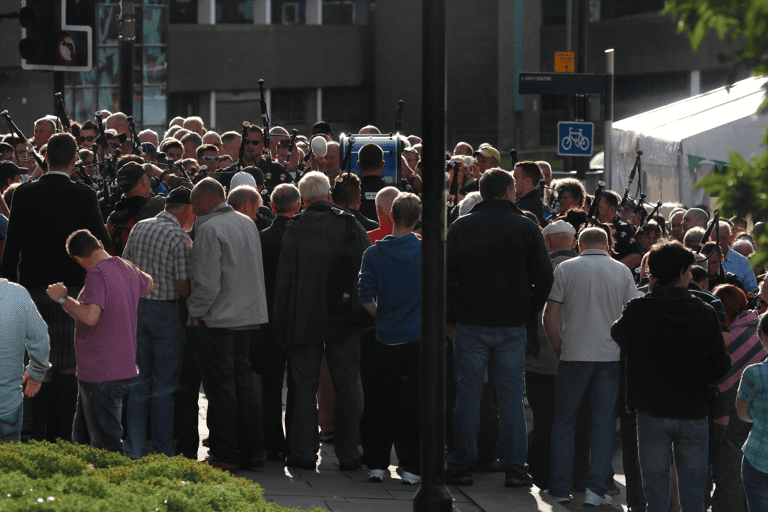}
    \captionsetup{font={small}}
    \text{DehazeFormer \cite{song2023vision}}
  \end{subfigure}
  \hfill
  \begin{subfigure}{0.151\linewidth}
    \centering
    \includegraphics[width=1.0\linewidth]{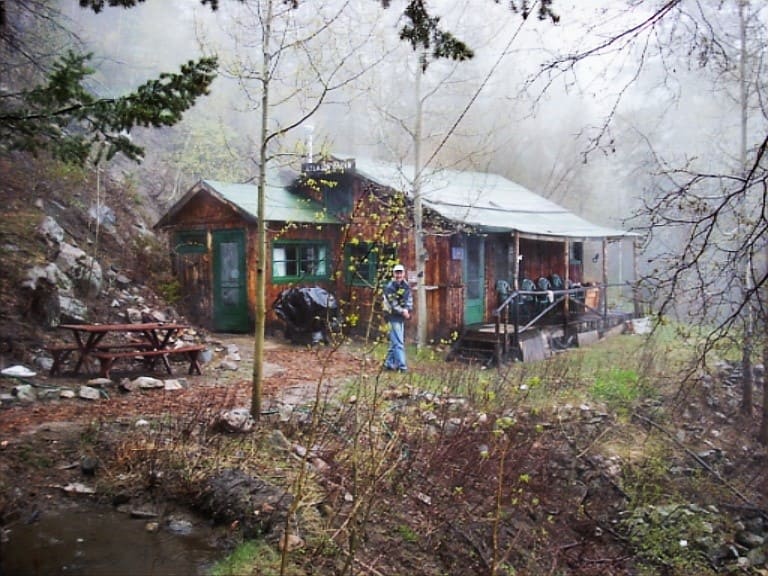}\\
    \vspace{0.05cm}
    \includegraphics[width=1.0\linewidth]{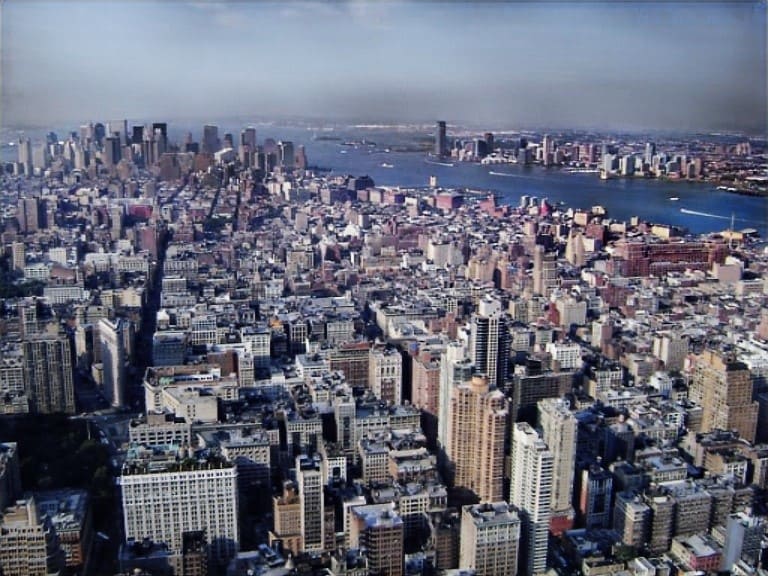}\\
    \vspace{0.05cm}
    \includegraphics[width=1.0\linewidth]{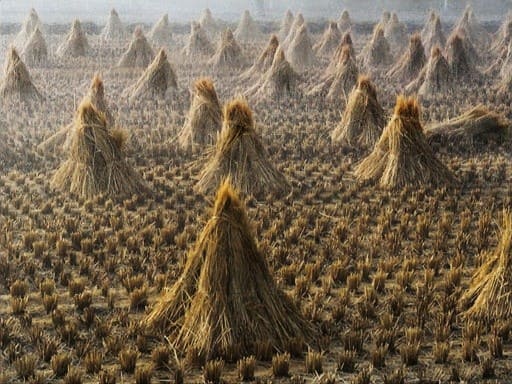}\\
    \vspace{0.05cm}
    \includegraphics[width=1.0\linewidth]{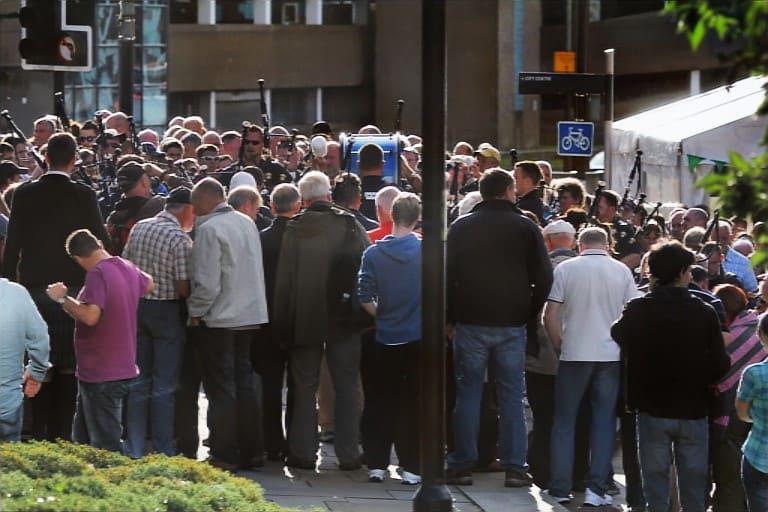}
    \captionsetup{font={small}}
    \text{DPTE-Net (ours)}
  \end{subfigure}}
  
  \caption{Qualitative results of various methods in real-world haze scenarios.}
  \label{fig:ablationasdas}
  
\end{figure*}

The effects of the aforementioned decoding structures on the dehazing performance are reported in Table \ref{tab:decoder}. As summarized in Table \ref{tab:decoder}, various decoder designs lead to minor variations in terms of \#Params and MACs, yet produce distinct outcomes in terms of dehazing effectiveness. It can be observed that the PTB module is ineffective for decoding, this may stem from the fact that PTB is based on the ViT encoder structure which is mainly developed for feature encoding \cite{yu2022metaformer} and may be inapplicable to be used in a decoding flow. Residual-Decoder, on the other hand, produces a marginal performance gain as compared to the baseline. Subpixel-Decoder offers a minor reduction in computational cost by utilizing the shuffle operation instead of the learnable transposed convolution during the up-sampling process. Despite this modest enhancement in efficiency, there is a trade-off with a downgrade in effectiveness. On the other hand, with a replacement of activation (ReLU $\rightarrow$ Swish), Swish-Decoder demonstrates a considerable performance improvement over the baseline ReLU-Decoder without affecting computational complexity. The impact of different decoder types on visual results is illustrated in Fig. \ref{fig:ablation_decoder}b - Fig. \ref{fig:ablation_decoder}e and Fig. \ref{fig:ablation_decoder}g.

\subsubsection{Adaptation Layer with PTB}
\label{subsubsec:adaptation}

This section delves into the analysis of the implementation of PTB in the adaptation layer. The purpose of an adaptation layer is typically to align the feature scales (number of neurons or feature map shape) of the student network with those of the teacher model. Additionally, it facilitates a seamless transition between the feature spaces of the student and teacher models, thereby enabling effective knowledge transfer. A $Conv1\times1$ layer is typically utilized as the adaptation layer to conserve memory. However, in an ill-posed challenging image restoration task like dehazing, relying solely on a single $Conv1\times1$ layer may be insufficient. Therefore, we incorporate the PTB module into the adaptation layer to leverage the feature learning and propagating prowess of ViT for feature space transition, followed by a $Conv1\times1$ layer for matching the feature shapes. To validate this approach, we compare the performance of this setting with that of an adaptation layer consisting only of a $Conv1\times1$ layer (without PTB). As the quantitative results summarized in Table \ref{tab:adaptation}, integrating the PTB module into the adaptation layer can facilitate an effective knowledge transfer process and result in improved dehazing performance. The impact of this configuration on the visibility of the restored image is illustrated in Fig. \ref{fig:ablation_decoder}f and Fig. \ref{fig:ablation_decoder}g.

\begin{table}
  \centering
  \caption{Effects of PTB in adaptation layer.}
  \resizebox{0.48\textwidth}{!}{%
  \begin{tabular}{cccccrr}
    \toprule
    \multirow{2}{*}{Adaptation} & \multicolumn{2}{c}{Dense-HAZE} & \multicolumn{2}{c}{NH-HAZE} & \multirow{2}{*}{\#Params$\downarrow$} & \multirow{2}{*}{MACs$\downarrow$}  \\
    \cmidrule{2-5}
        & PSNR & SSIM & PSNR & SSIM \\
    \midrule
    $Conv1\times1$ & 14.92 & 0.5065 & 19.13 & 0.5298 & \textbf{2.99M} & \textbf{19.0G} \\
    +PTB & \textbf{15.59} & \textbf{0.5248} & \textbf{20.18} & \textbf{0.5623} & 3.10M & 19.2G \\
    \bottomrule
  \end{tabular}}
  \label{tab:adaptation}
\end{table}

\subsubsection{Further Visual Comparisons}
\label{subsubsec:furthervisual}

In order to provide a more comprehensive evaluation of the proposed network's performance, additional visual outcomes generated by DPTE-Net and various state-of-the-art DL-based methods are presented in Fig. \ref{fig:ihaze-result-suppl}. As can be observed from Fig. \ref{fig:ihaze-result-suppl}, DPTE-Net can surpass existing methods such as GridDehaze and CGANet, while also achieving competitive visual quality as compared to other modern approaches with higher complexity including EDN-GTM, PPD-Net, and Dehamer.

In addition, a qualitative comparison on additional real-world hazy data was conducted to provide a more comprehensive evaluation of DPTE-Net’s performance. For this purpose, we trained the network on all outdoor images from the four datasets and compared DPTE-Net’s outputs with those of recent dehazing methods in realistic outdoor hazy scenes. Fig. \ref{fig:ablationasdas} presents the qualitative comparison between DPTE-Net and recent models. The results in Fig. \ref{fig:ablationasdas} highlight DPTE-Net’s adaptability and applicability across diverse environmental conditions, further supporting its potential for deployment in real-world dehazing applications.

\subsubsection{Limitations}
\label{subsubsec:limitations}

\begin{figure}
  \centering  
  \resizebox{0.49\textwidth}{!}{
  \begin{subfigure}{0.33\linewidth}
    \centering  
    \includegraphics[width=1.0\linewidth]{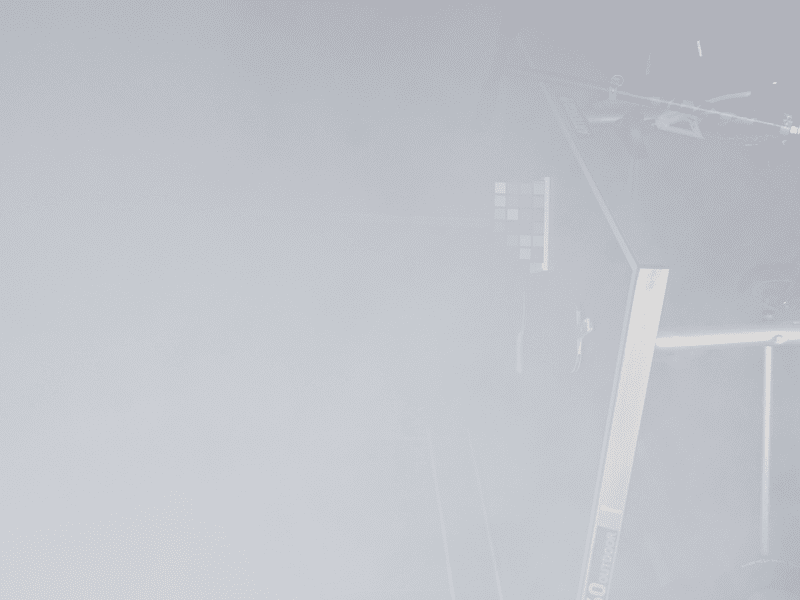}\\
    \vspace{0.10cm}
    \includegraphics[width=1.0\linewidth]{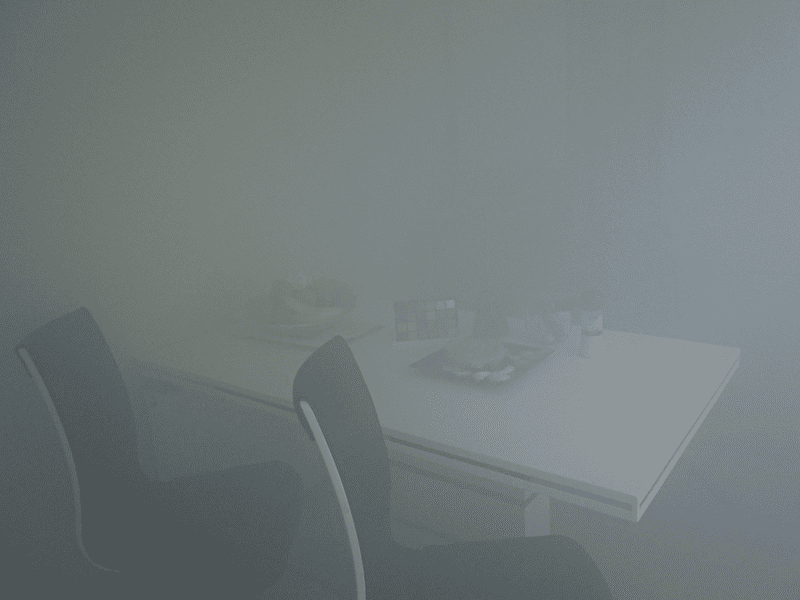}\\
    \text{\small Hazy}
  \end{subfigure}
  \begin{subfigure}{0.33\linewidth}
    \centering  
    \includegraphics[width=1.0\linewidth]{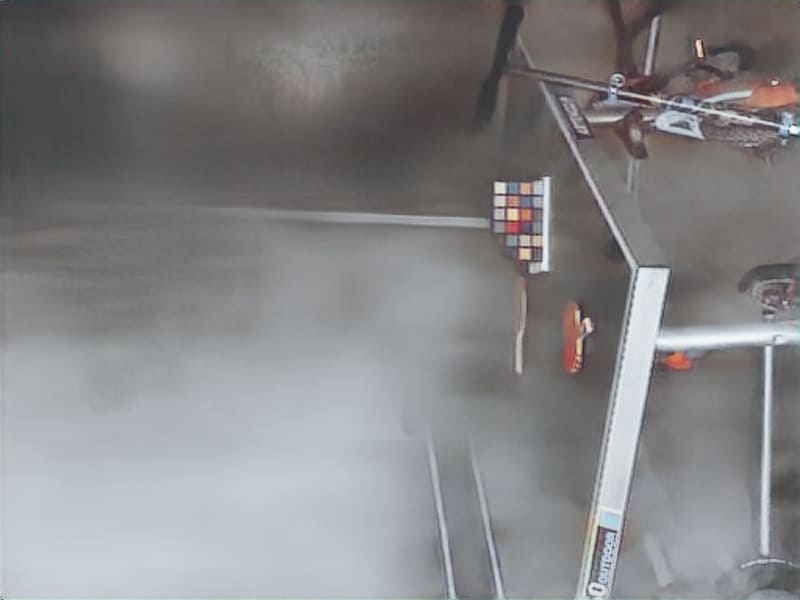}\\
    \vspace{0.10cm}
    \includegraphics[width=1.0\linewidth]{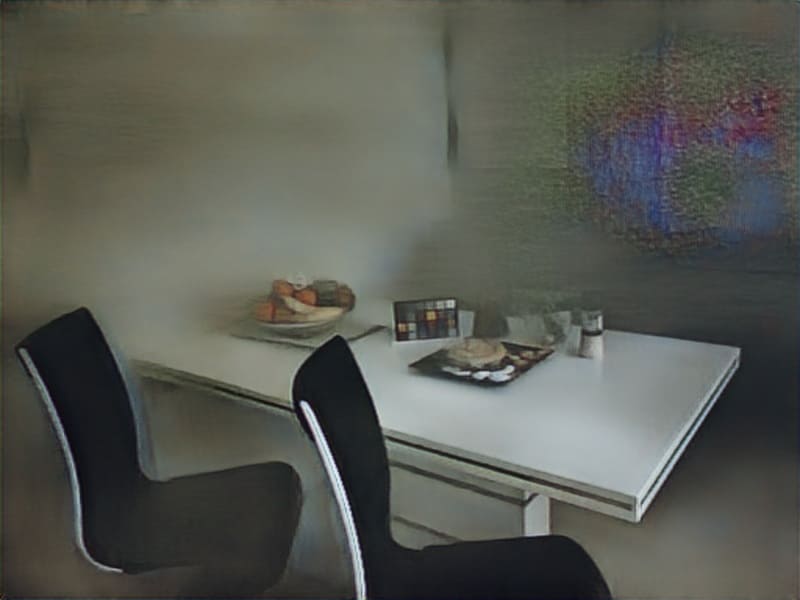}\\
    \text{\small DPTE-Net (ours)}
  \end{subfigure}
  \begin{subfigure}{0.33\linewidth}
    \centering  
    \includegraphics[width=1.0\linewidth]{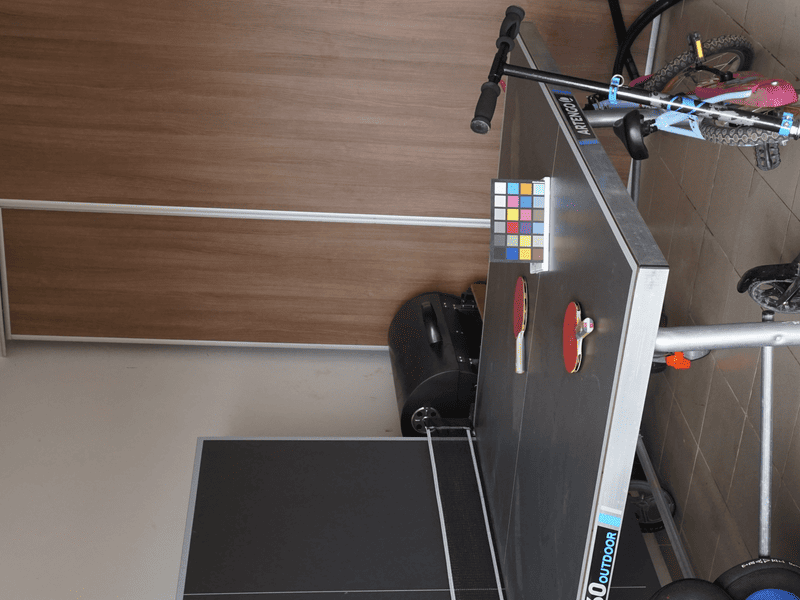}\\
    \vspace{0.10cm}
    \includegraphics[width=1.0\linewidth]{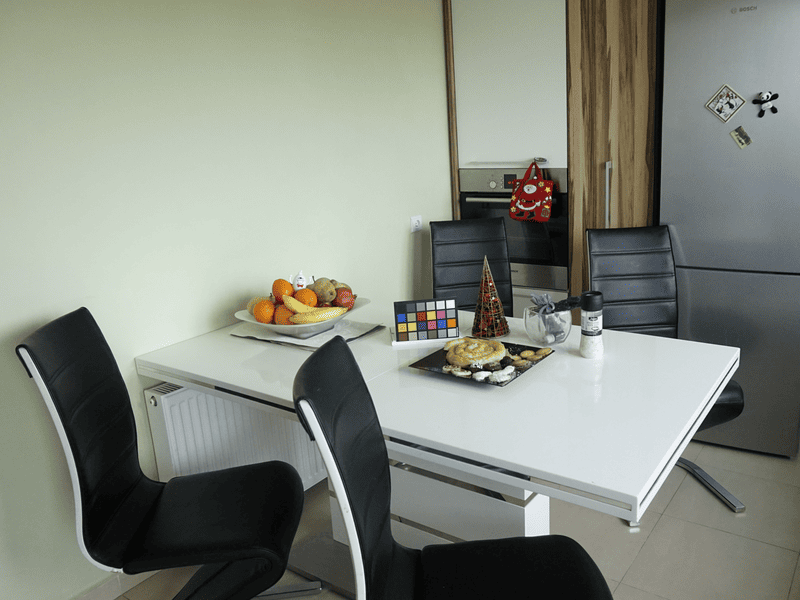}\\
    \text{\small Clean}
  \end{subfigure}}  
  \caption{Typical suboptimal results in cases of dense haze and unclear backgrounds.}
  \label{fig:limitations}
\end{figure}

Although the proposed network can produce competitive performance when compared against state-of-the-art approaches, the model still faces challenges that hinder its overall effectiveness. Akin to other existing lightweight dehazing methods, the proposed DPTE-Net particularly encounters difficulties in handling scenes with dense haze and unclear backgrounds, as illustrated in Fig. \ref{fig:limitations}. These shortcomings suggest that the framework may necessitate additional refinements to effectively deal with the varied and intricate nature of real-world haze distributions.

To enhance the generalizability of the model, future research will focus on expanding the dataset to capture a broader spectrum of haze scenarios, improving the student network’s architecture, and adopting advanced learning approaches such as meta learning \cite{ma2022flexible}. By addressing these challenges, it is expected that the framework's reliability will be improved, leading to more consistent dehazing outcomes across diverse environmental conditions.

\section{Conclusions}
\label{sec:conclusions}

In this paper, an efficient image dehazing network with a Distilled Pooling Transformer Encoder, named DPTE-Net, is proposed. DPTE-Net consists of an attention-free ViT-based encoder with pooling as the token mixing operation to reduce complexity. Knowledge from a same-task teacher model is aptly exploited via a two-stage training process to enrich the semantic features encoded at the bottleneck. The proposed DPTE-Net is trained based on a GAN framework to leverage the generalization power of image-generation models for dehazing task. A transmission-aware loss is adopted which utilizes the transmission feature to optimize the network based on different haze densities. Experimental results on various benchmark datasets have shown that DPTE-Net can achieve a competitive trade-off between performance and complexity as compared to other state-of-the-art image dehazing approaches. Future research will focus on improving the student network's architecture, and applying advanced methods like meta-learning to enhance model's generalizability.





\bibliographystyle{unsrt}
\bibliography{bibliography_v2}

\end{document}